\documentclass{article}[11pt] 

\usepackage{amsmath,amsfonts,bm}

\def\eqref#1{equation~\ref{#1}}

\def\1{\bm{1}}

\def\eps{{\epsilon}}

\def\vzero{{\bm{0}}}

\def\vb{{\bm{b}}}
\def\vc{{\bm{c}}}

\def\ve{{\bm{e}}}

\def\vu{{\bm{u}}}
\def\vv{{\bm{v}}}
\def\vw{{\bm{w}}}
\def\vx{{\bm{x}}}
\def\vy{{\bm{y}}}

\def\mA{{\bm{A}}}
\def\mB{{\bm{B}}}
\def\mC{{\bm{C}}}
\def\mD{{\bm{D}}}
\def\mE{{\bm{E}}}

\def\mG{{\bm{G}}}
\def\mH{{\bm{H}}}
\def\mI{{\bm{I}}}

\def\mK{{\bm{K}}}
\def\mL{{\bm{L}}}
\def\mM{{\bm{M}}}
\def\mN{{\bm{N}}}

\def\mP{{\bm{P}}}

\def\mR{{\bm{R}}}

\def\mT{{\bm{T}}}
\def\mU{{\bm{U}}}
\def\mV{{\bm{V}}}
\def\mW{{\bm{W}}}
\def\mX{{\bm{X}}}

\def\mZ{{\bm{Z}}}

\def\mSigma{{\bm{\Sigma}}}

\DeclareMathAlphabet{\mathsfit}{\encodingdefault}{\sfdefault}{m}{sl}
\SetMathAlphabet{\mathsfit}{bold}{\encodingdefault}{\sfdefault}{bx}{n}

\newcommand{\R}{\mathbb{R}}

\DeclareMathOperator*{\argmin}{arg\,min}

\usepackage{xcolor}
\usepackage{amssymb,enumerate,graphicx,verbatim,natbib,amsthm}
\usepackage{thmtools}
\usepackage{thm-restate}
\usepackage{hyperref}
\usepackage{subcaption}
\usepackage{mwe}
\usepackage[toc,page,header]{appendix}
\usepackage{minitoc}
\definecolor{amethyst}{rgb}{0.6, 0.4, 0.8}
\hypersetup{
    colorlinks=true,
    linkcolor=purple,
    filecolor=magenta,      
    urlcolor=cyan,
    citecolor=amethyst,
}
\usepackage{url}
\usepackage[nameinlink,noabbrev,capitalise]{cleveref}
\Crefname{table}{Table}{Tables}
\Crefname{assumption}{Assumption}{Assumptions}
\crefname{equation}{}{}
\newtheorem{theorem}{Theorem}[section]
\newtheorem{lemma}{Lemma}[section]
\newtheorem{assumption}{Assumption}[section]
\newtheorem{remark}{Remark}[section]
\newtheorem{proposition}{Proposition}[section]

\newtheorem{definition}{Definition}[section]

\newtheorem{corollary}{Corollary}[section]

\newcommand{\rank}[1]{\operatorname{rank}(#1)}
\newcommand{\A}{\mathcal{A}}
\renewcommand{\O}{\mathcal{O}}
\renewcommand{\d}{\mathrm{d}}
\newcommand{\diag}[1]{\operatorname{diag}\!\left(#1\right)}	
\newcommand{\tr}[1]{\operatorname{tr}\left(#1\right)}

\newcommand{\norm}[1]{\left\| #1 \right\|}
\newcommand{\normsm}[1]{\| #1 \|}

\newcommand{\normtwosm}[1]{\normsm{#1}_2}

\newcommand{\normF}[1]{\norm{#1}_{\mathrm{F}}}
\newcommand{\normFsm}[1]{\normsm{#1}_{\mathrm{F}}}

\newcommand{\hr}{\hat{r}}
\newcommand{\bmU}{\bar{\mU}}

\newcommand{\mDelta}{{\bm \Delta}}

\newcommand{\PLfull}{Polyak-\L{}ojasiewicz}

\newenvironment{proof-sketch}{\noindent\textbf{\textit{Proof sketch}:}}{\hfill$\square$}
\renewenvironment{proof}{\noindent\textbf{\textit{Proof }:}}{\hfill$\square$}

\renewcommand{\sp}{\mathtt{sp}}
\newcommand{\tpi}{\mathtt{pi}}
\renewcommand{\mSigma}{\bm{\Sigma}}

\title{Understanding Incremental Learning of Gradient Descent: A Fine-grained Analysis of Matrix Sensing}

\author{Jikai Jin \thanks{Peking University. Email: \texttt{jkjin@pku.edu.cn}}
~~~~Zhiyuan Li\thanks{Stanford University. Email: \texttt{zhiyuanli@stanford.edu}}~~~~Kaifeng Lyu\thanks{Princeton University. Email: \texttt{klyu@cs.princeton.edu}}~~~~Simon S. Du\thanks{University of Washington. Email: \texttt{ssdu@cs.washington.edu}}~~~~Jason D. Lee\thanks{Princeton University. Email: \texttt{jasonlee@princeton.edu}}}

\newcommand{\kaifeng}[1]{{\color{orange} Kaifeng:#1}}

\usepackage{times}
\usepackage[fontsize=11pt]{scrextend}
\usepackage[letterpaper,left=1.35in,right=1.35in,top=1.4in,bottom=1.5in]{geometry}

\begin{document}
\doparttoc 
\faketableofcontents 


\maketitle

\begin{abstract}
    It is believed that Gradient Descent (GD) induces an implicit bias towards good generalization in training machine learning models. This paper provides a fine-grained analysis of the dynamics of GD for the matrix sensing problem, whose goal is to recover a low-rank ground-truth matrix from near-isotropic linear measurements. It is shown that GD with small initialization behaves similarly to the greedy low-rank learning heuristics~\citep{li2020towards} and follows an incremental learning procedure~\citep{gissin2019implicit}: GD sequentially learns solutions with increasing ranks until it recovers the ground truth matrix. Compared to existing works which only analyze the first learning phase for rank-1 solutions, our result provides characterizations for the whole learning process. Moreover, besides the over-parameterized regime that many prior works focused on, our analysis of the incremental learning procedure also applies to the \textit{under-parameterized} regime. Finally, we conduct numerical experiments to confirm our theoretical findings.
\end{abstract}

\section{Introduction}
\label{sec-intro}
Understanding the optimization and generalization properties of optimization
algorithms is one of the central topics in deep learning theory
\citep{zhang2017understanding,sun2019optimization}. It has long been a mystery
why simple algorithms such as Gradient Descent (GD) or Stochastic Gradient Descent (SGD) can find global minima even
for highly non-convex functions~\citep{du2019gradient}, and why the global
minima being found can generalize well~\citep{hardt2016train}.

One influential line of works provides theoretical analysis of 
the {\em implicit bias} of GD/SGD.
These results typically exhibit theoretical settings
where the low-loss solutions found by GD/SGD
attain certain optimality conditions of a particular generalization metric,
\textit{e.g.}, the parameter norm (or the classifier margin)~\citep{soudry2018implicit,gunasekar2018characterizing,nacson2019lexicographic,lyu2020gradient,ji2020directional},
the sharpness of local loss landscape~\citep{blanc2020ou,damian2021label,li2022what,lyu2022understanding}.

Among these works, a line of works seek to characterize the implicit bias even
when the training is away from convergence. \citet{kalimeris2019sgd} empirically
observed that SGD learns model from simple ones, such as linear classifiers, to
more complex ones. As a result, SGD always tries to fit the training data with
minimal model complexity. This behavior, usually referred to as the {\em
simplicity bias} or the {\em incremental learning} behavior of GD/SGD, can be a
hidden mechanism of deep learning that prevents highly over-parameterized models
from overfitting.
In theory,
\citet{hu2020surprising,lyu2021gradient,frei2021provable}
established that GD on two-layer nets learns linear classifiers first.

The goal of this paper is to demonstrate this simplicity bias/incremental
learning in the {\em matrix sensing} problem, a non-convex optimization problem
that arises in a wide range of real-world applications, {\em e.g.}, image
reconstruction \citep{zhao2010low,peng2014reweighted}, object detection
\citep{shen2012unified,zou2013segmentation} and array processing systems
\citep{kalogerias2013matrix}. Moreover, this problem can serve as a standard
test-bed of the implicit bias of GD/SGD in deep learning theory, since it
retains many of the key phenomena in deep learning while being simpler to
analyze.

Formally, the matrix sensing problem asks for recovering a ground-truth matrix $\mZ^* \in \R^{d \times d}$ given $m$ observations $y_1, \dots, y_m$.
Each observation $y_i$ here is resulted from a linear measurement $y_i = \left\langle \mA_i, \mZ^* \right\rangle$,
where $\{\mA_i\}_{1\leq i\leq m}$ is a collection of symmetric measurement matrices.
In this paper, we focus on the case where $\mZ^*$ is symmetric,
positive semi-definite (PSD) and low-rank, {\em i.e.}, $\mZ^* \succeq \vzero$ and $\rank{\mZ^*}=r_{*}
\ll d$. 

An intriguing approach to solve this problem is to use 
the Burer-Monteiro type decomposition $\mZ^* = \mU \mU^{\top}$ with
$\mU \in \R^{d \times \hr}$,
and minimize the squared loss with GD:
\begin{equation} \label{matrix-sensing}
    \min_{\mU \in \R^{d \times \hr}} \quad f(\mU) := \frac{1}{4m}\sum_{i=1}^m \left( y_i - \left\langle \mA_i, \mU \mU^{\top}\right\rangle\right)^2.
\end{equation}
In the ideal case, the number of columns of $\mU$, denoted as $\hr$ above, should be set to $r_*$.
However, $r_*$ may not be known in advance.
This leads to two training regimes that are more likely to happen:
the {\em under-parameterized} regime where $\hr \le r_*$,
and the {\em over-parameterized} regime where $\hr > r_*$.

The over-parameterized regime may lead to overfitting at first glance, but
surprisingly, with small initialization, GD induces a good implicit bias towards
solutions with the exact or approximate recovery of the ground truth. It was first
conjectured in \citet{gunasekar2017implicit} that GD with small initialization
finds the matrix with the minimum nuclear norm. \citet{gunasekar2017implicit} also proved this conjecture for a special case where all measurements are commutable. However, a series of works point out
that this nuclear norm minimization view cannot capture the incremental learning behavior of GD,
which, in the context of matrix sensing, refers to the phenomenon that GD tends to learn solutions with
rank gradually increasing with training steps.
\citet{arora2019implicit}
exhibited this phenomenon when there is only one observation ($m=1$).
\citet{gissin2019implicit,Jiang2022AlgorithmicRI} studied the
full-observation case, where every entry of the ground truth is measured
independently $f(\mU) = \frac{1}{4d^2} \| \mZ^* - \mU\mU^\top
\|_{\mathrm{F}}^2$, and GD is shown to sequentially recover singular
components of the ground truth from the largest singular value to the smallest
one. \citet{li2020towards} provided theoretical evidence that the incremental
learning behavior generally occurs for matrix sensing. They specifically provided a
counterexample for \citet{gunasekar2017implicit}'s conjecture, where 
GD converges to a rank-1 solution with a very large nuclear norm.
\citet{razin2020implicit} also pointed out a case where GD
drives the norm to infinity while keeping the rank to be approximately $1$. 

Despite these progresses, theoretical understanding of the simplicity bias
of GD remains limited. In fact, a vast majority of existing analyses can only show
that GD is initially biased towards a rank-$1$ solution
and cannot be generalized to higher ranks, unless additional assumptions on
GD dynamics are made~\citep[Appendix H]{li2020towards}, \citep{belabbas2020implicit,jacot2021deep,razin2021implicit,razin2022implicit}. Recently \citet{li2022implicit} shows that the implicit bias of \citet{gunasekar2017implicit} essentially relies on rewriting gradient flow in the space of $\mU$ as continuous mirror descent in the space of $\mU\mU^\top$, which only works a special type of reparametrized model, named ``commuting parametrization''. However, \citet{li2022implicit} also shows that matrix sensing with general (non-commutable) measurements does not fall into this type.




\subsection{Our Contributions}
In this paper, we take a step towards understanding the generalization of GD with small initialization by
firmly demonstrating the simplicity bias/incremental learning behavior in the matrix sensing setting,
assuming the Restricted Isometry Property (RIP).
Our main result is informally stated below.
See \Cref{main_result} for the formal version.

\begin{definition}[Best Rank-$s$ Solution]
\label{def:best-rank}
   We define the {\em best rank-$s$ solution} as the unique global minimizer $\mZ_{s}^*$ of the following constrained optimization problem:
\begin{equation}
    \label{under-parameterize-sensing}
    \begin{aligned}
        & \min_{\mZ \in \R^{d \times d}}\quad \frac{1}{4m}\sum_{i=1}^m \left( y_i -\left\langle \bm{A}_i,\bm{Z}\right\rangle\right)^2\\
        & s.t.\quad \mZ\succeq \bm{0}, \quad \rank{\mZ}\leq s.
    \end{aligned}
\end{equation}
\end{definition}

\begin{theorem}[Informal version of \Cref{main_result}] \label{main-informal}
Consider the matrix sensing problem \Cref{matrix-sensing} with rank-$r_*$ ground-truth matrix $\mZ^*$ and measurements $\{\mA_i\}_{i=1}^{m}$. Assume that the measurements satisfy the RIP condition (\Cref{def-rip}).
With small learning rate $\mu > 0$ and small initialization $\mU_{\alpha,0} = \alpha\mU \in \R^{d \times \hr}$,
the trajectory of $\mU_{\alpha,t}\mU_{\alpha,t}^\top$ during GD training
enters an $o(1)$-neighborhood of each of the best rank-$s$ solutions in the order of $s = 1, 2, \dots, \hr\wedge r_*$ when $\alpha\to 0$. Moreover, when $\hr \leq r^*$, we have $\lim_{t\to\infty}\mU_{\alpha,t}\mU_{\alpha,t}^\top = \mZ_{\hr}^*$.
\end{theorem}

It is shown in \citet{li2018algorithmic,stoger2021small}
that GD exactly recovers the ground truth under the RIP condition, but our theorem
goes beyond them in a number of ways. First, in the over-parameterized regime (i.e., $\hr >
r_{*}$), it implies that GD performs \textit{incremental
learning}: learning solutions with increasing ranks until it finds
the ground truth. Second, this result also shows that in the under-parameterized
regime (i.e., $\hr \leq r_{*}$), GD exhibits the same implicit bias, but finally it
converges to the best low-rank solution of the matrix sensing loss. By
contrast, to the best of our knowledge, only the over-parameterized setting is analyzed in existing literature. 


\Cref{main-informal} can also be considered as a generalization of previous
results in \citet{gissin2019implicit,Jiang2022AlgorithmicRI} which show that
$\mU_{\alpha,t}\mU_{\alpha,t}^\top$ passes by the best low-rank solutions one by
one in the full observation case of matrix sensing $f(\mU) = \frac{1}{4d^2} \|
\mZ^* - \mU\mU^\top \|_{\mathrm{F}}^2$. However, our setting has two major
challenges which significantly complicate our analysis. First, since our setting
only gives partial measurements, the decomposition of signal and error terms in
\citet{gissin2019implicit,Jiang2022AlgorithmicRI} cannot be applied. Instead, we
adopt a different approach which is motivated by \citet{stoger2021small}.
Second, it is well-known that the optimal rank-$s$ solution of matrix
factorization is $\mX_s$ (defined in \Cref{sec-prelim}), but little is known for
$\mZ_s^*$.  In \Cref{subsec:key-lemma} we analyze the landscape of
\Cref{under-parameterize-sensing}, establishing the uniqueness of $\mZ_s^*$ and
local landscape properties under the RIP condition.
We find that when $\mU_{\alpha,t}\mU_{\alpha,t}^\top\approx\mZ_s^*$, GD follows
an approximate low-rank trajectory, so that it behaves similarly to GD in the
under-parameterized regime. Using our landscape results, we can finally prove
\Cref{main-informal}.


\textbf{Organization.} 
We review additional related works in \Cref{sec-related-work}. 
In \Cref{sec-prelim}, we provide an overview of necessary background and notations. We then present our main results in \Cref{sec-main-result} with proof sketch where we also state some key lemmas that are used in the proof, including \Cref{main-lemma} and some landscape results. In \Cref{sec-proof-sketch} we present a trajectory analysis of GD and prove \Cref{main-lemma}. Experimental results are presented in \Cref{sec-experiment} which verify our theoretical findings. Finally, in \Cref{sec-conclude}, we summarize our main contributions and discuss some promising future directions. Complete proofs of all results are given in the Appendix.

\section{Related work}
\label{sec-related-work}
\textbf{Low-rank matrix recovery.} The goal of low-rank matrix recovery is to
recover an unknown low-rank matrix from a number of (possibly noisy)
measurements. Examples include matrix sensing \citep{recht2010guaranteed},
matrix completion \citep{candes2009exact,candes2010matrix} and robust PCA
\citep{xu2010robust,candes2011robust}.
\citet{fornasier2011low,ngo2012scaled,wei2016guarantees,tong2021accelerating}
study efficient optimization algorithms with convergence guarantees.
Interested readers can refer to
\citet{davenport2016overview} for an overview of this topic.

\noindent\textbf{Simplicity bias/incremental learning of GD.} 
Besides the works mentioned in the introduction,
there are many other works studying the simplicity bias/incremental learning of GD
on tensor factorization~\citep{razin2021implicit,razin2022implicit},
deep linear networks~\citep{gidel2019implicit},
two-layer nets with orthogonal inputs~\citep{boursier2022gradient}.



\noindent\textbf{Landscape analysis of non-convex low-rank problems.} The strict saddle property~\citep{ge2016matrix,ge2015escaping,lee2016gradient} was established for non-convex low-rank problems in a unified framework by \citet{ge2017no}. \citet{tu2016low} proved a local PL property for matrix sensing with exact parameterization (i.e. the rank of parameterization and ground-truth matrix are the same). The optimization geometry of general objective function with Burer-Monteiro type factorization is studied in \citet{zhu2018global,li2019symmetry,zhu2021global}. We provide a comprehensive analysis in this regime for matrix factorization as well as matrix sensing that improves over their results.

\section{Preliminaries}
\label{sec-prelim}
In this section, we first list the notations used in this paper, and then
provide details of our theoretical setup and necessary preliminary results.

\subsection{Notations}
\label{notation}

We write $\min\{a, b\}$ as $a \land b$ for short. For any matrix $\mA$, we use $\normF{\mA}$ to denote the
Frobenius norm of $\mA$, use $\normsm{\mA}$ to denote the
spectral norm $\normtwosm{\mA}$,
and use $\sigma_{\min}(\mA)$ to denote the smallest singular value of $\mA$. We use the following notation for Singular Value Decomposition (SVD):

\begin{definition}[Singular Value Decomposition] \label{def:svd} For any matrix
    $\mA \in \R^{d_1 \times d_2}$ of rank $r$, we use $\mA = \mV_{\mA}
    \mSigma_{\mA} \mW_{\mA}^\top$ to denote a Singular Value Decomposition (SVD) of $\mA$, where $\mV_{\mA}
    \in \R^{d_1 \times r}, \mW_{\mA} \in \R^{d_2 \times r}$ satisfy
    $\mV_{\mA}^\top \mV_{\mA} = \mI, \mW_{\mA}^\top\mW_{\mA} = \mI$, and
    $\mSigma_{\mA} \in \R^{r \times r}$ is diagonal.
\end{definition}

For the matrix sensing problem \Cref{matrix-sensing}, we write the ground-truth
matrix as $\mZ^* = \mX \mX^{\top}$ for some $\mX = \left[ \vv_1,\vv_2,\cdots,
\vv_{r_{*}}\right] \in \R^{d\times r_{*}}$ with orthogonal columns from an orthogonal basis $\{\vv_i:i\in[d]\}$ of $\R^d$. 
We denote
the singular values of $\mX$ as $\sigma_1, \sigma_2, \dots, \sigma_{r_*}$, then
the singular values of $\mZ^*$ are $\sigma_1^2, \sigma_2^2, \dots,
\sigma_{r_*}^2$. We set $\sigma_{r_* + 1} := 0$ for convenience. For simplicity,
we only consider the case where $\mZ^*$ has distinct singular values, i.e.,
$\sigma_1^2 >\sigma_2 ^2 > \cdots >\sigma^2_{r_{*}} >  0$.
We use $\kappa := \frac{\sigma_1^2}{\min_{1\le s \le r_*} \{ \sigma_s^2 -
\sigma_{s+1}^2 \}}$ to quantify the degeneracy of the singular values of
$\mZ^*$. We also use the notation $\mX_s = \left[ \vv_1, \vv_2,\cdots,
\vv_s\right]$ for the matrix consisting of the first $s$ columns of $\mX$ and $\mX_s^{\perp} = \left[\vv_{s+1},\cdots,\vv_d \right]$. Following \Cref{def:svd}, we let $\mV_{\mX_s^\perp}=\left[ \frac{\vv_{s+1}}{\|\vv_{s+1}\|},\cdots,\frac{\vv_{d}}{\|\vv_{d}\|} \right]$. Note that the best rank-$s$ solution 
 $\mZ_s^*$ (\Cref{def:best-rank}) does not equal $\mX_s\mX_s^\top$ in general.

We write the results of the measurements $\{\mA_i\}_{i=1}^{m}$
as a linear mapping $\A:\R^{d\times d}\mapsto\R^m$, where $[\A(\mZ)]_{i} =
\frac{1}{\sqrt{m}}\langle \mA_i,\mZ\rangle$ for all $1 \le i \le m$. We use $\A^*: \R^m\to \R^{d\times d}, \A^*(\vw) =
\frac{1}{\sqrt{m}}\sum_{i=1}^m w_i \mA_i$ to denote the adjoint operator of $\A$. Our loss function
\Cref{matrix-sensing} can then be written as $f(\mU) = \frac{1}{4}
\left\|\A\left(\mZ^*-\mU\mU^{\top}\right)\right\|_2^2$.
The gradient is given by 
$\nabla f(\mU) = \A^*\left(\vy-\A(\mU\mU^{\top})\right) \mU =
\A^*\A\left(\mX\mX^{\top}-\mU\mU^{\top}\right) \mU$.

In this paper, we consider GD with learning rate $\mu > 0$ starting from $\mU_0$.
The update rule is
\begin{equation}
\label{GD}
    \mU_{t+1} =\mU_{t}-\mu \nabla f\left(\mU_{t}\right) = (\mI+\mu\mM_t)\mU_t,
\end{equation}
where $\mM_t := \mathcal{A}^{*} \mathcal{A}\left(\mX \mX^{T}-\mU_{t} \mU_{t}^{T}\right)$. 
We specifically focus on GD with \textit{small initialization}: letting $\mU_0 = \alpha\bmU$
for some matrix $\bmU\in\R^{d\times \hr}$ with $\normsm{\bmU}=1$, we are
interested in the trajectory of GD when $\alpha\to 0$. Sometimes we write
$\mU_t$ as $\mU_{\alpha, t}$ to highlight the dependence of the trajectory on
$\alpha$.

\subsection{Assumptions}
For our theoretical analysis of the matrix sensing problem,
we make the following standard assumption in the matrix sensing literature:
\begin{definition}[Restricted Isometry Property]
\label{def-rip}
We say that a measurement operator $\A$ satisfies the $(\delta,r)$-RIP condition if
$(1-\delta) \normFsm{\mZ}^2 \leq \normtwosm{\A(\mZ)}^2 \leq
(1+\delta)\normFsm{\mZ}^2$ for all matrices $\mZ \in \R^{d\times d}$ with
$\rank{\mZ} \leq r$.
\end{definition}
\begin{assumption}
\label{asmp:rip}
The measurement operator $\A$ satisfies the $(2r_{*}+1,\delta)$-RIP property, where $r_* = \rank{\mZ^*}$
and $\delta \le 10^{-12} \kappa^{-4.5} r_*^{-1}$. 
\end{assumption}
The RIP condition is the key to ensure the ground truth to be recoverable with partial observations.
An important consequence of RIP is that it guarantees $\A^*\A(\mZ)=\frac{1}{m}\sum_{i=1}^m \left\langle \mA_i,\mZ\right\rangle \mA_i \approx \mZ$ when $\mZ$ is low-rank. This is made rigorous in the following proposition.
\begin{proposition}
\label{rip-prop}
(\citealp[Lemma 7.3]{stoger2021small}) Suppose that $\A$ satisfies
$(r,\delta)$-RIP with $r\geq 2$, then for all symmetric $\mZ$, we have
$\left\|(\A^*\A-\mI)\mZ\right\|_2 \leq \delta\|\mZ\|_{*}$, where $\|\cdot\|_{*}$
is the nuclear norm. Moreover, if $\rank{\mZ} \leq r-1$, then
$\left\|(\A^*\A-\mI)\mZ\right\|_2 \leq \sqrt{r}\delta\|\mZ\|$.
\end{proposition}

We need the following regularity condition on initialization.
\begin{assumption}
\label{asmp-full-rank}
    For all $1 \le s \le \hr \land r_*$,
    $\sigma_{\min}\left(\mV_{\mX_s}^{\top} \bmU\right) \geq \rho$ for some constant $\rho > 0$,
    where $\mV_{\mX_s}$ is defined as \Cref{def:svd}.
\end{assumption}


The following proposition implies that \Cref{asmp-full-rank} is satisfied with high probability with a Gaussian initialization.

\begin{restatable}[]{proposition}{randomInit}
    \label{prop:random-init}
    Suppose that all entries of $\bar{\mU}\in\R^{d\times \hr}$ are independently drawn
    from $\mathcal{N}\left(0,\frac{1}{\hr}\right)$ and
    $\rho=\eps\frac{\sqrt{\hr}-\sqrt{\hr\wedge r_*-1}}{\sqrt{\hr}} \geq \frac{\eps}{2r_*}$, then
    $\sigma_{\min}\left(\mV_{\mX_s}^{\top}\bar{\mU}\right)\geq \rho$
    holds for all $1\leq s\leq \hr \land r_*$ with probability at least
    $1-\hr\left(C\eps+e^{-c\hr}\right)$, where $c,C>0$ are universal constants.
\end{restatable}

Lastly, we make the following assumption on the step size.

\begin{assumption}
    \label{asmp:step-size}
    The step size $\mu \leq 10^{-4}\delta\|\mX\|^{-2}$.
\end{assumption}

\subsection{Procrustes Distance} \label{sec:prelim-pro}
Our analysis 
uses the notion of Procrustes distance defined as in
\citet{goodall1991procrustes,tu2016low}.
\begin{definition}[Procrustes Distance]
\label{def-procrutes}
The Procrustes distance between two matrices
$\mU_1, \mU_2 \in \R^{d \times s}$
($d,s > 0$)
is defined as the optimal value of the classic {\em orthogonal Procrustes problem}:
\begin{equation}
    \mathrm{dist}(\mU_1,\mU_2) = \min_{\mR \in \R^{s \times s}: \mR^\top \mR = \mI} \left\|\mU_1-\mU_2\mR\right\|_F.
\end{equation}
\end{definition}
We note that the Procrustes distance is well-defined because the set of $s\times
s$ orthogonal matrices is compact and thus the continuous function
$\left\|\mU_1-\mU_2\mR\right\|_F$ in $\mR$ can attain its minimum. The Procrustes distance is a pseudometric, \emph{i.e.}, it is
symmetric and satisfies the triangle inequality. 

The following lemma is borrowed from \citet{tu2016low},
which connects the Procrustes distance between $\mU_1$ and $\mU_2$ with
the distance between $\mU_1\mU_1^\top$ and $\mU_2 \mU_2^\top$.
\begin{lemma}[\citealt{tu2016low}, Lemma 5.4]
\label{procrutes-property}
    For any two matrices $\mU_1,\mU_2\in\R^{d\times r}$,
    we have $\left\|\mU_1\mU_1^\top-\mU_2\mU_2^\top\right\|_F \geq (2\sqrt{2}-2)^{1/2} \cdot \sigma_r(\mU_1)\cdot\mathrm{dist}(\mU_1,\mU_2)$.
\end{lemma}

\section{Main results}
\label{sec-main-result}
In this section, we present our main theorems and their proof sketches,
following the theoretical setup in \Cref{sec-prelim}. Full proofs can be found
in \Cref{appsec:thm-fit,appsec-aux,appsec-landscape,appsec-main-proof}.

\begin{restatable}[]{theorem}{mainThm} \label{main_result}
Under \Cref{asmp:rip,asmp-full-rank,asmp:step-size},
consider GD~\Cref{GD}
with initialization $\mU_{\alpha, 0} = \alpha \bmU$
for solving the matrix sensing problem~\Cref{matrix-sensing}. 
There exist universal constants $c, M$, constant $C = C(\mX,\bar{\mU})$ 
and a sequence of time points $T_{\alpha}^1<T_{\alpha}^2<\cdots<T_{\alpha}^{\hr\wedge r_*}$ 
such that
for all $1 \le s \le \hr \land r_*$, 
the following holds when $\alpha$ is sufficiently small:
\begin{equation}
    \normF{\mU_{\alpha, T_{\alpha}^{s}} \mU_{\alpha, T_{\alpha}^{s}}^\top - \mZ_s^*} \le C \alpha^\frac{1}{M\kappa^2},
\end{equation}
where we recall that $\mZ_s^*$ is the best rank-$s$ solution defined in \Cref{def:best-rank}.
Moreover, GD follows an incremental learning procedure: we have $\lim_{\alpha\to 0}\max_{1\leq t\leq T_{\alpha}^s}\sigma_{s+1}(\mU_{\alpha,t})=0$ for all $1\leq s\leq \hr\wedge r_*$, where $\sigma_{i}(\mA)$ denotes the $i$-th largest singular value of a matrix $\mA$.

\end{restatable}

It is guaranteed that $\mZ_s^*$ is unique for all $1\leq s\leq \hr\wedge r_*$
under our assumptions (see \Cref{landscape-main-global-unique}). In short,
\Cref{main_result} states that GD with small initialization discovers the best
rank-$s$ solution ($s=1,2,\cdots,\hr\wedge r_*$) sequentially. In particular,
when $s=r_*$, the best rank-$s$ solution is exactly the ground truth
$\mX\mX^\top$. Hence with over-parameterization ($\hr\geq r_*$), GD can discover
the ground truth. 

At a high level, our
result characterizes the complete learning dynamics of GD and reveals an
incremental learning mechanism, \emph{i.e.}, GD starts from learning simple
solutions and then gradually increases the complexity of search space until it
finds the ground truth. 


In the under-parameterized setting, we can further establish the following convergence result:

\begin{restatable}[Convergence in the under-parameterized regime]{theorem}{mainThmUnderParam}
\label{main-thm-cor}
    Suppose that $\hr\leq r^*$, then there exists a constant $\bar{\alpha}>0$
    such that when $\alpha<\bar{\alpha}$, we have $\lim_{t\to
    +\infty}\mU_{\alpha,t}\mU_{\alpha,t}^{\top}=\mZ_{\hr}^*$.
\end{restatable}

\subsection{Key lemmas}
\label{subsec:key-lemma}

In this section, we present some key lemmas for proving our main results. First, we can show that with small initialization, GD can get into a small neighborhood of $\mZ_s^*$.

\begin{restatable}[]{lemma}{mainLemma} \label{main-lemma}
Under \Cref{asmp:rip,asmp-full-rank}, there exists $\hat{T}_{\alpha}^s>0$ for
all $\alpha>0$ and $1\leq s\leq \hr\wedge r_*$ such that $\lim_{\alpha\to
0}\max_{1\leq t\leq \hat{T}_{\alpha}^s}\sigma_{s+1}(\mU_{\alpha,t})=0$.
Furthermore, it holds that
$\left\|\mU_{\hat{T}_{\alpha}^s}\mU_{\hat{T}_{\alpha}^s}^\top -\mZ_s^*\right\|_F
=\O\left(\kappa^3 r_*\delta\|\mX\|^2\right)$.
\end{restatable}

The full proof can be found in \Cref{appsec:thm-fit}.
Motivated by \citet{stoger2021small},
we consider the following decomposition of $\mU_t$:
\begin{equation}
    \label{ut-decompose}
    \mU_t = \mU_t\mW_t\mW_t^{\top}+\mU_t\mW_{t,\perp}\mW_{t,\perp}^{\top},
\end{equation}
where 
$\mW_t := \mW_{\mV_{\mX_s}^{\top}\mU_t} \in \R^{\hr \times s}$ is the matrix
consisting of the right singular vectors of $\mV_{\mX_s}^{\top}\mU_t$
(\Cref{def:svd}) and $\mW_{t,\perp} \in \R^{\hr \times (\hr-s)}$ is any
orthogonal complement of $\mW_t$, \emph{i.e.}, $\mW_{t}\mW_{t}^\top +
\mW_{t,\perp}\mW_{t,\perp}^\top = \mI$. The dependence of $\mW_{t},\mW_{t,\perp}$
on $s$ is omitted but will be clear from the context.

We will refer to the term $\mU_t\mW_t\mW_t^\top$ as the \textit{parallel component} and
$\mU_t\mW_{t,\perp}\mW_{t,\perp}^\top$ as the \textit{orthogonal component}. The idea is to
show that the parallel component grows quickly until it gets close to the best
rank-$s$ solution at some time $\hat{T}_{\alpha}^s$ (namely $\mU_t\mW_t\mW_t^\top\mU_t^\top \approx \mZ_s^*$ when $t = \hat{T}_{\alpha}^s$). Meanwhile, 
the orthogonal term grows exponentially slower and stays $o(1)$ before $\hat{T}_{\alpha}^s$.
See 
\Cref{sec-proof-sketch}
for a detailed proof
sketch.

\Cref{main-lemma} shows that
$\mU_t \mU_t^\top$ would enter a neighborhood of $\mZ_s^*$ with
\textit{constant} radius. However, there is still a gap between
\Cref{main-lemma} and \Cref{main_result}, since the latter states that $\mU_t
\mU_t^\top$ would actually get $o(1)$-close to $\mZ_s^*$.

To proceed, we define the \textit{under-parameterized} matrix sensing loss $f_s$
for every $1\leq s \leq r_*$:
\begin{equation}
    \label{under-param-sensing}
    f_s(\mU) = \frac{1}{4}\left\|\A(\mZ^*-\mU\mU^{\top})\right\|_2^2,\quad \mU\in\R^{d\times s}.
\end{equation}
While the function we are minimizing is $f$ (defined in \Cref{matrix-sensing}) rather than $f_s$, \Cref{main-lemma} suggests that for $t \leq \hat{T}_{\alpha}^s$, $\mU_t$ is always approximately rank-$s$, so that we use a low-rank approximation for $\mU_{\hat{T}_{\alpha}^s}$ and associate the dynamics locally with the
GD dynamics of $f_s$. We will elaborate on how this is done in \Cref{subsec:proof-outline}.

When $\mathrm{dist}(\mU_1,\mU_2) = 0$, it can be easily shown that $f_s(\mU_1) = f_s(\mU_2)$ since $f_s$ is invariant to orthogonal transformations.
Moreover, we note that the global minimizer of $f_s$ is unique up to orthogonal transformations.
\begin{restatable}{lemma}{mainGlobalUnique}
    \label{landscape-main-global-unique}
    Under \Cref{asmp:rip}, if $\mU_s^* \in \R^{d\times s}$ is a global minimizer of $f_s$,
then the set of global minimizers
$\argmin f_s$ is equal to $\left\{\mU_s^*\mR: \mR\in\R^{s\times s}, \mR^\top \mR = \mI \right\}$.
\end{restatable}
Around the global minimizers, we show that $f_s$
satisfies the Restricted Secant Inequality (RSI) which is useful for optimization analysis.
\begin{restatable}[]{definition}{procrustesProjection}
    \label{def:procrutes-projection}
    For any $\mU \in \R^{d \times s}$, 
    we use $\Pi_s(\mU)$ to denote the set of {\em closest} global minimizers of
    $f_s$ to $\mU$, namely 
    $\Pi_s(\mU) = \argmin\{\|\mU - \mU^*_s\|_F: \mU^*_s \in \argmin f_s \}$.
\end{restatable}
\begin{restatable}[Restricted Secant Inequality]{lemma}{mainLandscape}
    \label{landscape-main-thm}
    Under \Cref{asmp:rip},
    if 
    a matrix $\mU\in\R^{d\times s}$ satisfies 
    $\left\|\mU- \mU_s^* \right\|_F \leq 10^{-2}\kappa^{-1}\|\mX\|$
    for some $\mU_s^* \in \Pi_s(\mU)$,
    then we have
    \begin{equation}
    \left\langle \nabla f_s(\mU), \mU -
    \mU_s^*\right\rangle \geq 0.1\kappa^{-1}\|\mX\|^2
    \| \mU - \mU_s^* \|_F^2.
    \end{equation}
\end{restatable}

\begin{remark}
    In general, 
    a function $g:\R^n\mapsto\R$ satisfies
    the RSI condition if for some $\mu > 0$, $\langle \nabla g(x), x-\pi(x)\rangle \ge \mu \|x-\pi(x)\|^2$ holds for all $x$,
    where $\pi(x)$ is a projection of $x$ onto $\argmin g$.
    This condition 
    can be used to prove linear convergence of GD~\citep{zhang2013gradient},
    but it is weaker than strong convexity and stronger than \PLfull  (PL) condition~\citep{karimi2016linear}.
\end{remark}

We end this subsection with the following lemma which says that all global minimizers of the $f_s$ must be close to $\mX_s$ under the procrustes distance, which is used in the proof sketch of \Cref{main_result} in \Cref{sec-main-result}.

\begin{restatable}[]{lemma}{minimaClose}
\label{lemma:minima-close}
Under \Cref{asmp:rip}, we have $\mathrm{dist}(\mU_s^*,\mX_s) \leq 40\delta\kappa\|\mX\|_F$ for any global minimizer $\mU_s^*$ of $f_s$. Moreover, $\left\|\mZ_s^*-\mX_s\mX_s^\top\right\|_F \leq 160\delta\kappa\sqrt{r_*}\|\mX\|^2$.
\end{restatable}

\begin{restatable}[]{corollary}{minimaCloseCor}
\label{cor:minima-close}
Under \Cref{asmp:rip}, we have $\sigma_{\min}(\mU_s^*) \geq \frac{1}{2} \sigma_{\min}(\mX_s) = \frac{1}{2}\sigma_s \ge \frac{1}{2} \kappa^{-\frac{1}{2}} \|\mX\|$.
\end{restatable}

The full proofs for \Cref{landscape-main-thm,lemma:minima-close,cor:minima-close} can be found in \Cref{appsec-landscape}.



\subsection{Proof outline}
\label{subsec:proof-outline}

Based on the key lemmas, here we provide the outlines of the proofs for our main theorems
and defer the details to \Cref{appsec-main-proof}.
We first prove \Cref{main-thm-cor} which can be directly derived by combining
the lemmas in \Cref{subsec:key-lemma}.

\begin{proof}[Proof Sketch of \Cref{main-thm-cor}]
    For any global minimizer $\mU_{\hr}^*$ of \Cref{matrix-sensing}, we have
    \begin{align*}
        &~~~~~\mathrm{dist}(\mU_{\hr}^*,\mU_{\alpha,\hat{T}_{\alpha}^{\hr}}) \\
        &\le (2\sqrt{2}-2)^{-1/2} \sigma_{\min}^{-1}\left(\mU_{\hr}^*\right)
        \left\| \mU_{\alpha,\hat{T}_{\alpha}^{\hr}}\mU_{\alpha,\hat{T}_{\alpha}^{\hr}}^\top - \mZ_{\hr}^*\right\|_F \\
        &\le \O(\kappa^{\frac{1}{2}}\|\mX\|^{-1}) \cdot \O(\kappa^3 r_*\delta\|\mX\|^2) \\
        &= \O(\kappa^{3.5} r_*\delta\|\mX\|),
    \end{align*}
    where the first inequality is due to \Cref{procrutes-property} 
    and the second one is due to 
    \Cref{cor:minima-close} and \Cref{main-lemma}.

    \Cref{asmp:rip} and \Cref{landscape-main-thm} then imply that $\mU_{\alpha,\hat{T}_{\alpha}^{\hr}}$ lies in the small neighborhood of the set of global minimizers of $f=f_{\hr}$, in which the RSI holds.
    Following a standard non-convex optimization analysis~\citep{karimi2016linear},
    we can show that GD converges linearly to $\argmin f_{\hr}$ (in the Procrustes distance), which yields the
    conclusion.
\end{proof}

Now we turn to prove \Cref{main_result}.
While $f$ is not necessarily local RSI,  we use a low-rank approximation for $\mU_{t}$ and associate the dynamics in this neighborhood with the GD dynamics of $f_s$.


\begin{proof}[Proof sketch of \Cref{main_result}]
    Recall that by \Cref{main-lemma}, $\mU_{\alpha,\hat{T}_{\alpha}^s}$ is
    approximately rank-$s$. So there must exist a matrix $\bar{\mU}_{\alpha,0} \in
    \R^{d\times \hr}$ with $\rank{\bar{\mU}_{\alpha,0}} \leq s$ such that
    \begin{equation}
        \label{eq:proof-main-result-1}
        \bar{\mU}_{\alpha,0}\bar{\mU}_{\alpha,0}^\top - \mU_{\alpha,T_{\alpha}^s}\mU_{\alpha,T_{\alpha}^s}^\top = o(1) \quad \text{as }\alpha\to 0.
    \end{equation}
    Indeed, we can let $\bar{\mU}_{\alpha,t}$
    be the parallel component of $\mU_{\alpha,T_{\alpha}^s}$
    because
    the orthogonal component stays $o(1)$ (see the discussions following
    \Cref{ut-decompose} and \Cref{main:refinement-main-cor} for details).
    

    Let $\left\{\bar{\mU}_{\alpha,t}\right\}_{t\geq 0}$ be the trajectory of GD
    with step size $\mu$, initialized at $\bar{\mU}_{\alpha,0}$.
    Since the gradient of the objective function $f$ is locally  Lipschitz,
    the solution obtained by the two GD trajectories
    $\{\bar{\mU}_{\alpha,t}\}_{t\geq 0}$ and
    $\{\mU_{\alpha,\hat{T}_{\alpha}^s+t}\}_{t\geq 0}$ will remain $o(1)$-close for at least constantly many steps. Indeed we can show that they will keep $o(1)$ close for some $\bar{t}_{\alpha} = \omega(1)$ steps, \emph{i.e.}, 
    for all $t \in [0, \bar{t}_{\alpha}]$,
    \begin{equation}
        \label{main-lemma-approx}
        \bar{\mU}_{\alpha,t}\bar{\mU}_{\alpha,t}^{\top}- \mU_{\alpha,\hat{T}_{\alpha}^s+t}\mU_{\alpha,\hat{T}_{\alpha}^s+t}^{\top} = o(1).
    \end{equation}
    From \Cref{GD} it is evident that GD initialized at $\bar{\mU}_{\alpha,0}$
    actually lives in the space of matrices with rank $\le s$. Indeed we can
  identify its dynamics with another GD on $f_s$ (defined in
    \Cref{under-param-sensing}). Concretely, let $\hat{\mU}_{\alpha,0}
    \in \R^{d\times s}$ be a matrix so that $\hat{\mU}_{\alpha,0}\hat{\mU}_{\alpha,0}^\top
    = \bar{\mU}_{\alpha,0}\bar{\mU}_{\alpha,0}^\top$, and let
    $\{\hat{\mU}_{\alpha,t}\}_{t\geq 0}$ be the trajectory of GD that optimizes $f_s$ with step size $\mu$ starting from $\hat{\mU}_{\alpha,0}$.
    Then we have
    $\hat{\mU}_{\alpha,t}\hat{\mU}_{\alpha,t}^\top =
    \bar{\mU}_{\alpha,t}\bar{\mU}_{\alpha,t}^\top$ for all $t\geq 0$.

    We can now apply our landscape results for $f_s$ to analyze
    the GD trajectory $\{\hat{\mU}_{\alpha,t}\}_{t\geq 0}$. By
    \Cref{eq:proof-main-result-1} and \Cref{main-lemma}, we have
    $\left\|\hat{\mU}_{\alpha,0}\hat{\mU}_{\alpha,0}^{\top} - \mZ_s^*\right\|_F
    = \O(\kappa^3 r_*\delta\|\mX\|^2)$, so using a similar argument as in the
    proof sketch of \Cref{main-thm-cor},
    \Cref{cor:minima-close,procrutes-property} imply that the initialization
    $\hat{\mU}_{\alpha,0}$ is within an $\O\left(\kappa^{3.5}
    r_*\delta\|\mX\|^2\right)$ neighborhood of the set of global minimizers of
    $f_s(\mU)$. From \Cref{asmp:rip,landscape-main-thm} we know that that
    $f_s(\mU)$ satisfies a local RSI condition in this neighborhood, so
    following standard non-convex optimization analysis
    \citep{karimi2016linear}, we can show that
    $\{\hat{\mU}_{\alpha,t}\}_{t\geq 0}$ converges linearly to its
    set of global minimizers in the Procrustes distance. We need to choose a
    time $t$ such that \Cref{main-lemma-approx} remains true while this linear
    convergence process takes place for sufficiently many steps. This is possible since
    $\bar{t}_{\alpha} = \omega(1)$; indeed we can show that there always exists
    some $t=t_{\alpha}^s \leq \bar{t}_{\alpha}$ such that both
    $\left\|\hat{\mU}_{\alpha,t}\hat{\mU}_{\alpha,t}^{\top}-
    \mU_{\alpha,\hat{T}_{\alpha}^s+t}\mU_{\alpha,\hat{T}_{\alpha}^s+t}^{\top}\right\|_F$
    and $\left\|\hat{\mU}_{\alpha,t}-\bm{U}_s^*\right\|_F$ are bounded by
    $\O(\alpha^{\frac{1}{M\kappa^2}})$. Hence
    $\left\|\mU_{\alpha,t}\mU_{\alpha,t}^\top-\mZ_s^*\right\|_F = \O(\alpha^{\frac{1}{M\kappa^2}})$ when
    $t=T_{\alpha}^s:=\hat{T}_{\alpha}^s+t_{\alpha}^s$. 
    
    For $1\leq s < \hr\wedge r_*$ and $t\leq t_{\alpha}^s$, since
    \Cref{main-lemma-approx} holds and $\rank{\hat{\mU}_{\alpha,t}}\leq s$, we
    have 
    \begin{equation}
        \notag
        \max_{1\leq t\leq T_{\alpha}^s}\sigma_{s+1}\left(\mU_{\alpha,t}\right) \to 0
    \end{equation}
    as $\alpha\to 0$.
    Finally, by \Cref{lemma:minima-close,asmp:rip} we have
    $\left\|\mZ_{s+1}^*-\mX_{s+1}\mX_{s+1}^\top\right\|
    = \O(\delta\kappa\sqrt{r_*})
    = \O(\kappa^{-1}\|\mX\|^2)$, so $\sigma_{s+1}(\mZ_{s+1}^*)\gtrsim \sigma_{s+1}^2$. Therefore, $\mU_{\alpha,t}\mU_{\alpha,t}^\top$ cannot be
    close to $\mZ_{s+1}^*$ when $t \leq T_{\alpha}^s$, so we must have
    $T_{\alpha}^{s+1}>T_{\alpha}^s$. This completes the proof of
    \Cref{main_result}. 
\end{proof}



\section{Proof sketch of \Cref{main-lemma}}
\label{sec-proof-sketch}
In this section, we outline the proof sketch of \Cref{main-lemma}. We divide the
GD dynamics intro three phases and characterize the dynamics separately.
Proof details for these three phases can be found in
\Cref{subsec:spectral,subsec:improve,subsec:refine}.

\subsection{The spectral phase}

Starting from a small initialization, GD
initially behaves similarly to power iteration since $\mU_{t+1} = (\mI + \mu \mM_t) \mU_t \approx
(\mI+\mu\mM)\mU_t$,
where $\mM := \A^*\A(\mX\mX^{\top})$ is a symmetric matrix.
Let $\mM=\sum_{k=1}^d \hat{\sigma}_k^2
\hat{\vv}_k\hat{\vv}_k^{\top}$ be the eigendecomposition of $\mM$, where $\hat{\sigma}_1\geq \hat{\sigma}_2 \geq \cdots\geq\hat{\sigma}_d\geq 0$. Using our assumption on $\delta$ (\Cref{asmp:rip}), we can show that $\left|\sigma_i-\hat{\sigma}_i\right|, 1\leq i\leq s$ are sufficiently small so that $\hat{\sigma}_i$'s are positive and well-separated. Then we have
\begin{equation}
    \label{main-spectral-eq}
    \begin{aligned}
        \mU_T \approx (\mI+\mu\mM)^{T}\mU_0
        &= \sum_{i=1}^d (1+\mu\hat{\sigma}_i^2)^{T} \hat{\vv}_i\hat{\vv}_i^{\top}\mU_0 \\
        &\approx \sum_{i=1}^s (1+\mu\hat{\sigma}_i^2)^{T} \hat{\vv}_i\hat{\vv}_i^{\top}\mU_0,
    \end{aligned}
\end{equation}
where the last step holds because $(1 + \mu \hat{\sigma}_s)^T \gg (1 + \mu \hat{\sigma}_{s+1})^T$. In other words, we can expect that there is an exponential separation between the parallel and orthogonal component of $\mU_T$. Formally, we can prove the following property at the end of the spectral phase:

\begin{lemma}[\Cref{spectral-end}, simplified version]
\label{main:spectral-end}
Suppose that \Cref{asmp:rip,asmp-full-rank,asmp:step-size} hold. Then there
exist positive constants $C_i = C_i(\mX,\bar{\mU}), \gamma_i =
\gamma_i(\mX,\bar{\mU}), i=2,3$ independent of $\alpha$ such that
$\gamma_2<\gamma_3$ and the following inequalities hold for $t =
T_{\alpha}^{\sp} = \O\left( \frac{\log\alpha^{-1}}{\log(1+\mu\|\mX\|^2)}
\right)$ when $\alpha$ is sufficiently small:
\begin{subequations}
    \label{main:induction-base}
    \begin{align*}
    \|\mU_t\| \leq \|\mX\|, &\quad \sigma_{\min}\left( \mU_{t}\mW_{t}\right) \geq C_2\cdot \alpha^{\gamma_2}, \\
    \left\|\mU_{t}\mW_{t,\perp}\right\| \leq C_3\cdot\alpha^{\gamma_3},\text{and} &\quad \left\| \mV_{\mX_s,\perp}^{\top} \mV_{\mU_{t}\mW_{t}}\right\| \leq 200\delta.
    \end{align*}
\end{subequations}
\end{lemma}

\subsection{The parallel improvement phase}

For small $\alpha$, we have $\sigma_{\min}\left(\mU_t\mW_t\right)
\gg \left\|\mU_t\mW_{t,\perp}\right\|$ by the end of the spectral phase. When \Cref{main-spectral-eq} no longer holds, we enter 
a new phase which we call the \textit{parallel improvement phase}. In this phase, the ratio
$\frac{\sigma_{\min}\left(\mU_t\mW_t\right)}{\left\|\mU_t\mW_{t,\perp}\right\|}$
grows exponentially in $t$, until the former reaches a constant scale. Formally, let $T_{\alpha,s}^{\tpi}=\min \left\{t \geqslant 0: \sigma_{\min }^2\left(\mV_{\mX_{s}}^{\top} \mU_{\alpha,t+1}\right)>0.3\kappa^{-1}\|\mX\|^2\right\}$,
then we can prove the following lemma via induction.

\begin{restatable}[]{lemma}{inductionLemma}
\label{main:induction-lemma}
Suppose that \Cref{asmp:rip,asmp-full-rank,asmp:step-size} hold and let $c_3 = 10^{4}\kappa r_{*}^{\frac{1}{2}} \delta$. Then for sufficiently small $\alpha$,  the following inequalities hold when $T_{\alpha}^{\sp} \leq t < T_{\alpha,s}^{\tpi}$:

\begin{subequations}
\label{main:induction-eq}
\begin{align}
    &\quad \sigma_{\min }\left(\mV_{\mX_{s}}^{\top} \mU_{t+1}\right)
    \geq \sigma_{\min }\left(\mV_{\mX_{s}}^{\top} \mU_{t+1}\mW_{t}\right) \geq \left(1+0.5 \mu\left(\sigma_{s}^{2}+\sigma_{s+1}^{2}\right)\right) \sigma_{\min }\left(\mV_{\mX_{s}}^{\top} \mU_t\right), \label{main:induction-1} \\
    & \quad \left\|\mU_{t+1} \mW_{t+1, \perp}\right\|  \leq \left(1+\mu \left(0.4 \sigma_s^2+0.6 \sigma_{s+1}^{2}\right)\right)\left\|\mU_{t} \mW_{t,\perp}\right\|,  \label{main:induction-2}\\
    & \quad\left\|\mV_{\mX_s,\perp}^{\top} \mV_{\mU_{t+1}\mW_{t+1}}\right\| \leq c_3,\label{main:induction-3} \\
    &\quad \rank{\mV_{\mX_{s}}^{\top} \mU_{t+1}} = \rank{\mV_{\mX_{s}}^{\top} \mU_{t+1}\mW_{t}} = s.\label{main:induction-4}
\end{align}
\end{subequations}
\end{restatable}

We can immediately deduce from \Cref{main:spectral-end,main:induction-lemma} that the orthogonal term $\|\mU_t\mW_{t,\perp}\|$ remains $o(1)$ by the end of the parallel improvement phase:

\begin{corollary}[\Cref{refinement-start}, simplified version]
\label{main:refinement-start}
Under the conditions of \Cref{main:induction-lemma}, when $\alpha$ is sufficiently small we have $\left\| \mU_{T_{\alpha,s}^{\tpi}} \mW_{T_{\alpha,s}^{\tpi},\perp}\right\| \leq C_5\cdot \alpha^{\frac{1}{4\kappa}}$
for some constant $C_5 = C_5(\mX,\bar{\mU})$.
\end{corollary}

\subsection{The refinement phase}

After $\sigma_{\min}\left(\mU_t\mW_t\right)$ grows to constant scale, we enter the
\textit{refinement phase} for which we show that
$\normF{\mX_{s} \mX_{s}^{\top} - \mU_{t} \mU_{t}^{\top}}$ keeps decreasing
until it is $\O\left(\delta\kappa^{3}r_*\|\mX\|^2\right)$. Formally, let $\tau=\kappa^{-1}\|\mX\|^2$ and $T_{\alpha, s}^{\mathrm{ft}}=T_{\alpha, s}^{\mathrm{pi}}-\frac{\log \left(10^{-2}\|\boldsymbol{X}\|^{-2} \kappa^{-1} c_3^{-1}\right)}{\log \left(1-\frac{1}{2} \mu \tau\right)}> T_{\alpha, s}^{\mathrm{pi}}$ where $c_3$ is defined in \Cref{main:induction-lemma},
then the following lemma holds.

\begin{restatable}[]{lemma}{refineMain}
    \label{main:refinement-main}
    Suppose that $T_{\alpha,s}^{\tpi}\leq t \leq T_{\alpha, s}^{\mathrm{ft}}$ and all the conditions in \Cref{main:induction-lemma} hold, then we have
    \begin{align}
        &\left\| \mV_{\mX_s}^{\top}( \mX\mX^{\top}-\mU_{t+1}\mU_{t+1}^{\top})\right\|_F \nonumber \\
        \leq &\left( 1-\frac{1}{2}\mu\tau\right)\left\| \mV_{\mX_s}^{\top}\left(\mX \mX^{\top}-\mU_{t} \mU_{t}^{\top}\right)\right\|_F  + 20\mu\|\mX\|^4\left( \delta+5c_3\right) + 2000\mu^2\sqrt{r_*}\|\mX\|^6. \nonumber
    \end{align}
    Moreover, it holds that $\left\|\mU_{t+1}\mW_{t+1,\perp}\right\| \leq (1+\sigma_s^2)\|\mU_t\mW_{t,\perp}\|$ and $\left\| \mV_{\mX_s^\perp}\mV_{\mU_t \mW_t}\right\| \leq c_3$.
\end{restatable}

Using \Cref{main:refinement-main}, we arrive at the following result:

\begin{corollary}
\label{main:refinement-main-cor}
    For sufficiently small $\alpha$, at $t=T_{\alpha, s}^{\mathrm{ft}}$ we have $\left\|\boldsymbol{V}_{\boldsymbol{X}_s}^{\top}\left(\boldsymbol{X} \boldsymbol{X}^{\top}-\boldsymbol{U}_{t} \boldsymbol{U}_{t}^{\top}\right)\right\|_F \leq 80 \delta\kappa^3 r_*\|\mX\|^2$ and $\|\mU_t\mW_{t,\perp}\|=o(1)$ ($\alpha\to 0$).
\end{corollary}

\textit{\textbf{Concluding the proof of \Cref{main-lemma}.}} At $t=T_{\alpha, s}^{\mathrm{ft}}$, we have
\begin{subequations}
\label{eq:linear-convergence}
    \begin{align}
    &\quad \left\|\mX_{s} \mX_{s}^{\top}-\mU_{t} \mU_{t}^{\top}\right\|_F \nonumber \\& \leq \left\|\left(\mX_{s} \mX_{s}^{\top}-\mU_{t} \mU_{t}^{\top}\right) \mV_{\mX_{s}} \mV_{\mX_{s}}^{\top}\right\|_F +\left\|\mU_{t} \mU_{t}^{\top} \mV_{\mX_{s}^{\perp}} \mV_{\mX_{s}^{\perp}}^{\top}\right\|_F\nonumber \\
    &\leq \left\|\left(\mX_{s} \mX_{s}^{\top}-\mU_{t} \mU_{t}^{\top}\right) \mV_{\mX_{s}} \mV_{\mX_{s}}^{\top}\right\|_F +\left\|\mV_{\mX_{s}^{\perp}}^{\top} \mU_{t} \mU_{t}^{\top} \mV_{\mX_{s}^{\perp}}\right\|_F\nonumber \\
    &\leq \left\| \mV_{\mX_{s}}^{\top}\left(\mX_{s} \mX_{s}^{\top}-\mU_{t} \mU_{t}^{\top}\right)\right\|_F  + \sqrt{r_*}\left\|\mV_{\mX_{s}^{\perp}}^{\top} \mU_{t} \mW_{t}\right\|^2+\sqrt{d}\left\|\mV_{\mX_{s}^{\perp}}^{\top} \mU_{t} \mW_{t, \perp}\right\|^2\label{eq:linear-convergence-1} \\
    &\leq \left\| \mV_{\mX_{s}}^{\top}\left(\mX_{s} \mX_{s}^{\top}-\mU_{t} \mU_{t}^{\top}\right)\right\|_F \! +\! 9\sqrt{r_*}\|\mX\|^2 \left\| \mV_{\mX_s^\perp}\mV_{\mU_t \mW_t}\right\|^2\! +\! \sqrt{d}\| \mU_t \mW_{t,\perp}\|^2\!\!\label{eq:linear-convergence-2}\\
    &= \O\left( \delta\kappa^3 r_*\|\mX\|^2 + \|\mX\|^2 c_3^2\sqrt{r_*}\right) + o(1)\label{eq:linear-convergence-3} \\
    &= \O\left( \delta\kappa^3 r_*\|\mX\|^2\right),\label{eq:linear-convergence-4}
    \end{align}
\end{subequations}
where \Cref{eq:linear-convergence-1} uses $\|\mA\|_F \leq \sqrt{\rank{\mA}}\|\mA\|$, \Cref{eq:linear-convergence-2} uses $\|\mU_t\|\leq 3\|\mX\|$, \Cref{eq:linear-convergence-3} follows from \Cref{main:refinement-main,main:refinement-main-cor} and the last step follows from $c_3=10^4\kappa\sqrt{r_*}\delta$ and \Cref{asmp:rip}.

By \Cref{lemma:minima-close}, the best rank-$s$ solution is close to the matrix factorization minimizer \emph{i.e.} $\left\|\mZ_s^*-\mX_s\mX_s^\top\right\|_F = \O\left(\delta\kappa\sqrt{r_*}\|\mX\|^2\right)$. We thus obtain that $\normF{\mZ_s^* - \mU_{t} \mU_{t}^{\top}} = \O\left(\delta\kappa^3 r_*\|\mX\|^2\right)$. Finally, since $\rank{\mU_t\mW_t}\leq s$ (recall the decomposition \Cref{ut-decompose}), we have $\sigma_{s+1}(\mU_t) \leq \left\|\mU_t\mW_{t,\perp}\right\| = o(1)$. The conclusion follows.


\section{Experiments}
\label{sec-experiment}
In this section, we perform some numerical experiments to illustrate our theoretical findings.

\begin{figure*}
    \centering
    \begin{subfigure}{0.28\textwidth}
        \centering
        \includegraphics[width=\textwidth]{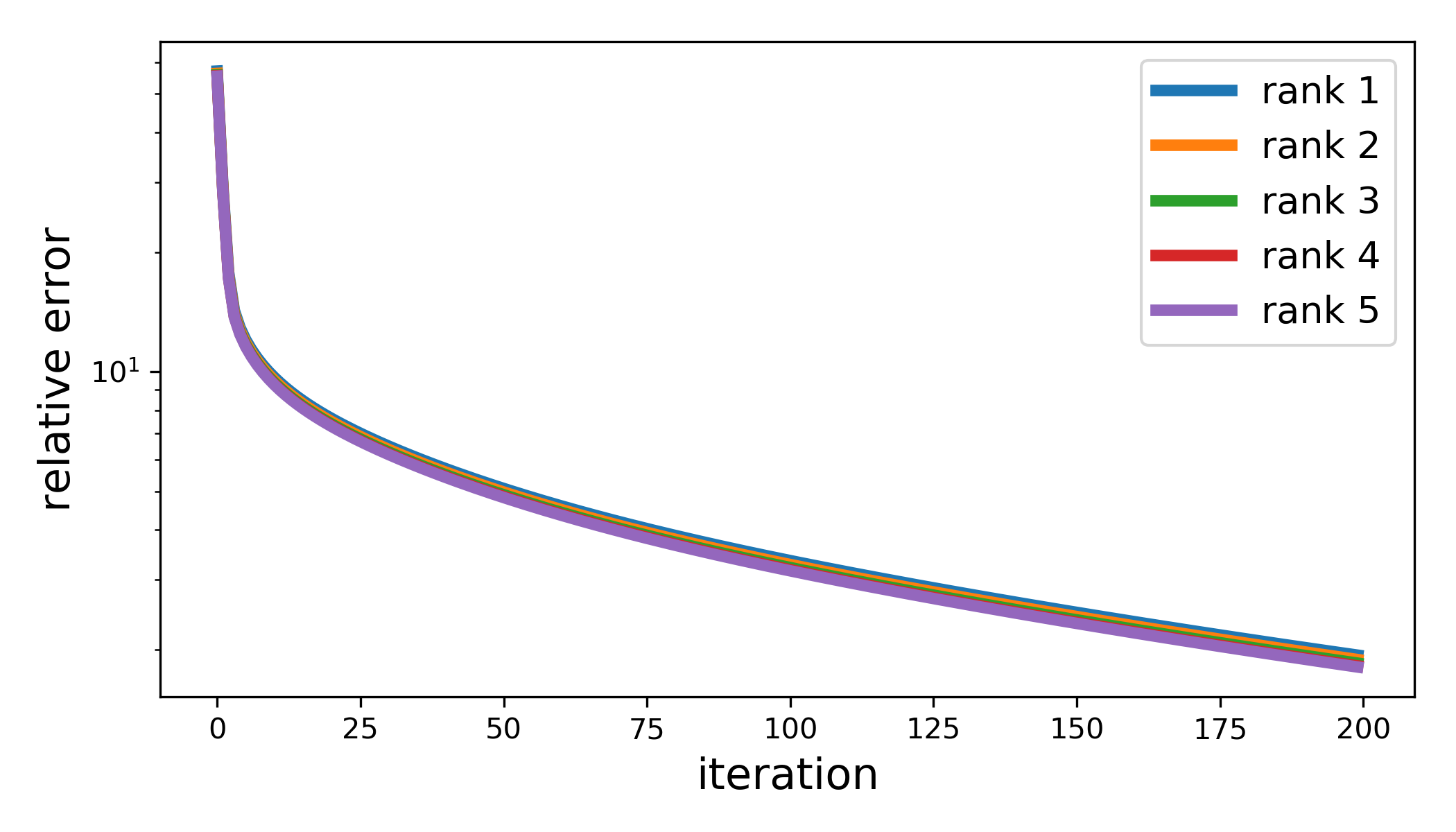}
        \caption[]%
        {{\small $\alpha=1$, $1000$ measurements.}}    
        \label{low_rank0}
    \end{subfigure}
    \qquad
    \begin{subfigure}{0.28\textwidth}  
        \centering 
        \includegraphics[width=\textwidth]{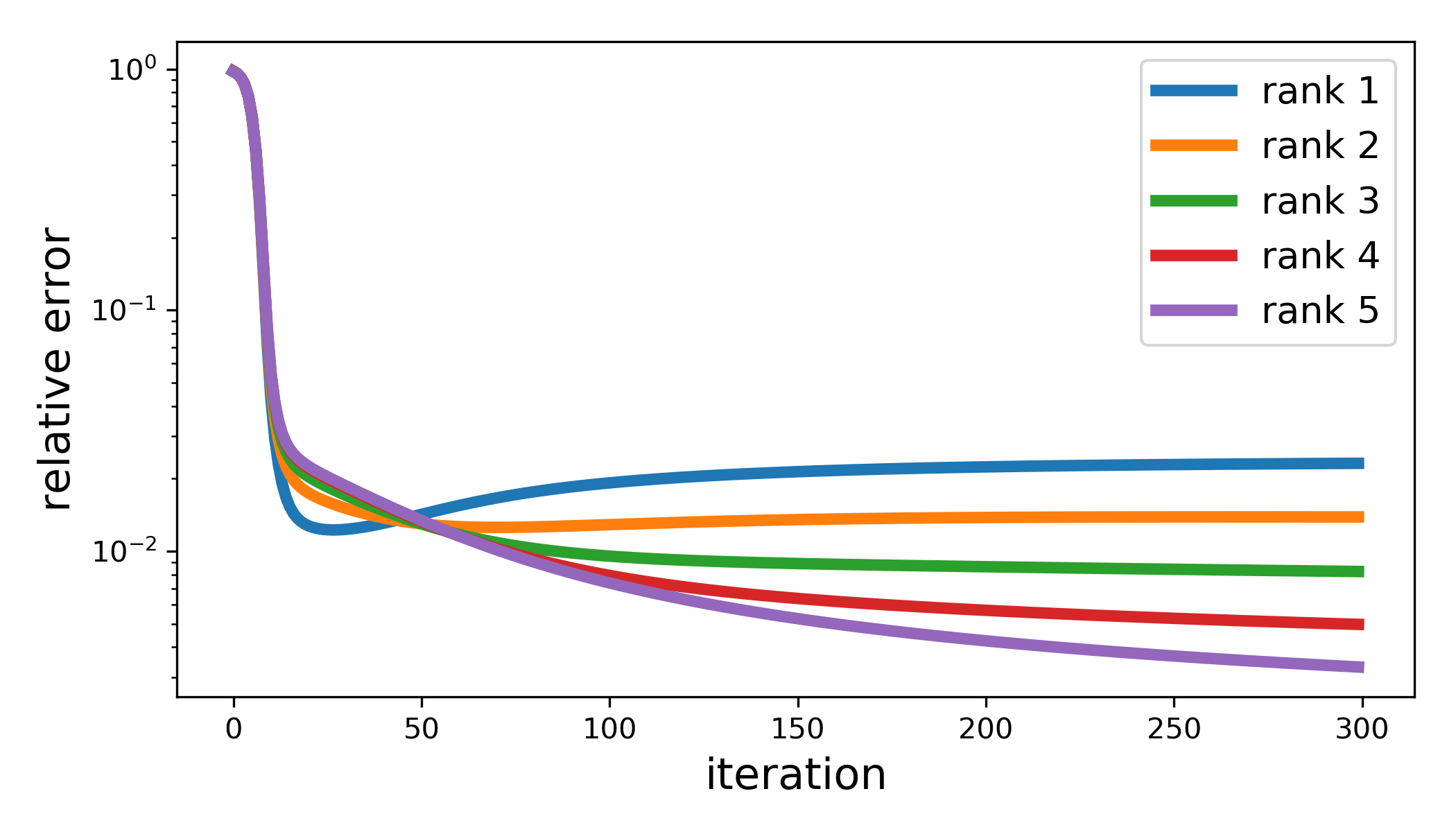}
        \caption[]%
        {{\small $\alpha=0.1$, $1000$ measurements.}}    
        \label{low_rank1}
    \end{subfigure}
    \qquad
    \begin{subfigure}{0.28\textwidth}   
        \centering 
        \includegraphics[width=\textwidth]{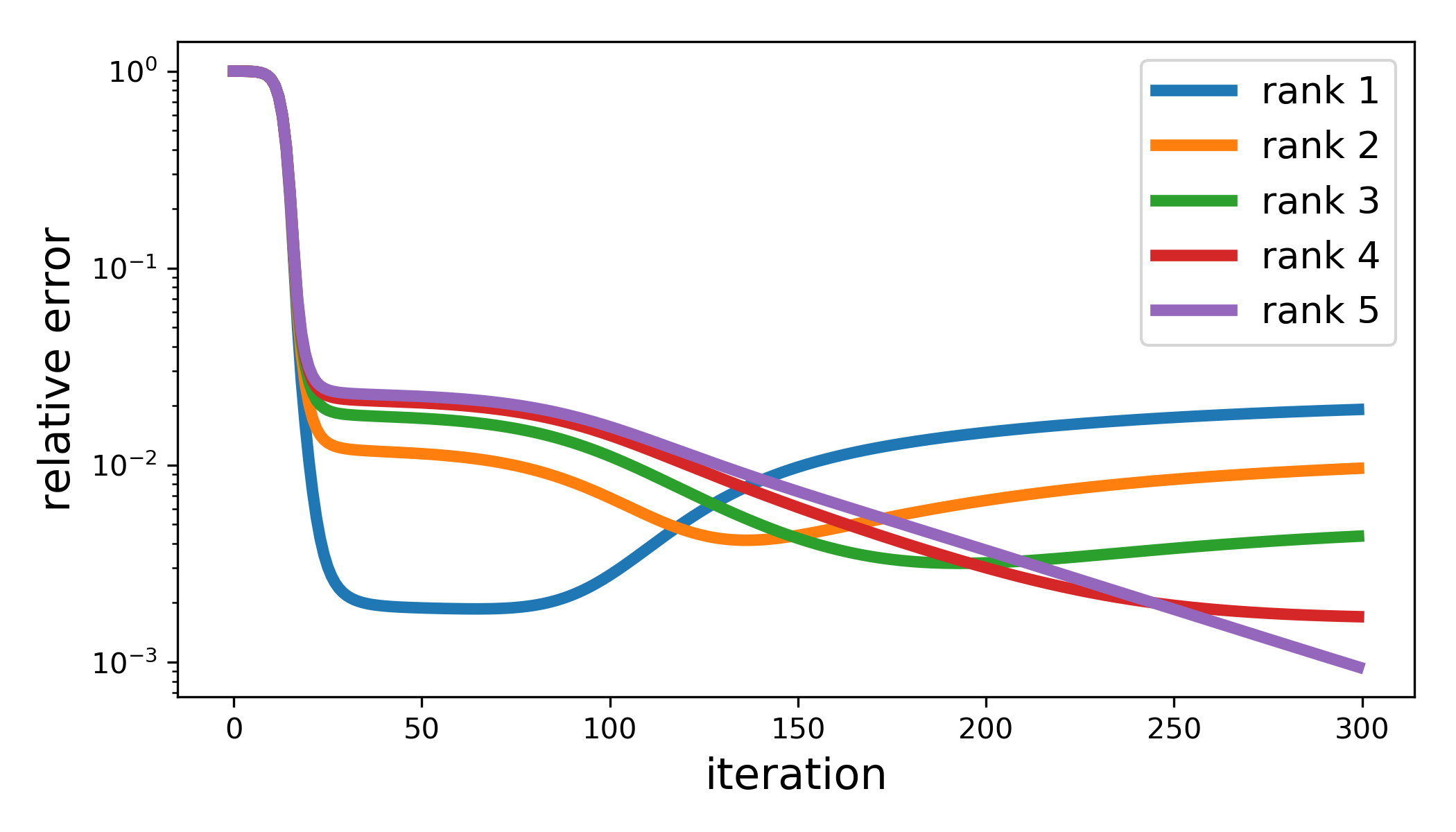}
        \caption[]%
        {{\small $\alpha=0.01$, $1000$ measurements.}}    
        \label{low_rank2}
    \end{subfigure}
    \vskip\baselineskip
    \begin{subfigure}{0.28\textwidth}   
        \centering 
        \includegraphics[width=\textwidth]{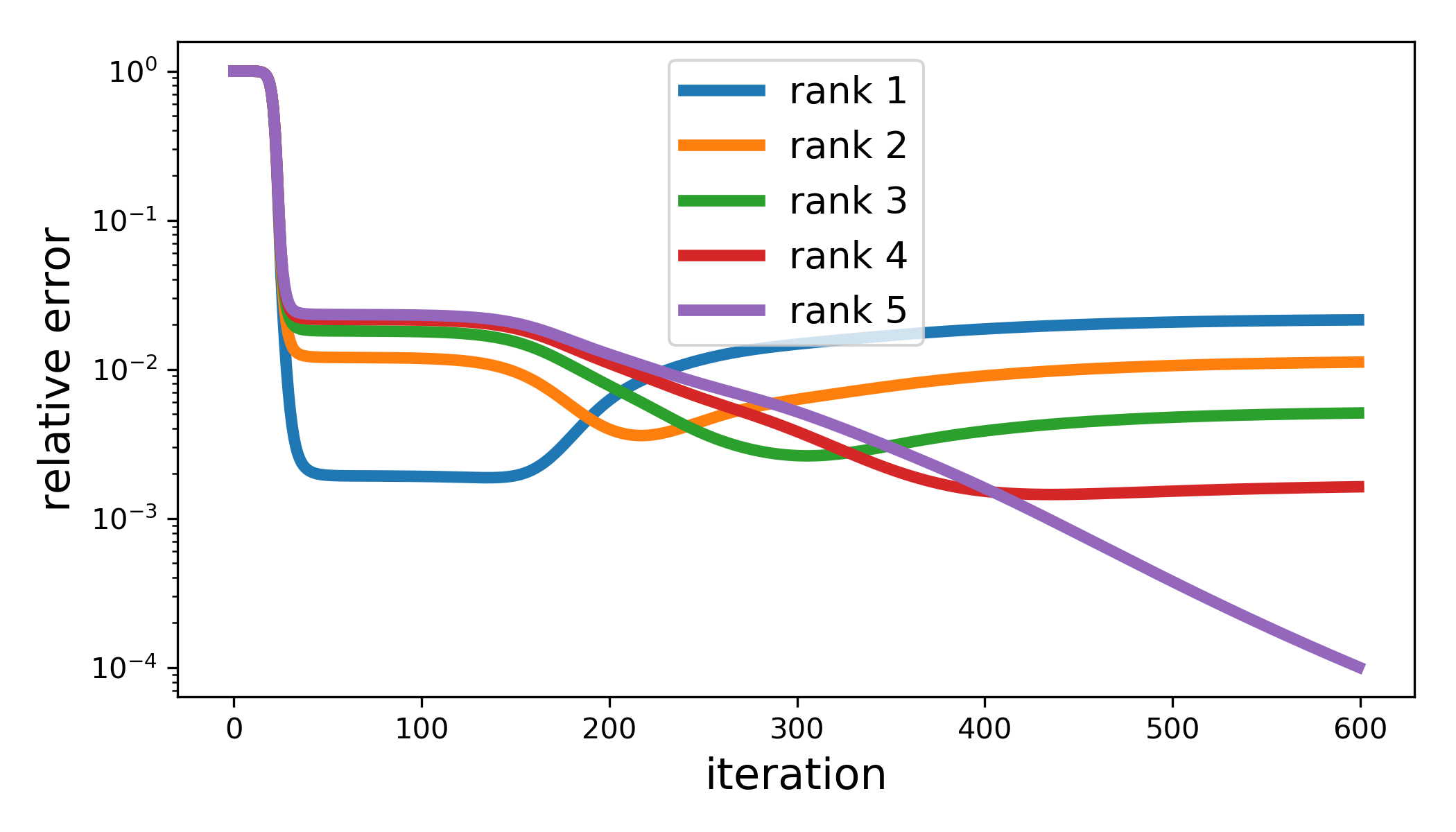}
        \caption[]%
        {{\small $\alpha=0.001$, $1000$ measurements.}}    
        \label{low_rank3}
    \end{subfigure}
    \qquad
    \begin{subfigure}{0.28\textwidth}   
        \centering 
        \includegraphics[width=\textwidth]{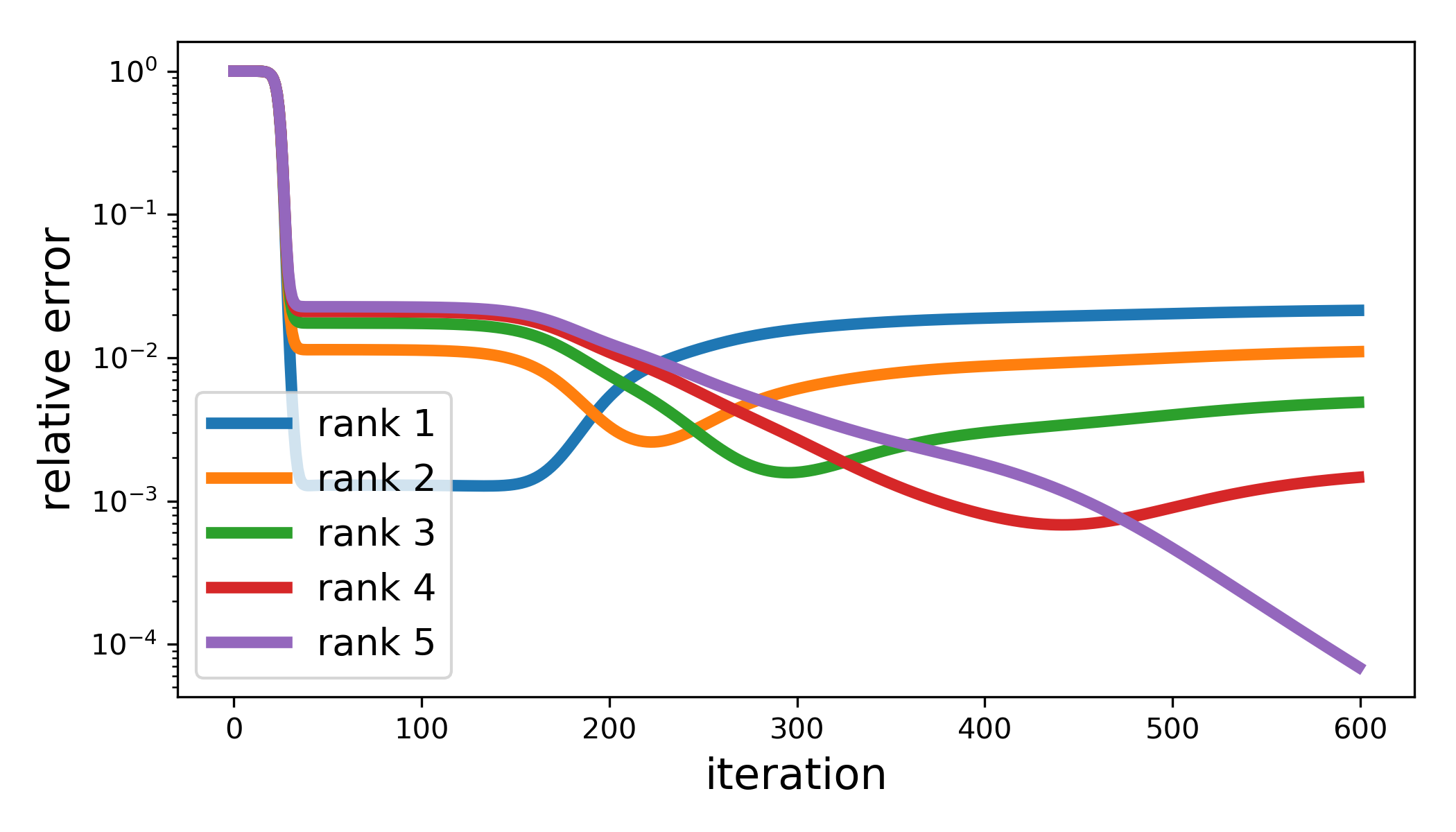}
        \caption[]%
        {{\small $\alpha=0.001$, $2000$ measurements.}}    
        \label{low_rank4}
    \end{subfigure}
    \qquad
    \begin{subfigure}{0.28\textwidth}   
        \centering 
        \includegraphics[width=\textwidth]{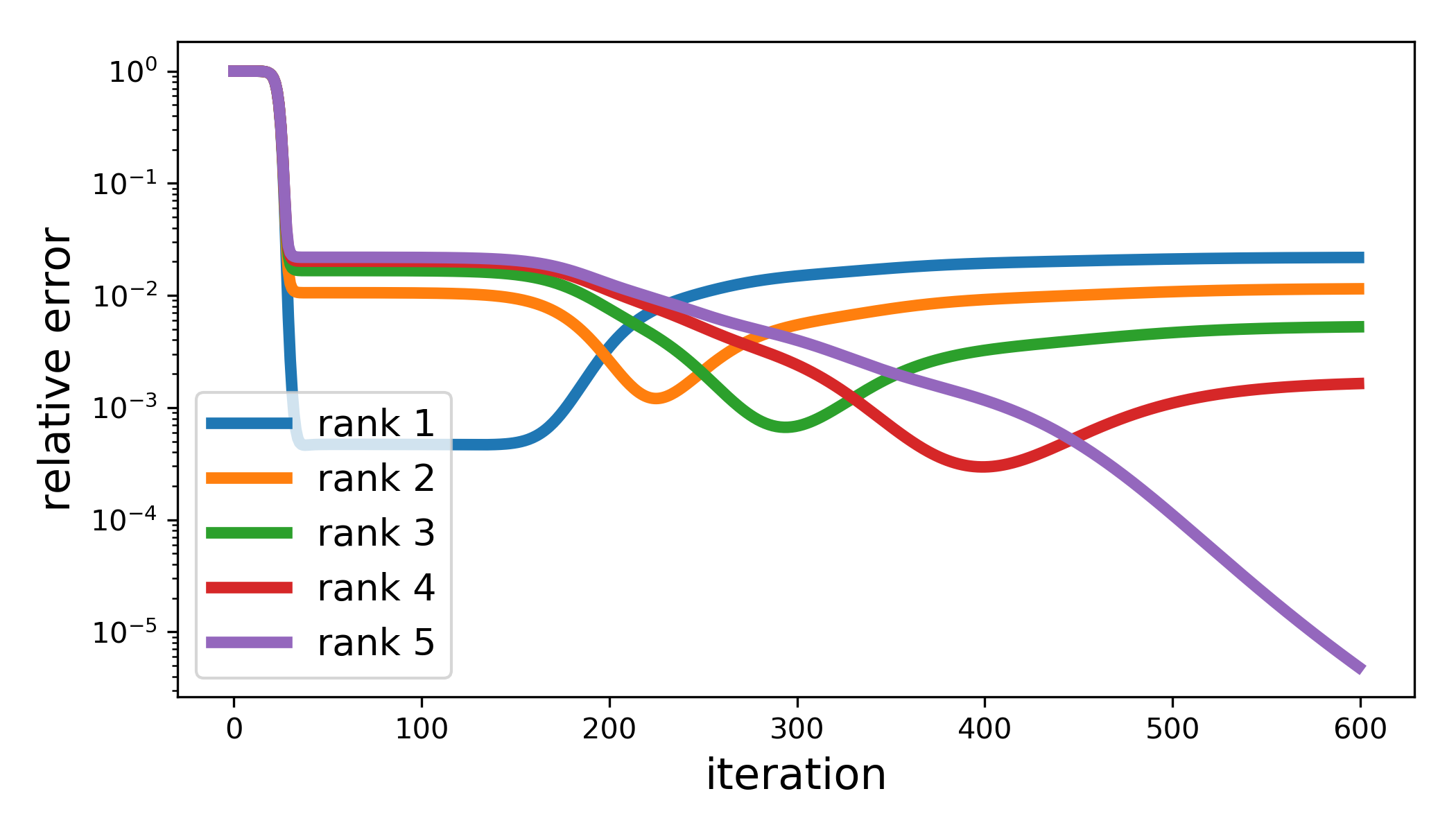}
        \caption[]%
        {{\small $\alpha=0.001$, $5000$ measurements.}}    
        \label{low_rank5}
    \end{subfigure}
    \caption[]%
    {The evolution of relative error against the best solution of different ranks over time.} 
    \label{low_rank}
\end{figure*}

\begin{figure*}
    \centering
    \begin{subfigure}[t]{0.3\textwidth}
        \centering
        \includegraphics[width=\textwidth]{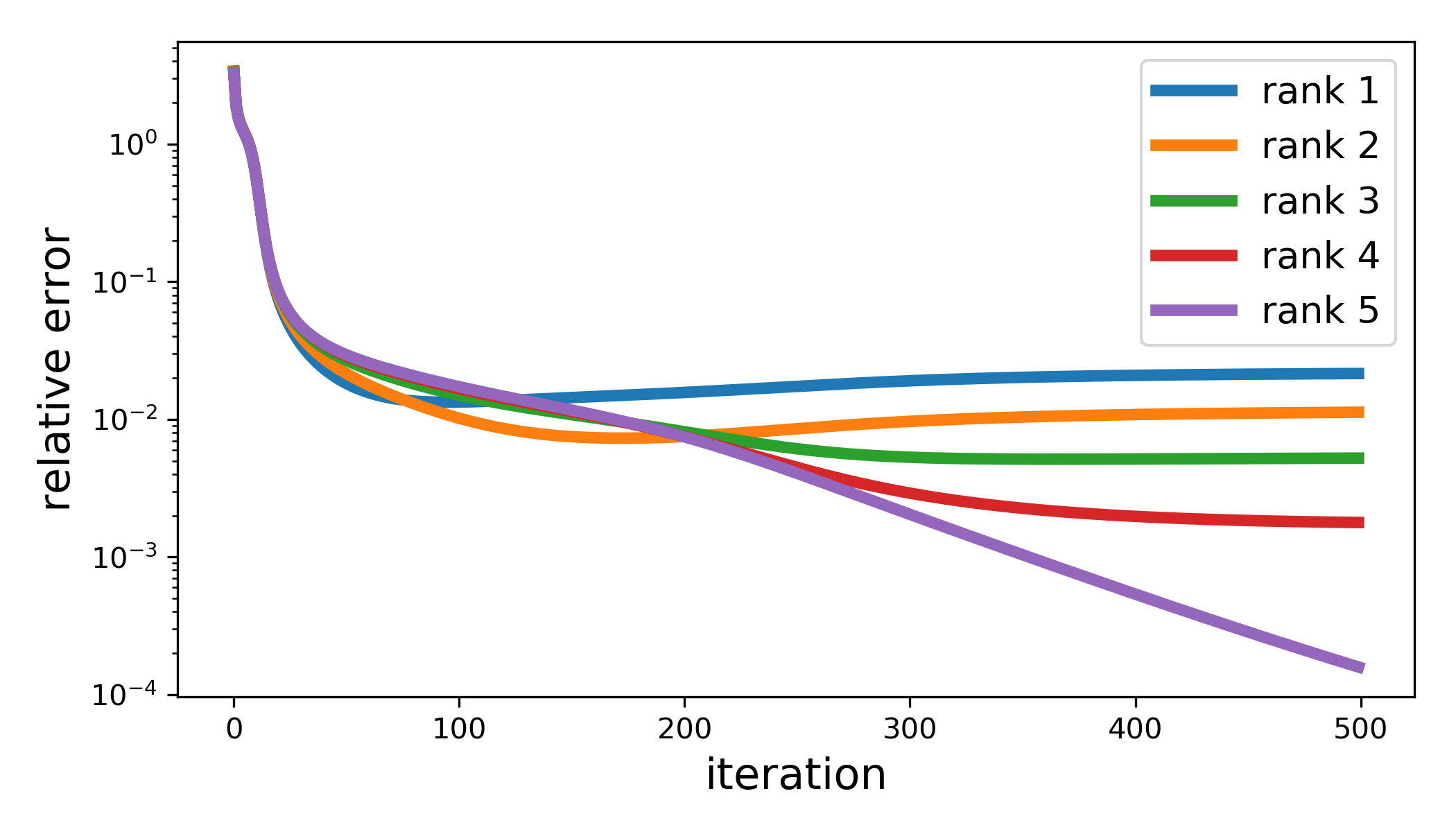}
        \caption[]%
        {{\small $\alpha=1$, $1000$ measurements.}}    
        \label{low_rank_exact0}
    \end{subfigure}
    \qquad
    \begin{subfigure}[t]{0.3\textwidth}  
        \centering 
        \includegraphics[width=\textwidth]{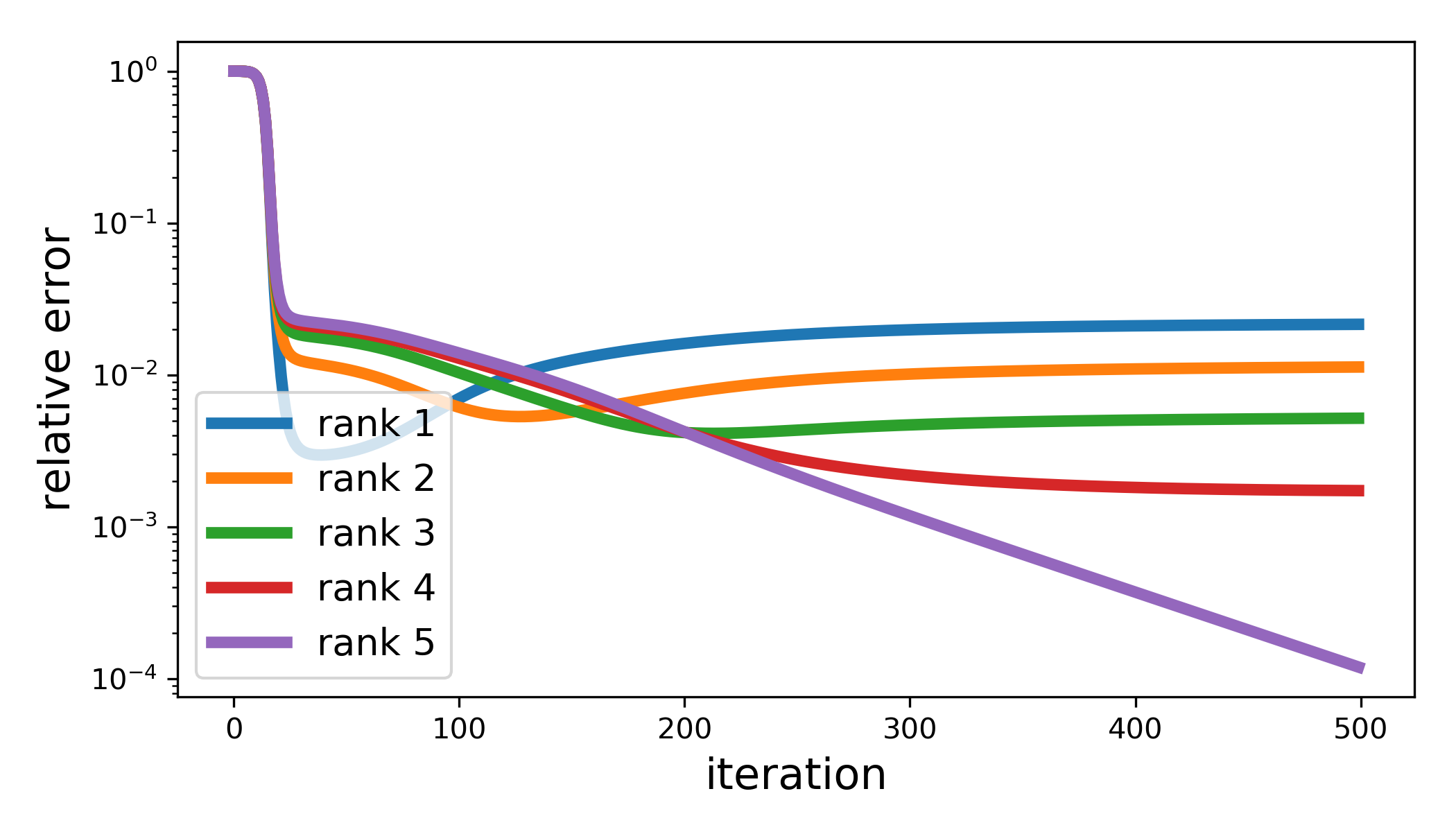}
        \caption[]%
        {{\small $\alpha=0.1$, $1000$ measurements.}}    
        \label{low_rank_exact1}
    \end{subfigure}
    \vskip\baselineskip
    \begin{subfigure}[t]{0.3\textwidth}   
        \centering 
        \includegraphics[width=\textwidth]{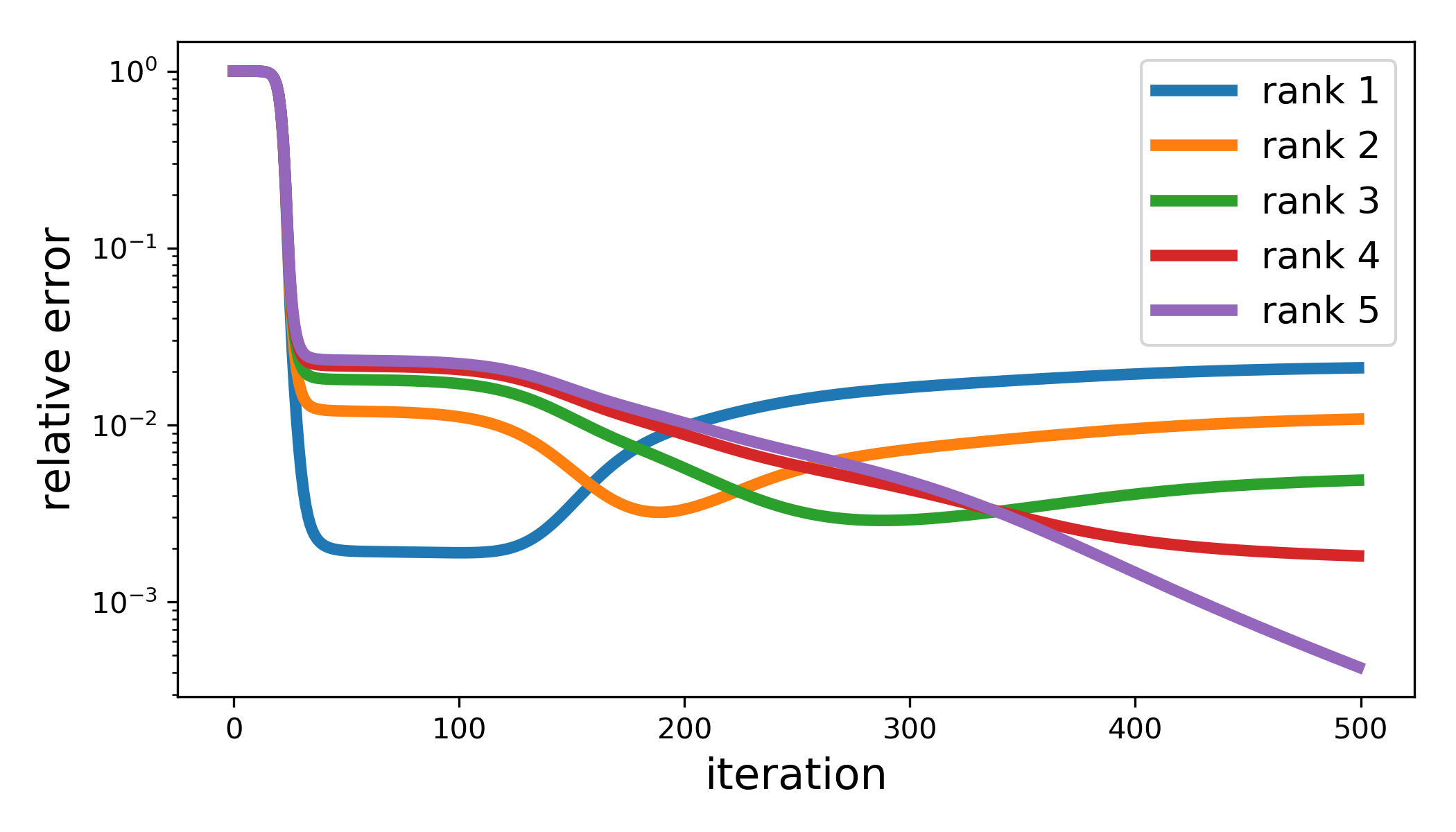}
        \caption[]%
        {{\small $\alpha=0.01$, $1000$ measurements.}}    
        \label{low_rank_exact2}
    \end{subfigure}
    \qquad
    \begin{subfigure}[t]{0.3\textwidth}   
        \centering 
        \includegraphics[width=\textwidth]{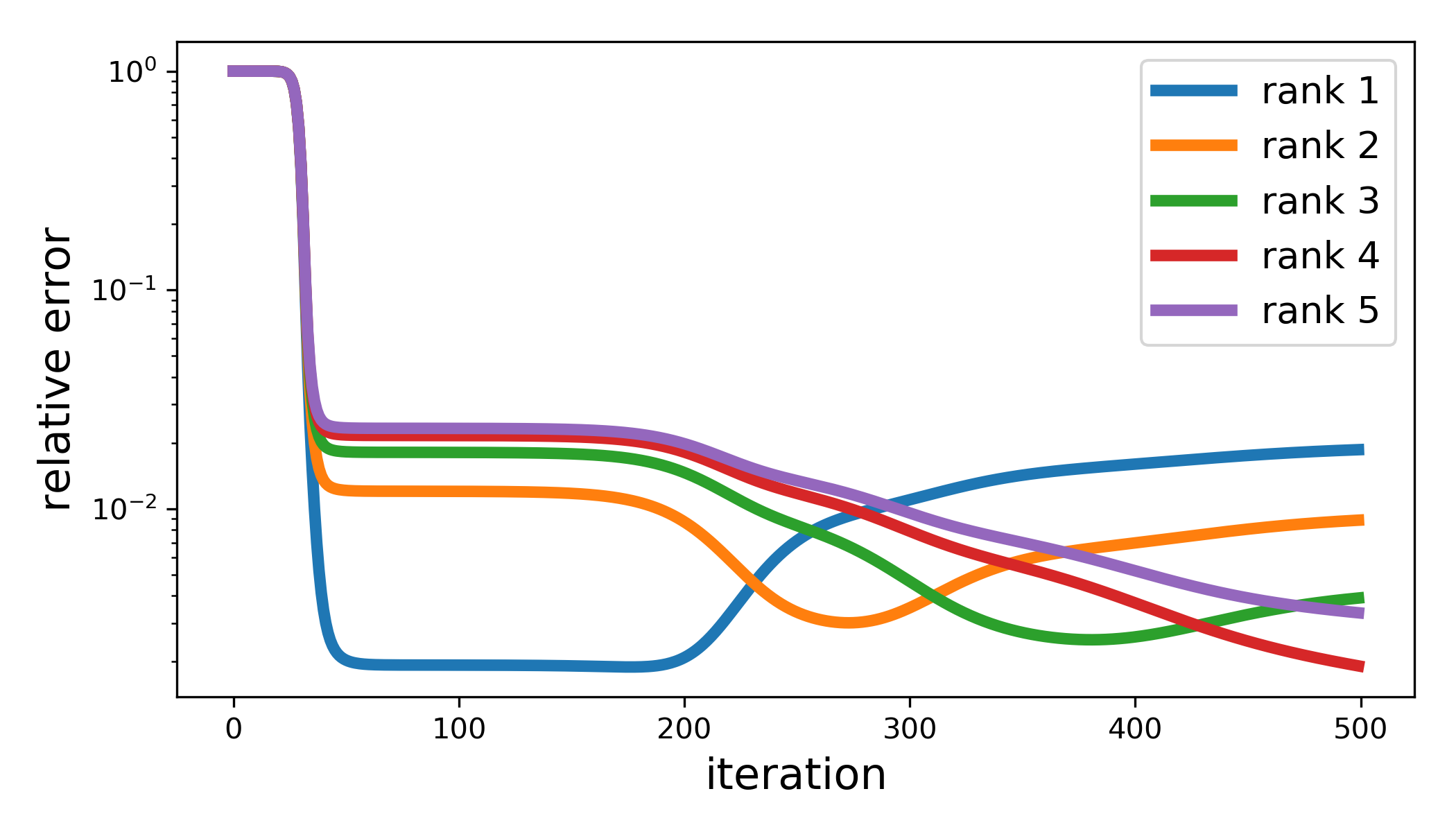}
        \caption[]%
        {{\small $\alpha=0.001$, $1000$ measurements.}}    
        \label{low_rank_exact3}
    \end{subfigure}
    \caption[]%
    {The evolution of the loss and relative error against best solution of different ranks in the exact-parameterized case $r=5$.}
    \label{low_rank_exact}
\end{figure*}

\textbf{Experimental setup.} We consider the matrix
sensing problem \Cref{matrix-sensing} with $d=50$, $r_*=5$,
$\alpha\in\left\{1,0.1,0.01,0.001\right\}$, $m\in\left\{1000,2000,5000\right\}$.
We will consider different choices for $\hr$ in the experiments. The ground truth
$\mZ^* = \mX\mX^{\top}$ is generated such that the entries of $\mX$ are i.i.d.
standard Gaussian variables.
We use the same
ground truth throughout our experiments.

For
$i=1,2,\cdots,m$, all entries of the measurement $\mA_i\in\R^{d\times d}$ are chosen i.i.d. from the standard Gaussian $\mathcal{N}(0,1)$.  When
$m \gtrsim d r_*\delta^{-2}$, this set of measurements satisfies the RIP with
high probability~\citep[Theorem 4.2]{recht2010guaranteed}.

We solve the problem \Cref{matrix-sensing} via running GD for $T=10^4$
iterations starting with small initialization with scale $\alpha$. Specifically,
we choose $\mU_0 = \alpha\bar{\mU}$ where the entries of $\bar{\mU}\in\R^{d\times \hr}$ are
drawn i.i.d.~from $\mathcal{N}(0,1)$. We consider both the
over-parameterized and the exact/under-parameterized regime. The learning rate
of GD is set to be $\mu=0.005$.

\subsection{Implicit low-rank bias}

In this subsection, we consider the over-parameterized setting with $r=50$. 
For each iteration $t \in [T]$ and rank $s \in [r_*]$, we define the relative error $\mathcal{E}_s(t) = \frac{\left\|\mU_t\mU_t^{\top}-\mX_s\mX_s^{\top}\right\|_F^2}{\left\|\mX_s\mX_s^{\top}\right\|_F^2}$ to measure the proximity of the GD iterates to $\mX_s$. We plot the relative error in \Cref{low_rank} for different choices of $\alpha$ and $m$ (which affects the measurement error $\delta$).

\textbf{Small initialization.} The implicit low-rank bias of GD is evident when
the initialization scale $\alpha$ is small. Indeed, one can observe that GD
first visits a small neighborhood of $\mX_1$, spends a long period of time near
it, and then moves towards $\mX_2$. It then proceeds to learn
$\mX_3,\mX_4,\cdots$ in a similar way, until it finally fits the ground truth.
This is in align with \Cref{main_result}. By
contrast, for large initialization we do not have this implicit bias.

\textbf{The effect of measurement error.} For fixed $\alpha$, one can observe
the relative error becomes smaller when the number of measurements increases.
This is in align with \Cref{main-lemma} in which the bound
depends on $\delta$. In particular, for the case $s=r_*$, in the end the distance to the set
of global minima goes to zero as $\alpha \to 0$.

\subsection{Matrix sensing with exact parameterization}

Now we study the behavior of GD in the exact parameterization regime ($r=r_*$).
We fix $m=1000$, $r=r_*=5$ and run GD for $T=500$ iterations. We plot the
relative error in \Cref{low_rank_exact}.
As predicted by \Cref{main_result}, we can observe that when $\alpha$ is small, GD exhibits an implicit low-rank bias and takes a longer time to converge.
The latter is because GD would get into a $\mathrm{poly}(\alpha)$-small neighborhood of the
saddle point $\mZ_s$ and take a long time to escape the saddle. As guaranteed by \Cref{main-thm-cor}, we also observe the final convergence to global minimizers for sufficiently small $\alpha$.

\section{Conclusion}
\label{sec-conclude}

In this paper, we study the matrix sensing problem with RIP measurements and
show that GD with small initialization follows an incremental learning procedure,
where GD finds near-optimal solutions with increasing ranks until it finds
the ground-truth. We take a step towards understanding the optimization and
generalization aspects of simple optimization methods, thereby providing insights
into their success in modern applications such as deep learning
\citep{goodfellow2016deep}. Also, we provide a detailed landscape analysis in
the under-parameterized regime, which to the best of our knowledge is the first
analysis of this kind.

Although we focus on matrix sensing in this paper, it has been revealed in a
line of works that the implicit regularization effect may vary for different
models, including deep matrix factorization~\citep{arora2019implicit} and
nonlinear ReLU/LeakyReLU networks~\citep{lyu2021gradient,timor2022implicit}. Also, it is shown
in~\citet{woodworth2020kernel} that different initialization scales can lead to
distinct inductive bias and affect the generalization and optimization
behaviors. All these results indicate that 
we need further studies to comprehensively understand
gradient-based
optimization methods from the generalization aspect.

\bibliography{Files/iclr2023_conference,Files/simondupub,Files/simonduref}

\begin{thebibliography}{63}
\providecommand{\natexlab}[1]{#1}
\providecommand{\url}[1]{\texttt{#1}}
\expandafter\ifx\csname urlstyle\endcsname\relax
  \providecommand{\doi}[1]{doi: #1}\else
  \providecommand{\doi}{doi: \begingroup \urlstyle{rm}\Url}\fi

\bibitem[Arora et~al.(2019)Arora, Cohen, Hu, and Luo]{arora2019implicit}
Arora, S., Cohen, N., Hu, W., and Luo, Y.
\newblock Implicit regularization in deep matrix factorization.
\newblock \emph{Advances in Neural Information Processing Systems}, 32, 2019.

\bibitem[Belabbas(2020)]{belabbas2020implicit}
Belabbas, M.~A.
\newblock On implicit regularization: Morse functions and applications to
  matrix factorization.
\newblock \emph{arXiv preprint arXiv:2001.04264}, 2020.

\bibitem[Blanc et~al.(2020)Blanc, Gupta, Valiant, and Valiant]{blanc2020ou}
Blanc, G., Gupta, N., Valiant, G., and Valiant, P.
\newblock Implicit regularization for deep neural networks driven by an
  ornstein-uhlenbeck like process.
\newblock In Abernethy, J. and Agarwal, S. (eds.), \emph{Proceedings of Thirty
  Third Conference on Learning Theory}, volume 125 of \emph{Proceedings of
  Machine Learning Research}, pp.\  483--513. PMLR, 09--12 Jul 2020.

\bibitem[Boursier et~al.(2022)Boursier, Pillaud-Vivien, and
  Flammarion]{boursier2022gradient}
Boursier, E., Pillaud-Vivien, L., and Flammarion, N.
\newblock Gradient flow dynamics of shallow relu networks for square loss and
  orthogonal inputs.
\newblock \emph{arXiv preprint arXiv:2206.00939}, 2022.

\bibitem[Candes \& Plan(2010)Candes and Plan]{candes2010matrix}
Candes, E.~J. and Plan, Y.
\newblock Matrix completion with noise.
\newblock \emph{Proceedings of the IEEE}, 98\penalty0 (6):\penalty0 925--936,
  2010.

\bibitem[Cand{\`e}s \& Recht(2009)Cand{\`e}s and Recht]{candes2009exact}
Cand{\`e}s, E.~J. and Recht, B.
\newblock Exact matrix completion via convex optimization.
\newblock \emph{Foundations of Computational mathematics}, 9\penalty0
  (6):\penalty0 717--772, 2009.

\bibitem[Cand{\`e}s et~al.(2011)Cand{\`e}s, Li, Ma, and
  Wright]{candes2011robust}
Cand{\`e}s, E.~J., Li, X., Ma, Y., and Wright, J.
\newblock Robust principal component analysis?
\newblock \emph{Journal of the ACM (JACM)}, 58\penalty0 (3):\penalty0 1--37,
  2011.

\bibitem[Damian et~al.(2021)Damian, Ma, and Lee]{damian2021label}
Damian, A., Ma, T., and Lee, J.~D.
\newblock Label noise {SGD} provably prefers flat global minimizers.
\newblock In Ranzato, M., Beygelzimer, A., Dauphin, Y., Liang, P., and Vaughan,
  J.~W. (eds.), \emph{Advances in Neural Information Processing Systems},
  volume~34, pp.\  27449--27461. Curran Associates, Inc., 2021.

\bibitem[Davenport \& Romberg(2016)Davenport and
  Romberg]{davenport2016overview}
Davenport, M.~A. and Romberg, J.
\newblock An overview of low-rank matrix recovery from incomplete observations.
\newblock \emph{IEEE Journal of Selected Topics in Signal Processing},
  10\penalty0 (4):\penalty0 608--622, 2016.

\bibitem[Du et~al.(2019)Du, Lee, Li, Wang, and Zhai]{du2019gradient}
Du, S., Lee, J., Li, H., Wang, L., and Zhai, X.
\newblock Gradient descent finds global minima of deep neural networks.
\newblock In \emph{International Conference on Machine Learning}, pp.\
  1675--1685. PMLR, 2019.

\bibitem[Fornasier et~al.(2011)Fornasier, Rauhut, and Ward]{fornasier2011low}
Fornasier, M., Rauhut, H., and Ward, R.
\newblock Low-rank matrix recovery via iteratively reweighted least squares
  minimization.
\newblock \emph{SIAM Journal on Optimization}, 21\penalty0 (4):\penalty0
  1614--1640, 2011.

\bibitem[Frei et~al.(2021)Frei, Cao, and Gu]{frei2021provable}
Frei, S., Cao, Y., and Gu, Q.
\newblock Provable generalization of sgd-trained neural networks of any width
  in the presence of adversarial label noise.
\newblock In Meila, M. and Zhang, T. (eds.), \emph{Proceedings of the 38th
  International Conference on Machine Learning}, volume 139 of
  \emph{Proceedings of Machine Learning Research}, pp.\  3427--3438. PMLR,
  18--24 Jul 2021.

\bibitem[Ge et~al.(2015)Ge, Huang, Jin, and Yuan]{ge2015escaping}
Ge, R., Huang, F., Jin, C., and Yuan, Y.
\newblock Escaping from saddle points—online stochastic gradient for tensor
  decomposition.
\newblock In \emph{Conference on learning theory}, pp.\  797--842. PMLR, 2015.

\bibitem[Ge et~al.(2016)Ge, Lee, and Ma]{ge2016matrix}
Ge, R., Lee, J.~D., and Ma, T.
\newblock Matrix completion has no spurious local minimum.
\newblock In \emph{Advances in Neural Information Processing Systems}, pp.\
  2973--2981, 2016.

\bibitem[Ge et~al.(2017)Ge, Jin, and Zheng]{ge2017no}
Ge, R., Jin, C., and Zheng, Y.
\newblock No spurious local minima in nonconvex low rank problems: A unified
  geometric analysis.
\newblock In \emph{International Conference on Machine Learning}, pp.\
  1233--1242. PMLR, 2017.

\bibitem[Gidel et~al.(2019)Gidel, Bach, and Lacoste-Julien]{gidel2019implicit}
Gidel, G., Bach, F., and Lacoste-Julien, S.
\newblock Implicit regularization of discrete gradient dynamics in linear
  neural networks.
\newblock In \emph{Advances in Neural Information Processing Systems},
  volume~32. Curran Associates, Inc., 2019.

\bibitem[Gissin et~al.(2019)Gissin, Shalev-Shwartz, and
  Daniely]{gissin2019implicit}
Gissin, D., Shalev-Shwartz, S., and Daniely, A.
\newblock The implicit bias of depth: How incremental learning drives
  generalization.
\newblock In \emph{International Conference on Learning Representations}, 2019.

\bibitem[Goodall(1991)]{goodall1991procrustes}
Goodall, C.
\newblock Procrustes methods in the statistical analysis of shape.
\newblock \emph{Journal of the Royal Statistical Society. Series B
  (Methodological)}, 53\penalty0 (2):\penalty0 285--339, 1991.
\newblock ISSN 00359246.

\bibitem[Goodfellow et~al.(2016)Goodfellow, Bengio, Courville, and
  Bengio]{goodfellow2016deep}
Goodfellow, I., Bengio, Y., Courville, A., and Bengio, Y.
\newblock \emph{Deep learning}, volume~1.
\newblock MIT Press, 2016.

\bibitem[Gunasekar et~al.(2017)Gunasekar, Woodworth, Bhojanapalli, Neyshabur,
  and Srebro]{gunasekar2017implicit}
Gunasekar, S., Woodworth, B.~E., Bhojanapalli, S., Neyshabur, B., and Srebro,
  N.
\newblock Implicit regularization in matrix factorization.
\newblock \emph{Advances in Neural Information Processing Systems}, 30, 2017.

\bibitem[Gunasekar et~al.(2018)Gunasekar, Lee, Soudry, and
  Srebro]{gunasekar2018characterizing}
Gunasekar, S., Lee, J., Soudry, D., and Srebro, N.
\newblock Characterizing implicit bias in terms of optimization geometry.
\newblock In \emph{International Conference on Machine Learning}, pp.\
  1832--1841. PMLR, 2018.

\bibitem[Hardt et~al.(2016)Hardt, Recht, and Singer]{hardt2016train}
Hardt, M., Recht, B., and Singer, Y.
\newblock Train faster, generalize better: Stability of stochastic gradient
  descent.
\newblock In \emph{International conference on machine learning}, pp.\
  1225--1234. PMLR, 2016.

\bibitem[Hu et~al.(2020)Hu, Xiao, Adlam, and Pennington]{hu2020surprising}
Hu, W., Xiao, L., Adlam, B., and Pennington, J.
\newblock The surprising simplicity of the early-time learning dynamics of
  neural networks.
\newblock In Larochelle, H., Ranzato, M., Hadsell, R., Balcan, M., and Lin, H.
  (eds.), \emph{Advances in Neural Information Processing Systems}, volume~33,
  pp.\  17116--17128. Curran Associates, Inc., 2020.

\bibitem[Jacot et~al.(2021)Jacot, Ged, Gabriel, {\c{S}}im{\c{s}}ek, and
  Hongler]{jacot2021deep}
Jacot, A., Ged, F., Gabriel, F., {\c{S}}im{\c{s}}ek, B., and Hongler, C.
\newblock Deep linear networks dynamics: Low-rank biases induced by
  initialization scale and l2 regularization.
\newblock \emph{arXiv preprint arXiv:2106.15933}, 2021.

\bibitem[Ji \& Telgarsky(2020)Ji and Telgarsky]{ji2020directional}
Ji, Z. and Telgarsky, M.
\newblock Directional convergence and alignment in deep learning.
\newblock In Larochelle, H., Ranzato, M., Hadsell, R., Balcan, M., and Lin, H.
  (eds.), \emph{Advances in Neural Information Processing Systems}, volume~33,
  pp.\  17176--17186. Curran Associates, Inc., 2020.

\bibitem[Jiang et~al.(2022)Jiang, Chen, and Ding]{Jiang2022AlgorithmicRI}
Jiang, L., Chen, Y., and Ding, L.
\newblock Algorithmic regularization in model-free overparametrized asymmetric
  matrix factorization.
\newblock \emph{ArXiv}, abs/2203.02839, 2022.

\bibitem[Kalimeris et~al.(2019)Kalimeris, Kaplun, Nakkiran, Edelman, Yang,
  Barak, and Zhang]{kalimeris2019sgd}
Kalimeris, D., Kaplun, G., Nakkiran, P., Edelman, B., Yang, T., Barak, B., and
  Zhang, H.
\newblock Sgd on neural networks learns functions of increasing complexity.
\newblock In \emph{Advances in Neural Information Processing Systems},
  volume~32. Curran Associates, Inc., 2019.

\bibitem[Kalogerias \& Petropulu(2013)Kalogerias and
  Petropulu]{kalogerias2013matrix}
Kalogerias, D.~S. and Petropulu, A.~P.
\newblock Matrix completion in colocated mimo radar: Recoverability, bounds \&
  theoretical guarantees.
\newblock \emph{IEEE Transactions on Signal Processing}, 62\penalty0
  (2):\penalty0 309--321, 2013.

\bibitem[Karimi et~al.(2016)Karimi, Nutini, and Schmidt]{karimi2016linear}
Karimi, H., Nutini, J., and Schmidt, M.
\newblock Linear convergence of gradient and proximal-gradient methods under
  the polyak-{\l}ojasiewicz condition.
\newblock In \emph{Joint European conference on machine learning and knowledge
  discovery in databases}, pp.\  795--811. Springer, 2016.

\bibitem[Lee et~al.(2016)Lee, Simchowitz, Jordan, and Recht]{lee2016gradient}
Lee, J.~D., Simchowitz, M., Jordan, M.~I., and Recht, B.
\newblock Gradient descent only converges to minimizers.
\newblock In \emph{Conference on Learning Theory}, pp.\  1246--1257, 2016.

\bibitem[Li et~al.(2019)Li, Lu, Arora, Haupt, Liu, Wang, and
  Zhao]{li2019symmetry}
Li, X., Lu, J., Arora, R., Haupt, J., Liu, H., Wang, Z., and Zhao, T.
\newblock Symmetry, saddle points, and global optimization landscape of
  nonconvex matrix factorization.
\newblock \emph{IEEE Transactions on Information Theory}, 65\penalty0
  (6):\penalty0 3489--3514, 2019.

\bibitem[Li et~al.(2018)Li, Ma, and Zhang]{li2018algorithmic}
Li, Y., Ma, T., and Zhang, H.
\newblock Algorithmic regularization in over-parameterized matrix sensing and
  neural networks with quadratic activations.
\newblock In \emph{Conference On Learning Theory}, pp.\  2--47. PMLR, 2018.

\bibitem[Li et~al.(2020)Li, Luo, and Lyu]{li2020towards}
Li, Z., Luo, Y., and Lyu, K.
\newblock Towards resolving the implicit bias of gradient descent for matrix
  factorization: Greedy low-rank learning.
\newblock In \emph{International Conference on Learning Representations}, 2020.

\bibitem[Li et~al.(2022{\natexlab{a}})Li, Wang, and Arora]{li2022what}
Li, Z., Wang, T., and Arora, S.
\newblock What happens after {SGD} reaches zero loss? --a mathematical
  framework.
\newblock In \emph{International Conference on Learning Representations},
  2022{\natexlab{a}}.

\bibitem[Li et~al.(2022{\natexlab{b}})Li, Wang, Lee, and Arora]{li2022implicit}
Li, Z., Wang, T., Lee, J.~D., and Arora, S.
\newblock Implicit bias of gradient descent on reparametrized models: On
  equivalence to mirror descent.
\newblock In Oh, A.~H., Agarwal, A., Belgrave, D., and Cho, K. (eds.),
  \emph{Advances in Neural Information Processing Systems}, 2022{\natexlab{b}}.
\newblock URL \url{https://openreview.net/forum?id=k4KHXS6_zOV}.

\bibitem[Lyu \& Li(2020)Lyu and Li]{lyu2020gradient}
Lyu, K. and Li, J.
\newblock Gradient descent maximizes the margin of homogeneous neural networks.
\newblock In \emph{International Conference on Learning Representations}, 2020.

\bibitem[Lyu et~al.(2021)Lyu, Li, Wang, and Arora]{lyu2021gradient}
Lyu, K., Li, Z., Wang, R., and Arora, S.
\newblock Gradient descent on two-layer nets: Margin maximization and
  simplicity bias.
\newblock \emph{Advances in Neural Information Processing Systems},
  34:\penalty0 12978--12991, 2021.

\bibitem[Lyu et~al.(2022)Lyu, Li, and Arora]{lyu2022understanding}
Lyu, K., Li, Z., and Arora, S.
\newblock Understanding the generalization benefit of normalization layers:
  Sharpness reduction.
\newblock \emph{arXiv preprint arXiv:2206.07085}, 2022.

\bibitem[Nacson et~al.(2019)Nacson, Gunasekar, Lee, Srebro, and
  Soudry]{nacson2019lexicographic}
Nacson, M.~S., Gunasekar, S., Lee, J., Srebro, N., and Soudry, D.
\newblock Lexicographic and depth-sensitive margins in homogeneous and
  non-homogeneous deep models.
\newblock In \emph{International Conference on Machine Learning}, pp.\
  4683--4692. PMLR, 2019.

\bibitem[Ngo \& Saad(2012)Ngo and Saad]{ngo2012scaled}
Ngo, T. and Saad, Y.
\newblock Scaled gradients on grassmann manifolds for matrix completion.
\newblock \emph{Advances in neural information processing systems}, 25, 2012.

\bibitem[Peng et~al.(2014)Peng, Suo, Dai, and Xu]{peng2014reweighted}
Peng, Y., Suo, J., Dai, Q., and Xu, W.
\newblock Reweighted low-rank matrix recovery and its application in image
  restoration.
\newblock \emph{IEEE transactions on cybernetics}, 44\penalty0 (12):\penalty0
  2418--2430, 2014.

\bibitem[Razin \& Cohen(2020)Razin and Cohen]{razin2020implicit}
Razin, N. and Cohen, N.
\newblock Implicit regularization in deep learning may not be explainable by
  norms.
\newblock In Larochelle, H., Ranzato, M., Hadsell, R., Balcan, M., and Lin, H.
  (eds.), \emph{Advances in Neural Information Processing Systems}, volume~33,
  pp.\  21174--21187. Curran Associates, Inc., 2020.

\bibitem[Razin et~al.(2021)Razin, Maman, and Cohen]{razin2021implicit}
Razin, N., Maman, A., and Cohen, N.
\newblock Implicit regularization in tensor factorization.
\newblock In \emph{International Conference on Machine Learning}, pp.\
  8913--8924. PMLR, 2021.

\bibitem[Razin et~al.(2022)Razin, Maman, and Cohen]{razin2022implicit}
Razin, N., Maman, A., and Cohen, N.
\newblock Implicit regularization in hierarchical tensor factorization and deep
  convolutional neural networks.
\newblock \emph{arXiv preprint arXiv:2201.11729}, 2022.

\bibitem[Recht et~al.(2010)Recht, Fazel, and Parrilo]{recht2010guaranteed}
Recht, B., Fazel, M., and Parrilo, P.~A.
\newblock Guaranteed minimum-rank solutions of linear matrix equations via
  nuclear norm minimization.
\newblock \emph{SIAM review}, 52\penalty0 (3):\penalty0 471--501, 2010.

\bibitem[Rudelson \& Vershynin(2009)Rudelson and
  Vershynin]{rudelson2009smallest}
Rudelson, M. and Vershynin, R.
\newblock Smallest singular value of a random rectangular matrix.
\newblock \emph{Communications on Pure and Applied Mathematics: A Journal
  Issued by the Courant Institute of Mathematical Sciences}, 62\penalty0
  (12):\penalty0 1707--1739, 2009.

\bibitem[Shen \& Wu(2012)Shen and Wu]{shen2012unified}
Shen, X. and Wu, Y.
\newblock A unified approach to salient object detection via low rank matrix
  recovery.
\newblock In \emph{2012 IEEE Conference on Computer Vision and Pattern
  Recognition}, pp.\  853--860. IEEE, 2012.

\bibitem[Soudry et~al.(2018)Soudry, Hoffer, Nacson, Gunasekar, and
  Srebro]{soudry2018implicit}
Soudry, D., Hoffer, E., Nacson, M.~S., Gunasekar, S., and Srebro, N.
\newblock The implicit bias of gradient descent on separable data.
\newblock \emph{Journal of Machine Learning Research}, 19\penalty0
  (70):\penalty0 1--57, 2018.

\bibitem[St{\"o}ger \& Soltanolkotabi(2021)St{\"o}ger and
  Soltanolkotabi]{stoger2021small}
St{\"o}ger, D. and Soltanolkotabi, M.
\newblock Small random initialization is akin to spectral learning:
  Optimization and generalization guarantees for overparameterized low-rank
  matrix reconstruction.
\newblock \emph{Advances in Neural Information Processing Systems}, 34, 2021.

\bibitem[Sun(2019)]{sun2019optimization}
Sun, R.
\newblock Optimization for deep learning: theory and algorithms.
\newblock \emph{arXiv preprint arXiv:1912.08957}, 2019.

\bibitem[Timor et~al.(2022)Timor, Vardi, and Shamir]{timor2022implicit}
Timor, N., Vardi, G., and Shamir, O.
\newblock Implicit regularization towards rank minimization in relu networks.
\newblock \emph{arXiv preprint arXiv:2201.12760}, 2022.

\bibitem[Tong et~al.(2021)Tong, Ma, and Chi]{tong2021accelerating}
Tong, T., Ma, C., and Chi, Y.
\newblock Accelerating ill-conditioned low-rank matrix estimation via scaled
  gradient descent.
\newblock \emph{J. Mach. Learn. Res.}, 22:\penalty0 150--1, 2021.

\bibitem[Tu et~al.(2016)Tu, Boczar, Simchowitz, Soltanolkotabi, and
  Recht]{tu2016low}
Tu, S., Boczar, R., Simchowitz, M., Soltanolkotabi, M., and Recht, B.
\newblock Low-rank solutions of linear matrix equations via {P}rocrustes flow.
\newblock In \emph{International Conference on Machine Learning}, pp.\
  964--973. PMLR, 2016.

\bibitem[Wedin(1972)]{wedin1972perturbation}
Wedin, P.-{\AA}.
\newblock Perturbation bounds in connection with singular value decomposition.
\newblock \emph{BIT Numerical Mathematics}, 12\penalty0 (1):\penalty0 99--111,
  1972.

\bibitem[Wei et~al.(2016)Wei, Cai, Chan, and Leung]{wei2016guarantees}
Wei, K., Cai, J.-F., Chan, T.~F., and Leung, S.
\newblock Guarantees of riemannian optimization for low rank matrix recovery.
\newblock \emph{SIAM Journal on Matrix Analysis and Applications}, 37\penalty0
  (3):\penalty0 1198--1222, 2016.

\bibitem[Woodworth et~al.(2020)Woodworth, Gunasekar, Lee, Moroshko, Savarese,
  Golan, Soudry, and Srebro]{woodworth2020kernel}
Woodworth, B., Gunasekar, S., Lee, J.~D., Moroshko, E., Savarese, P., Golan,
  I., Soudry, D., and Srebro, N.
\newblock Kernel and rich regimes in overparametrized models.
\newblock In \emph{Conference on Learning Theory}, pp.\  3635--3673. PMLR,
  2020.

\bibitem[Xu et~al.(2010)Xu, Caramanis, and Sanghavi]{xu2010robust}
Xu, H., Caramanis, C., and Sanghavi, S.
\newblock Robust pca via outlier pursuit.
\newblock \emph{Advances in neural information processing systems}, 23, 2010.

\bibitem[Zhang et~al.(2017)Zhang, Bengio, Hardt, Recht, and
  Vinyals]{zhang2017understanding}
Zhang, C., Bengio, S., Hardt, M., Recht, B., and Vinyals, O.
\newblock Understanding deep learning requires rethinking generalization.
\newblock In \emph{International Conference on Learning Representations}, 2017.
\newblock URL \url{https://openreview.net/forum?id=Sy8gdB9xx}.

\bibitem[Zhang \& Yin(2013)Zhang and Yin]{zhang2013gradient}
Zhang, H. and Yin, W.
\newblock Gradient methods for convex minimization: better rates under weaker
  conditions.
\newblock \emph{arXiv preprint arXiv:1303.4645}, 2013.

\bibitem[Zhao et~al.(2010)Zhao, Haldar, Brinegar, and Liang]{zhao2010low}
Zhao, B., Haldar, J.~P., Brinegar, C., and Liang, Z.-P.
\newblock Low rank matrix recovery for real-time cardiac mri.
\newblock In \emph{2010 ieee international symposium on biomedical imaging:
  From nano to macro}, pp.\  996--999. IEEE, 2010.

\bibitem[Zhu et~al.(2018)Zhu, Li, Tang, and Wakin]{zhu2018global}
Zhu, Z., Li, Q., Tang, G., and Wakin, M.~B.
\newblock Global optimality in low-rank matrix optimization.
\newblock \emph{IEEE Transactions on Signal Processing}, 66\penalty0
  (13):\penalty0 3614--3628, 2018.

\bibitem[Zhu et~al.(2021)Zhu, Li, Tang, and Wakin]{zhu2021global}
Zhu, Z., Li, Q., Tang, G., and Wakin, M.~B.
\newblock The global optimization geometry of low-rank matrix optimization.
\newblock \emph{IEEE Transactions on Information Theory}, 67\penalty0
  (2):\penalty0 1308--1331, 2021.

\bibitem[Zou et~al.(2013)Zou, Kpalma, Liu, and Ronsin]{zou2013segmentation}
Zou, W., Kpalma, K., Liu, Z., and Ronsin, J.
\newblock Segmentation driven low-rank matrix recovery for saliency detection.
\newblock In \emph{24th British machine vision conference (BMVC)}, pp.\  1--13,
  2013.

\end{thebibliography}
\bibliographystyle{icml2023}


\newpage
\appendix
\addcontentsline{toc}{section}{Appendix} 
\part{Appendix} 
\parttoc 

\newpage

The appendix is organized as follows: in \Cref{appsec-prelim} we present a number of results that will be used for later proof. \Cref{appsec-main-idea} sketches the main idea for proving our main results. \Cref{appsec:thm-fit} is devoted to a rigorous proof of \Cref{main-lemma} ,with some auxiliary lemmas proved in \Cref{appsec-aux}. In \Cref{appsec-landscape} we analyze the landscape of low-rank matrix sensing and prove our landscape results in \Cref{subsec:key-lemma}. These results are then used in \Cref{appsec-main-proof} to prove \Cref{main_result,main-thm-cor}. Finally, \Cref{appendix_rank1} studies the landscape of rank-$1$ matrix sensing, which enjoys a strongly convex property, as we mentioned in \Cref{subsec:key-lemma} without proof.

\section{Preliminaries}
\label{appsec-prelim}
In this section, we present some useful results that is needed in subsequent analysis. 

\subsection{The RIP condition and its properties}
\label{rip-properties}
In this subsection, we collect a few useful properties of the RIP condition, which we recall below:

\begin{definition}
We say that the measurement $\A$ satisfies the $(\delta,r)$-RIP condition if for all matrices $\mZ \in \R^{d\times d}$ with $\rank{\mZ} \leq r$, we have
\begin{equation}
    \notag
    (1-\delta)\|\mZ\|_F^2 \leq \left\|\A(\mZ)\right\|_2^2 \leq (1-\delta)\|\mZ\|_F^2.
\end{equation}
\end{definition}

The key intuition behind RIP is that $\A^*\A \approx I$, where $\A^*: \vv\mapsto \frac{1}{\sqrt{m}}\sum_{i=1}^m v_i \mA_i$ is the adjoint of $\A$. This intuition is made rigorous by the following proposition:

\begin{proposition}
\label{rip-prop-1}
(\citealp[Lemma 7.3]{stoger2021small}) Suppose that $\A$ satisfies $(r,\delta)$-RIP with $r\geq 2$, then for all symmetric $\mZ$,
\begin{enumerate}[(1).]
    \item if $\rank{\mZ} \leq r-1$, we have $\left\|(\A^*\A-\mI)\mZ\right\|_2 \leq \sqrt{r}\delta\|\mZ\|$.
    \item $\left\|(\A^*\A-\mI)\mZ\right\|_2 \leq \delta\|\mZ\|_{*}$, where $\|\cdot\|_{*}$ is the nuclear norm.
\end{enumerate}
\end{proposition}

\subsection{Matrix analysis}

The following lemma is a direct corollary of \Cref{rip-prop-1} and will be frequently used in our proof.

\begin{lemma}
\label{rip-cor}
Suppose that the measurement $\A$ satisfies $(\delta,2r_*+1)$-RIP condition, then for all matrices $\mU \in \R^{d\times r}$ such that $\rank{\mU}\leq r_*$, we have
\begin{equation}
    \notag
    \left\| \left(\A^*\A-\mI\right) (\mX\mX^{\top}-\mU \mU^{\top})\right\| \leq \delta\sqrt{r_*}\left( \|\mX\|^2+\|\mU\|^2\right).
\end{equation}
\end{lemma}

In our proof we will frequently make use of the Weyl's inequality for singular values:

\begin{lemma}[Weyl's inequality]
\label{weyl}
Let $\mA,\Delta \in \R^{d\times d}$ be two matrices, then for all $1\leq k\leq d$, we have
\begin{equation}
    \notag
    \left| \sigma_k(\mA)-\sigma_k(\mA+\Delta)\right| \leq \|\Delta\|.
\end{equation}
\end{lemma}

We will also need the Wedin's sin theorem for singular value decomposition:

\begin{lemma}
(\citealp[Section 3]{wedin1972perturbation}) Define $R(\cdot)$ to be the column space of a matrix. Suppose that matrices $\mB = \mA + \mT$, $\mA_1$, $\mB_1$ are the top-$s$ components in the SVD of $\mA$ and $\mB$ respectively, and $\mA_0 = \mA-\mA_1, \mB_0 = \mB-\mB_1$. If $\delta = \sigma_{\min}(\mB_1) - \sigma_{\max}(\mA_0) > 0$, then we have
\begin{equation}
    \notag
    \left\| \sin\Theta\left( R(\mA_1),R(\mB_1)\right)\right\| \leq \frac{\|\mT\|}{\delta}
\end{equation}
where $\Theta(\cdot,\cdot)$ denotes the angle between two subspaces.
\end{lemma}

Equipped with \Cref{rip-cor}, we can have the following characterization of the eigenvalues of $\mM$ (recall that $\mM = \A^*\A(\mX\mX^{\top})$):
\begin{lemma}
\label{M-eigen}
Let $\mM := \A^*\A(\mX\mX^{\top})$ and $\mM=\sum_{k=1}^d \hat{\sigma}_k^2 \hat{\vv}_k\hat{\vv}_k^{\top}$ be the eigen-decomposition of $\mM$. For $1 \leq i\leq d$ we have
\begin{equation}
    \notag
    \left| \sigma_i^2 - \hat{\sigma}_i^2 \right| \leq \delta\|\mX\|^2.
\end{equation}
\end{lemma}

\begin{proof}
By Weyl's inequality we have
\begin{equation}
    \notag
    \left| \sigma_i^2 - \hat{\sigma}_i^2\right| \leq \left\| \mM - \mX\mX^{\top}\right\| \leq \delta\|\mX\|^2
\end{equation}
as desired.
\end{proof}

\subsection{Optimization}

\begin{lemma}
\label{opt-prelim-1}
Suppose that a smooth function $f \in \R^m \mapsto \R$ with minimum value $f^*> -\infty$ satisfies the following conditions with some $\eps > 0$:
\begin{enumerate}[(1).]
    \item $\lim_{\|\vx\|\to +\infty}  f(\vx) = +\infty$.
    \item There exists an open subset $S \subset \R^m$ such that the set $S^*$ of global minima of $f$ is contained in $S$, and for all stationary points $x$ of $f$ in $\R^m - S$, we have $f(\vx)-f^* \geq 2\eps$. Moreover, we also have $f(\vx)-f^* \geq 2\eps$ on $\partial S$. 
\end{enumerate}
Then we have
\begin{equation}
    \notag
    \left\{ \vx\in\R^m : f(\vx)-f^* \leq \eps \right\} \subset S.
\end{equation}
\end{lemma}

\begin{proof}
Let $\vx^*$ be the minimizer of $f$ on $\R^m-S$. By condition (1) we can deduce that $\vx^*$ always exists. Moreover, since any local minimizer of a function defined on a compact set must either be a stationary point or lie on the boundary of its domain, we can see that either $\vx^* \in \partial S$ or $\nabla f(\vx^*)=0$ holds. By condition (2), either cases would imply that $f(\vx^*)-f^*\geq 2\eps$, as desired.
\end{proof}

\begin{lemma}
\label{gd-error}
Let $\{\vx_k\},\{\vy_k\}\subset \R^n$ be two sequences generated by $\vx_{k+1} = \vx_k - \mu\nabla f(\vx_k)$ and $\vy_{k+1} = \vy_k - \mu\nabla f(\vy_k)$. Suppose that $\|\vx_k\| \leq B$ and $\|\vy_k\| \leq B$ for all $k$ and $f$ is $L$-smooth in $\left\{ \vx\in\R^n: \|\vx\|\leq B\right\}$, then we have
\begin{equation}
    \notag
    \left\| \vx_{k}-\vy_{k}\right\| \leq (1+\mu L)^k  \left\| \vx_{0}-\vy_{0}\right\|.
\end{equation}
\end{lemma}

\begin{proof}
The update rule implies that
\begin{equation}
    \notag
    \begin{aligned}
    \left\|\vx_{k+1}-\vy_{k+1}\right\| &= \left\| \vx_k - \vy_k - \mu\nabla f(\vx_k) +\mu\nabla f(\vy_k)\right\| \\
    &\leq \|\vx_k-\vy_k\| + \mu\left\|\nabla f(\vx_k)-f(\vy_k)\right\| \\
    &\leq (1+\mu L)\|\vx_k-\vy_k\|
    \end{aligned}
\end{equation}
which yields the desired inequality.
\end{proof}

\subsection{Proof for Proposition \ref{prop:random-init}}

\randomInit*

\Cref{prop:random-init} immediately follows from the following result:

\begin{proposition}
    \label{prop-random-init-known}
    (Restatement of \citealp[Theorem 1.1]{rudelson2009smallest}) Let $A$ be an $N \times n$ random matrix, $N \geq n$, whose elements are independent copies of a mean zero sub-gaussian random variable with unit variance. Then, for every $\varepsilon>0$, we have
    $\mathbb{P}\left(s_n(A) \leq \varepsilon(\sqrt{N}-\sqrt{n-1})\right) \leq(C \varepsilon)^{N-n+1}+e^{-c N}$
    where $C, c>0$ depend (polynomially) only on the sub-Gaussian moment.
\end{proposition}

Now we can complete the proof of \Cref{prop:random-init}. Note that the entries of $U \in R^{d\times \hat{r}}$ are independently drawn from $N(0,\frac{1}{\hat{r}})$ and $\mV_{\mX_s} \in R^{d\times s}$ is an orthonormal matrix. We write $\mV_{\mX_s}^+ \in R^{d\hat{r}\times s\hat{r}}$ as a block diagonal matrix with $\hat{r}$ copies of $\mV_{\mX_s}$ on the diagonal, and $\mathrm{vec}(\mU) \in R^{d\hat{r}}$ be a vector formed by the concatenation of the columns of $U$. Then $\mV_{\mX_s}^+$ is still orthonormal, and $\mathrm{vec}(\mU) \sim \mathcal{N}(0,\frac{1}{\hat{r}}\mI)$. Since multivariate Gaussian distributions are invariant under orthonormal transformations, we deduce that $(\mV_{\mX_s}^+)^{\top}\mathrm{vec}(\mU) \sim \mathcal{N}(0,\frac{1}{\hat{r}}\mI)$. Equivalently, the entries of $\mV_{\mX_s}^{\top}\mU$ are i.i.d. $\mathcal{N}(0,\frac{1}{\hat{r}})$.

The matrix $\sqrt{\hat{r}} \mV_{\mX_s}^{\top}\mU$ satisfies all the conditions in \Cref{prop-random-init-known}. Thus, with probability at least $1-(C\epsilon)^{\hat{r}-s+1}-e^{-c\hat{r}}$, we have $\sigma_{\min}(\sqrt{\hat{r}} \mV_{\mX_s}^{\top}\mU) \geq \epsilon (\sqrt{\hat{r}}-\sqrt{s-1})$, or equivalently $\sigma_{\min}(\mV_{\mX_s}^{\top}\mU) \geq \epsilon \frac{\sqrt{\hat{r}}-\sqrt{s-1}}{\sqrt{\hat{r}}}$. Finally, the conclusion follows from a union bound:
\begin{equation}
\begin{aligned}
& \mathbb{P}\left[ \exists 1 \leq s \leq \hat{r} \wedge r_* \text { s.t. } \sigma_{\min }\left(\mV_{\mX_s}^{\top} \mU \right)<\frac{\epsilon}{2\hat{r}}\right] \\
\leq & \sum_{s=1}^{\hat{r} \wedge r_*} \mathbb{P}\left[\sigma_{\min }\left( \mV_{\mX_s}^{\top} \mU\right)<\epsilon \frac{\sqrt{\hat{r}}-\sqrt{s-1}}{\sqrt{r}}\right] 
\leq  \sum_{s=1}^{\hat{r} \wedge r_*}\left(e^{-c \hat{r}}+(C \epsilon)^{\hat{r}-s+1}\right) \leq r\left(e^{-c \hat{r}}+C \epsilon\right).
\end{aligned}
\end{equation}

\subsection{Procrustes Distance} \label{sec:proof-for-lemma-pro-pro}

Procrustes distance is introduced in \Cref{sec:prelim-pro}.
The following characterization of the optimal $\mR$ in \Cref{def-procrutes} is known in the literature (see e.g. \citealp[Section 5.2.1]{tu2016low}) but we provide a proof for completeness.

\begin{lemma} \label{lm:R-orth} Let $\mU_1,\mU_2\in\R^{d\times r}$ where $r\leq
    d$. Then for any orthogonal matrix $\mR \in \R^{r\times r}$ 
    that minimizes $\|\mU_1 - \mU_2 \mR \|_F$
    ({\em i.e.}, any orthogonal $\mR$ s.t.~$\left\|\mU_1-\mU_2 \mR\right\|_F = \mathrm{dist}(\mU_1,\mU_2)$),
    $\mU_1^{\top} \mU_2 \mR$ is a symmetric positive
    semi-definite matrix.
\end{lemma}

\begin{proof}
We only need to consider the case when $\mU_2^\top\mU_1 \neq \bm{0}$.
Observe that
\begin{equation}
    \notag
    \begin{aligned}
    \left\|\mU_1-\mU_2 \mR\right\|_{F}^2 
    &= \|\mU_1\|_F^2 + \|\mU_2 \mR \|_F^2 - 2\tr{\mR^{\top} \mU_2^{\top} \mU_1} \\
    &= \|\mU_1\|_F^2 + \|\mU_2\|_F^2 - 2\tr{\mR^{\top} \mU_2^{\top} \mU_1}.
    \end{aligned}
\end{equation}
Let
$\mA\mSigma \mB^{\top}$
be the SVD of $\mU_2^{\top} \mU_1$, where $\mA^\top \mA = \mI, \mB^\top \mB = \mI$ and $\mSigma \succ \vzero$.
Then
\begin{equation}
    \notag
    \tr{\mR^{\top} \mU_2^{\top} \mU_1} = \tr{\mB^{\top} \mR^{\top} \mA\mSigma} \leq \left\|\mB^{\top} \mR^{\top} \mA\right\|\tr{\mSigma} = \tr{\mSigma},
\end{equation}
where the final step is due to orthogonality of $\mB^{\top} \mR^{\top} \mA \in
\R^{s\times s}$, and equality holds if and only if $\mB^{\top} \mR^{\top}
\mA=\mI$. Let $C = \mR^\top\mA$. Let $\vb_i, \vc_i \in \R^d$ be the $i$-th
column of $\mB$ and $\mC$ respectively, then $\mB^\top\mC=\mI$ implies that
$\vb_i^\top\vc_i = 1$. Note that $\|\vb_i\|_2=\|\vc_i\|_2=1$, so we must have
$\vb_i=\vc_i$ for all $i$, \emph{i.e.,} $\mB=\mC = \mR^\top\mA$. Therefore,
$\mU_1^{\top} \mU_2 \mR = \mB \mSigma \mA^\top \mR = \mB\mSigma \mB^{\top}$, which implies that
$\mU_1^{\top} \mU_2 \mR$ is symmetric and positive semi-definite.
\end{proof}

\section{Main idea for the proof of \Cref{main_result}}
\label{appsec-main-idea}
In this section, we briefly introduce our main ideas for proving
\Cref{main_result}. Motivated by \citet{stoger2021small}, we decompose the matrix
$\mU_t$ into a parallel component and an orthogonal component. Specifically, we
write
\begin{equation}
    \label{decomposition}
    \mU_t = \underbrace{\mU_t \mW_t \mW_t^{\top}}_{\text{parallel component}} + \underbrace{\mU_t \mW_{t,\perp}\mW_{t,\perp}^{\top}}_{\text{orthogonal component}},
\end{equation}
where
$\mW_t := \mW_{\mV_{\mX_s}^{\top}\mU_t} \in \R^{\hr \times s}$
is the matrix consisting of the right singular vectors of $\mV_{\mX_s}^{\top}\mU_t$ (\Cref{def:svd})
and $\mW_{t,\perp} \in \R^{\hr \times (\hr-s)}$ is an orthogonal complement of $\mW_t$.
Our goal is to prove that at some time $t$, we have $\mV_{\mX_s}^{\top}\left( \mU_t \mU_t^{\top} - \mX_s \mX_s^{\top}\right) \approx 0$ and $\left\|\mU_t \mW_{t,\perp} \right\|\approx 0$. As we will see later, these imply that $\left\| \mU_t \mU_t^{\top} - \mX_s \mX_s^{\top}\right\|\approx 0$. In the remaining part of this section we give a heuristic explanation for considering \Cref{decomposition}.

\paragraph{Additional Notations.} Let $\mV_{\mX_s,\perp} \in \R^{d \times (d-s)}$ be an orthogonal complement of $\mV_{\mX_s} \in \R^{d \times s}$.
Let $\mSigma_s = \diag{\sigma_1, \dots, \sigma_s}$
and $\mSigma_{s,\perp} = \diag{\sigma_{s+1}, \dots, \sigma_r,0,\cdots,0}\in \R^{(d-s)\times(d-s)}$.
We use $\mDelta_t := (\A^* \A - \mI) (\mX\mX^{\top}-\mU_t \mU_t^{\top})$
to denote the vector consisting of measurement errors for $\mX\mX^{\top}-\mU_t \mU_t^{\top}$.

\subsection{Heuristic explanations of the decomposition}
\label{heuristic}

A simple and intuitive approach for showing the implicit low rank bias is to
directly analyze the growth of $\mV_{\mX_s}^{\top} \mU_t$ versus
$\mV_{\mX_s,\perp}^{\top} \mU_t$. Ideally, the former grows faster than the
latter, so that GD only learns the components in $\mX_s$.

By the update rule of GD~\Cref{GD},
\begin{align*}
    \mV_{\mX_s,\perp}^{\top} \mU_{t+1}
    &= \mV_{\mX_s,\perp}^\top \left[ \mI + \mu\A^*\A(\mX\mX^{\top}-\mU_t \mU_t^{\top})\right] \mU_t \\
    &= \underbrace{\mV_{\mX_s,\perp}^\top \left[ \mI + \mu\mX\mX^{\top}- \mu\mU_t \mU_t^{\top}\right] \mU_t}_{=: \mG_{t,1}}
    + \mu \underbrace{\mV_{\mX_s,\perp}^\top \mDelta_t \mU_t}_{=: \mG_{t,2}} \\
    &= \mG_{t,1} + \mu\mG_{t,2}.
\end{align*}
For the first term $\mG_{t,1}$, we have
\begin{align*}
    \mG_{t,1}
    &= (\mI+\mu \mSigma_{s,\perp}^2)\mV_{\mX_s,\perp}^{\top} \mU_t - \mu \mV_{\mX_s,\perp}^{\top} \mU_t \mU_t^{\top} \mU_t \\
    &= (\mI+\mu \mSigma_{s,\perp}^2)\mV_{\mX_s,\perp}^{\top} \mU_t (\mI-\mu \mU_t \mU_t^{\top}) + \O(\mu^2),
\end{align*}
where the last term $\O(\mu^2)$ is negligible when $\mu$ is  sufficiently small.
Since $\|\mSigma_{s,\perp}\| = \sigma_{s+1}$,
the spectral norm of $\mG_{t,1}$ can be bounded by
\begin{align*}
\normsm{\mG_{t,1}}
&\le \normsm{\mI+\mu \mSigma_{s,\perp}^2} \cdot \normsm{\mV_{\mX_s,\perp}^{\top} \mU_t} \cdot \normsm{\mI-\mu \mU_t \mU_t^{\top}} + \O(\mu^2) \\
&\le (1+\mu\sigma_{s+1}^2) \normsm{\mV_{X_s,\perp}^\top \mU_t} + \O(\mu^2).
\end{align*}
However, the main difference with the full-observation case
\citep{Jiang2022AlgorithmicRI} is the second term $\mG_{t,2} := 
\mV_{\mX_s,\perp}^\top \mDelta_t \mU_t$. 
Since the measurement errors $\mDelta_t$ are small but arbitrary,
it is hard to compare this term with $\mV_{\mX_s,\perp}^\top
\mU_{t+1}$.
As a result, we cannot directly bound the growth of
$\normsm{\mV_{X_s,\perp}^\top \mU_t}$.

However, the aforementioned problem
disappears if we turn to bound the growth of
$\normsm{\mV_{\mX_s,\perp}^{\top} \mU_{t+1} \mW_{t,\perp}}$.
To see this, first we deduce the following by repeatedly using $\mV_{\mX_s}^{\top}
\mU_t \mW_{t,\perp} = 0$ due to the definition of $\mW_{t,\perp}$.
\begin{align*}
    \mG_{t,1} \mW_{t,\perp}
    &= \mV_{\mX_s,\perp}^\top \left[ \mI + \mu\mX\mX^{\top}- \mu\mU_t \mU_t^{\top}\right] \mU_t\mW_{t,\perp} \\
    &= \mV_{\mX_s,\perp}^\top (\mI + \mu\mX\mX^{\top})\mU_t\mW_{t,\perp}  - \mu \mV_{\mX_s,\perp}^\top \mU_t \mU_t^{\top} \mU_t\mW_{t,\perp} \\
    &= (\mI+\mu \mSigma_{s,\perp}^2) \mV_{\mX_s,\perp}^{\top} \mU_t \mW_{t,\perp} - \mu \mV_{\mX_s,\perp}^\top \mU_t (\mW_t \mW_t^\top + \mW_{t,\perp}\mW_{t,\perp}^\top) \mU_t^{\top} \mU_t\mW_{t,\perp} \\
    &= (\mI+\mu \mSigma_{s,\perp}^2) \mV_{\mX_s,\perp}^{\top} \mU_t \mW_{t,\perp}(\mI - \mu \mW_{t,\perp}^\top \mU_t^\top \mU_t \mW_{t,\perp}) \\
    &\quad - \mu \mV_{\mX_s,\perp}^\top \mU_t \mW_t \mW_t^\top \mU_t^{\top} \mU_t\mW_{t,\perp} + \O(\mu^2),
\end{align*}
\begin{align*}
    \mG_{t,2} \mW_{t,\perp}
    &= \mV_{\mX_s,\perp}^\top \mDelta_t \mU_t \mW_{t,\perp}
    = \mV_{\mX_s,\perp}^\top \mDelta_t \mV_{\mX_s,\perp}\mV_{\mX_s,\perp}^{\top} \mU_t \mW_{t,\perp},
\end{align*}
So we have the following recursion:
\begin{align*}
\mV_{\mX_s,\perp}^{\top} \mU_{t+1} \mW_{t,\perp} &= 
    (\mI+\mu \mSigma_{s,\perp}^2 + \mu \mV_{\mX_s,\perp}^\top \mDelta_t \mV_{\mX_s,\perp}) \mV_{\mX_s,\perp}^{\top} \mU_t \mW_{t,\perp}(\mI - \mu \mW_{t,\perp}^\top \mU_t^\top \mU_t \mW_{t,\perp}) \\
    &\quad - \mu \mV_{\mX_s,\perp}^\top \mU_t \mW_t \mW_t^\top \mU_t^{\top} \mU_t\mW_{t,\perp} + \O(\mu^2),
\end{align*}
We further note that
\begin{equation}
    \label{noise_recursion}
    \mV_{\mX_s,\perp}^\top \mU_{t+1} \mW_{t+1, \perp}=\mV_{\mX_s,\perp}^\top \mU_{t+1} \mW_{t} \mW_{t}^\top \mW_{t+1, \perp}+\mV_{\mX_s,\perp}^\top \mU_{t+1} \mW_{t, \perp} \mW_{t, \perp}^\top \mW_{t+1, \perp},
\end{equation}
which establishes the relationship between $\mV_{\mX_s,\perp}^{\top} \mU_{t+1}\mW_{t,\perp}$ and $\mV_{\mX_s,\perp}^{\top} \mU_{t+1}\mW_{t+1,\perp}$. To complete the proof we need to prove the following:
\begin{itemize}
    \item The minimal eigenvalue of the \textit{parallel component} $\mU_t \mW_t \mW_t^{\top}$ grows at a linear rate with speed strictly faster than $\sigma_{s+1}$.
    \item The term $\left\|\mV_{\mX_s,\perp}^\top\mV_{\mU_t\mW_t}\right\| \ll 1$, which implies that the first term in \Cref{noise_recursion} is negligible.
\end{itemize}

\section{Proof of \Cref{main-lemma}}
\label{appsec:thm-fit}
In this section, we give the full proof of \Cref{main-lemma}, with some additional technical lemmas left to \Cref{appsec-aux}. 

\mainLemma*

\Cref{subsec:spectral,subsec:improve} are devoted to analyzing the spectral phase and parallel improvement phase, respectively. \Cref{subsec:induct} uses induction to characterize the low-rank GD trajectory in the parallel improvement phase. In \Cref{subsec:refine} we study the refinement phase, which allows us to derive \Cref{main-lemma}.
\subsection{The spectral phase}
\label{subsec:spectral}
Starting from a small initialization $\mU_0 = \alpha \bar{\mU}, \alpha \ll 1$, we first enter the spectral phase where GD behaves similar to power iteration. As in \citet{stoger2021small}, we refer to this phase as the spectral phase. Specifically, we have in the spectral phase that
\begin{equation}
    \notag
    \mU_{t+1} = \left( \mI+\mu\left(\A^*\A\right)(\mX\mX^{\top}-\mU_t \mU_t^{\top})\right) \mU_t \approx \left( \mI+\mu\left(\A^*\A\right)(\mX\mX^{\top})\right) \mU_t.
\end{equation}
The approximation holds with high accuracy as long as $\|\mU_t\|\ll 1$. Moreover we have $\mM:=\left(\A^*A\right)(\mX\mX^{\top}) \approx \mX\mX^{\top}$ by the RIP condition; when $\delta$ is sufficiently small, we can still ensure a positive eigen-gap of $\mM$. As a result, with small initialization $\mU_t$ would become approximately aligned with the top eigenvector $u_1$ of $\mM$. Since $\|\mM-\mX\mX^{\top}\|=\O(\delta\sqrt{r_*})$ by \Cref{rip-prop-1}, we have $\|\vu_1-\vv_1\| = \O(\delta\sqrt{r_*})$ so that $\|\mV_{\mX_s}^{\top}\mV_{\mU_t\mW_t}\| = \O(\delta\sqrt{r_*})$. This proves the base case for the induction.

Formally, we define $\mM = \A^*\A(\mX\mX^{\top})$, $\mK_t = (\mI+\mu \mM)^t$ and $\mU_t^{\mathtt{sp}} = \mK_t \mU_0$. Suppose that $\mM = \sum_{i=1}^{\rank{\mM}} \hat{\sigma}_i^2\hat{\vv}_i\hat{\vv}_i^{\top}$ is the spectral decomposition of $\mM$ where $\{\hat{\sigma}_i\}_{i\geq 1}$ is sorted in non-increasing order. We additionally define $\mM_s = \sum_{i=1}^{\min\left\{s,\rank{\mM}\right\}}\hat{\sigma}_i^2\hat{\vv}_i\hat{\vv}_i^{\top}$. By \Cref{M-eigen} and $\delta\sqrt{r_*}\leq 10^{-3}\kappa$ by \Cref{asmp:rip}, we have $\hat{\sigma}_s^2\geq \sigma_s^2-0.01\tau$ and $\hat{\sigma}_{s+1}^2 \leq \sigma_{s+1}^2+0.01\tau$, where we recall that $\tau=\min_{s\in[r_*]}\left(\sigma_s^2-\sigma_{s+1}^2\right)>0$. Additionally, let $\mL_t$ be the span of the top-$s$ left singular vectors of $\mU_t$. Recall that \Cref{asmp-full-rank} is made on the initialization. Let 
\begin{equation}
    \notag
    t^{\star}:=\min \left\{i \in \mathbb{N}:\left\|\mU_{i-1}^{\sp}-\mU_{i-1}\right\|>\left\|\mU_{i-1}^{\sp}\right\|\right\},
\end{equation}
the following lemma bounds the error of approximating $\mU_t$ via $\mU_t^{\sp}$:

\begin{lemma}
\label{spectral-phase-gap}
(\citealp[Lemma 8.1]{stoger2021small}) Suppose that $\mathcal{A}$ satisfies the rank-1 RIP with constant $\delta_{1}$. For all integers $t$ such that $1 \leq t \leq t^{\star}$ it holds that
\begin{equation}
    \left\|\mE_{t}\right\|=\left\|\mU_{t}-\mU_{t}^{\sp}\right\| \leq 4\hat{\sigma}_1^{-2} \alpha^{3} r_*\left(1+\delta_{1}\right)\left(1+\mu \hat{\sigma}_1^2\right)^{3 t}.
\end{equation}
\end{lemma}

We can derive the following lower bound on $t^*$ from \Cref{spectral-phase-gap}.

\begin{corollary}
\label{t-star-lower-bound}
We have
\begin{equation}
    \notag
    t^* \geq \frac{\log\alpha^{-1}+ \frac{1}{2}\log \frac{\rho\hat{\sigma}_1^2}{4(1+\delta_1)r_*}}{\log\left(1+\mu\hat{\sigma}_1^2\right)}.
\end{equation}
\end{corollary}

\begin{proof}
By \Cref{spectral-phase-gap} we have
\begin{equation}
    \notag
    \|\mE_{t}\| \leq 4\hat{\sigma}_1^{-2} \alpha^{3} r_*\left(1+\delta_{1}\right)\left(1+\mu \hat{\sigma}_1^2\right)^{3 t}.
\end{equation}
for all $t\leq t^*$. On the other hand, we have
\begin{equation}
    \notag
    \begin{aligned}
    \|\mU_t^{\sp}\| &= \alpha \left\| (\mI+\mu \mM)^t \bar{\mU}\right\| \\
    &\geq \alpha(1+\mu \hat{\sigma}_1^2)^t \left\| \hat{\vv}_1\hat{\vv}_1^{\top} \bar{\mU}\right\| \\
    &\geq \left(1+\mu \hat{\sigma}_1^2\right)^t \alpha\rho.
    \end{aligned}
\end{equation}
Thus, it follows from $\|\mE_{t^*}\| \geq \|\mU_{t^*}^{\sp}\|$ that
\begin{equation}
    \notag
    \left(1+\mu \hat{\sigma}_1^2\right)^{t^*} \geq \sqrt{\frac{\rho\hat{\sigma}_1^2}{4(1+\delta_1)r_*}} \cdot \alpha^{-1} \Rightarrow t^* \geq \frac{\log\alpha^{-1}+ \frac{1}{2}\log \frac{\rho\hat{\sigma}_1^2}{4(1+\delta_1)r_*}}{\log\left(1+\mu\hat{\sigma}_1^2\right)}
\end{equation}
as desired.
\end{proof}

Note that a trivial bound for the rank-$1$ RIP constant is $\delta_1\leq \delta$. We can now show that for small $t$, GD can be viewed as approximate power iteration.

\begin{lemma}
\label{approx_rank1}
There exists a time  
\begin{equation}
    \notag
    t = T_{\alpha}^{\sp} := \frac{2\log\alpha^{-1}+ \log\frac{\rho\hat{\sigma}_1^2}{4r_*(1+\delta)} }{3\log(1+\mu\hat{\sigma_1^2})-\log(1+\mu\hat{\sigma}_{s+1}^2)} \leq t^*
\end{equation}
such that
\begin{equation}
    \notag
    \left\| \mU_t - \sum_{i=1}^s \alpha (1+\mu\hat{\sigma}_i^2)^t \hat{\vv}_i\hat{\vv}_i^{\top} \bar{\mU}\right\| \leq C_1\cdot \alpha^{\gamma}
\end{equation}
where $\gamma = 1-\frac{2\log(1+\mu\hat{\sigma}_{s+1}^2)}{3\log(1+\mu\hat{\sigma}_1^2)-\log(1+\mu\hat{\sigma}_{s+1}^2)}$ and $C_1 = C_1(\mX,\bar{\mU})$ is a constant that only depends on $\mX$ and $\bar{\mU}$.
\end{lemma}

\begin{proof}
It's easy to check that $T_{\alpha}^{\sp} \leq t^*$ by applying \Cref{t-star-lower-bound}.

We consider the following decomposition:
\begin{equation}
    \notag
    \left\| \mU_t - \sum_{i=1}^s \alpha (1+\mu\hat{\sigma}_i^2)^t \hat{\vv}_i\hat{\vv}_i^{\top} \bar{\mU}\right\| \leq \left\| \mU_t-\mU_t^{\sp}\right\| + \left\|\mU_t^{\sp} - \sum_{i=1}^s \alpha (1+\mu\hat{\sigma}_i^2)^t \hat{\vv}_i\hat{\vv}_i^{\top} \bar{\mU}\right\|.
\end{equation}
When $t \leq t^*$, the first term can be bounded as
\begin{equation}
    \notag
    \|\mE_{t}\| \leq 4\hat{\sigma}_1^{-2} \alpha^{3} r_*\left(1+\delta\right)\left(1+\mu \hat{\sigma}_1^2\right)^{3 t}.
\end{equation}
For the second term we have
\begin{equation}
    \notag
    \left\|\mU_t^{\sp} - \sum_{i=1}^s \alpha (1+\mu\hat{\sigma}_i^2)^t \hat{\vv}_i\hat{\vv}_i^{\top} \mU\right\| \leq
    \left\| \sum_{i=s+1}^{r_*} \alpha (1+\mu\hat{\sigma}_i^2)^t \hat{\vv}_i\hat{\vv}_i^{\top} \mU\right\| \leq \alpha \left(1+\mu\hat{\sigma}_{s+1}^2\right)^t.
\end{equation}
In particular, the definition of $T_{\alpha}^{\sp}$ implies that
\begin{equation}
    \notag
    \begin{aligned}
        \left\| \mU_t - \sum_{i=1}^s \alpha (1+\mu\hat{\sigma}_i^2)^t \hat{\vv}_i\hat{\vv}_i^{\top} \mU\right\| &\leq 2\left(\frac{\rho\hat{\sigma}_1^2}{4r_*(1+\delta)}\right)^{\frac{1-\gamma}{2}} \alpha^{\gamma} \\
        &\leq 2\max\left\{ 1, \frac{\rho\hat{\sigma}_1^2}{4r_*(1+\delta)}\right\} \alpha^{\gamma} \\
        &\leq \underbrace{\max\left\{ 2, \frac{\rho\sigma_1^2}{r_*}\right\}}_{:= C_1} \alpha^{\gamma}.
    \end{aligned}
\end{equation}
as desired.
\end{proof}

We conclude this section with the following lemma, which states that initially the parallel component $\mU_t \mW_t$ would grow much faster than the noise term, and would become well-aligned with $\mX_s$.

\begin{lemma}[\Cref{main:spectral-end}, formal version]
\label{spectral-end}
There exists positive constants $C_2 = C_2(\mX,\bar{\mU})$ and $C_3 = C_3(\mX,\bar{\mU})$ such that the following inequalities hold for $t = T_{\alpha}^{\sp}$ when $\alpha \in \left( 0, \left(\frac{\rho}{10C_1(\mX,\bar{\mU})}\right)^{10\kappa}\right)$:
\begin{subequations}
    \label{induction-base}
    \begin{align}
    \|\mU_t\| &\leq \|\mX\|\label{induction-base1} \\
    \sigma_{\min}\left( \mU_{t}\mW_{t}\right) &\geq C_2\cdot \alpha^{1-\frac{2\log(1+\mu\hat{\sigma}_s^2)}{3\log(1+\mu\hat{\sigma}_1^2)-\log(1+\mu\hat{\sigma}_{s+1}^2)}}\label{induction-base2} \\
    \left\|\mU_{t}\mW_{t,\perp}\right\| &\leq C_3\cdot\alpha^{1-\frac{2\log(1+\mu\hat{\sigma}_{s+1}^2)}{3\log(1+\mu\hat{\sigma}_1^2)-\log(1+\mu\hat{\sigma}_{s+1}^2)}}\label{induction-base3} \\
    \left\| \mV_{\mX_s,\perp}^{\top} \mV_{\mU_{t}\mW_{t}}\right\| &\leq 200\delta\label{induction-base4}
    \end{align}
\end{subequations}
\end{lemma}

\begin{proof}
We prove this lemma by applying \Cref{signal-noise-separation} to $t = T_{\alpha}^{\sp}$ defined in the previous lemma. 

The inequality \Cref{induction-base1} can be directly verified by using \Cref{approx_rank1}:
\begin{equation}
    \notag
    \|\mU_t\| \leq \alpha\left(1+\mu\hat{\sigma}_1^2\right)^{t} + \alpha^{\gamma} \leq \left(1+ \left(\frac{C_1(\mX,\bar{\mU})\hat{\sigma}_1^2}{4r_*(1+\delta)}\right)^{\frac{1}{3}}\right)\cdot \alpha^{\gamma/3} \leq \hat{C}_1(\mX,\bar{\mU}) \cdot \alpha^{\gamma/3}\|\mX\|.
\end{equation}
where $\hat{C}_1(\mX,\bar{\mU})=1+ \left(\frac{C_1(\mX,\bar{\mU})\|\mX\|^2}{2r_*(1+\delta)}\right)^{\frac{1}{3}}$ (the constant $C_1$ is defined in the previous lemma). The last inequality holds when $\alpha$ is sufficiently small.
For the remaining inequalities, we first verify that the assumption in \Cref{signal-noise-separation}:
\begin{equation}
    \label{signal-noise-separation-cond}
    \alpha\sigma_s(\mK_t) > 10\left( \alpha\sigma_{s+1}(\mK_t)+\|\mE_t\|\right).
\end{equation}
By definition of $\mK_t$, we can see that for $\alpha \leq \left( \frac{\rho}{10 C_1}\right)^{10\kappa}$,
\begin{equation}
    \notag
    \begin{aligned}
    \alpha\sigma_{s+1}(\mK_{t})+\|\mE_{t}\|
    &\leq  \alpha \left(1+\mu\hat{\sigma}_{s+1}^2\right)^{t} + 4 \hat{\sigma}_1^{-2} \alpha^3 r_*(1+\delta)\left(1+\mu \hat{\sigma}_1^2\right)^{3 t}  \\
    &\leq C_1(\mX,\bar{\mU})\cdot \alpha^{\gamma} \leq 0.1\rho \alpha^{1-\frac{2\log(1+\mu\hat{\sigma}_s^2)}{3\log(1+\mu\hat{\sigma}_1^2)-\log(1+\mu\hat{\sigma}_{s+1}^2)}}\\
    &\leq 0.1\alpha\sigma_s(\mK_{t})
    \end{aligned}
\end{equation}
where $\|\mE_{T_{\alpha}^{\sp}}\|$ is bounded in the previous lemma. Hence \Cref{signal-noise-separation-cond} holds. Let $\mL$ be the span of top-$s$ eigenvectors of $\mM$, then by \Cref{signal-noise-separation}, at $t = T_{\alpha}^{\sp}$ we have
\begin{equation}
    \begin{aligned}
    \sigma_{s}\left(\mU_{t} \mW_{t}\right) &\geqslant 0.4 \alpha \sigma_{s}\left(\mK_{t}\right) \sigma_{\min }\left(\mV_{\mL}^{\top} \bar{\mU}\right) \\
    &\geq 0.1 \alpha\rho \left( 1+\mu\hat{\sigma}_s^2\right)^{t} \\
    &= 0.1\rho \left(\frac{\rho\hat{\sigma}_1^2}{4r_*(1+\delta)}\right)^{\frac{ \log \left(1+\mu \hat{\sigma}_{s}^2\right)}{3 \log \left(1+\mu \hat{\sigma}_1^2\right)-\log \left(1+\mu \hat{\sigma}_{s+1}^2\right)}} \alpha^{1-\frac{2\log(1+\mu\hat{\sigma}_s^2)}{3\log(1+\mu\hat{\sigma}_1)-\log(1+\mu\hat{\sigma}_{s+1}^2)}}\\
    &\geq \underbrace{0.1\rho\left(\frac{\rho\sigma_1^2}{8r_*}\right)^{\frac{1}{10\kappa}}}_{:=C_2(\mX,\bar{\mU})} \alpha^{1-\frac{2\log(1+\mu\hat{\sigma}_s^2)}{3\log(1+\mu\hat{\sigma}_1)-\log(1+\mu\hat{\sigma}_{s+1}^2)}} \\
    \left\|\mU_{t} \mW_{t, \perp}\right\| & \leqslant 2\alpha \sigma_{s+1}^2\left(\mK_{t}\right)+\left\|\mE_{t}\right\| \\
    &\leq \underbrace{2 C_1(\mX,\bar{\mU})}_{:= C_3(\mX,\bar{\mU})} \alpha^{1-\frac{2\log(1+\mu\hat{\sigma}_{s+1}^2)}{3\log(1+\mu\hat{\sigma}_1^2)-\log(1+\mu\hat{\sigma}_{s+1}^2)}} \\
    \left\|\mV_{\mX_s,\perp}^{\top} \mV_{\mU_{t} \mW_{t}}\right\| & \leqslant 100\left(\delta+\frac{\alpha \sigma_{s+1}\left(\mK_{t}\right)+\left\|\mE_{t}\right\|}{\alpha \rho \sigma_{s}\left(\mK_{t}\right)}\right) \\
    &\leq 100\left( \delta \alpha^{\frac{2\log(1+\mu\hat{\sigma}_{s}^2)-2\log(1+\mu\hat{\sigma}_{s+1}^2)}{3\log(1+\mu\hat{\sigma}_1^2)-\log(1+\mu\hat{\sigma}_{s+1}^2)}}\right) \\
    &\leq 200\delta.
    \end{aligned}
\end{equation}
The conclusion follows.
\end{proof}

\subsection{The parallel improvement phase}
\label{subsec:improve}

This subsection is devoted to proving \Cref{main:induction-lemma} which we recall below.

\inductionLemma*

\subsubsection{The parallel component}
In the following we bound $\sigma_{\min}\left( \mV_{\mX_s}^{\top} \mU_{t+1}\mW_t\right)$. We state our main result of this section in the lemma below.

\begin{lemma}
\label{signal-main}
Suppose that \Cref{asmp:rip,asmp-full-rank,asmp:step-size} holds, $\|\mV_{\mX_s^{\perp}}\mV_{\mU_t \mW_t}\| \leq c_3 < 10^{-2}\kappa^{-1}$ and $\mDelta_t = \left(\A^*\A-\mI\right)(\mX\mX^{\top}-\mU_t \mU_t^{\top})$ satisfies $\left\|\mDelta_t\right\|\leq 0.2\kappa^{-1}r_{*}^{-\frac{1}{2}}\|\mX\|^2$, then we have
\begin{equation}
    \notag
    \begin{aligned}
        &\quad \sigma_{\min}(\mV_{\mX_s}^{\top} \mU_{t+1}) \geq \sigma_{\min}(\mV_{\mX_s}^{\top} \mU_{t+1}\mW_t)\\ 
        &\geq \left[ 1+\mu\left(\sigma_s^2-5c_3\|\mX\|^2-2\|\mDelta_t\|\right)-20\mu^2\|\mX\|^4\right] \left( 1- \mu\sigma_{\min}^2(\mV_{\mX_s}^{\top}\mU_t)\right)\sigma_{\min}(\mV_{\mX_s}^{\top}\mU_t).
    \end{aligned}
\end{equation}
\end{lemma}

\begin{proof}
The update rule of GD implies that
\begin{subequations}
    \label{signal-1}
    \begin{align}
    &\quad \mV_{\mX_s}^{\top} \mU_{t+1}\mW_t\nonumber \\
    &= \mV_{\mX_s}^{\top}\left( \mI + \mu( \mX_s \mX_s^{\top}-\mU_t \mU_t^{\top})+\mu\mDelta_t\right)\mU_t \mW_t\label{signal-1-1} \\
    &= (\mI+\mu\mSigma_{s}^2) \mV_{\mX_s}^{\top} \mU_{t} \mW_{t} -\mu \mV_{\mX_s}^{\top} \mU_{t} \mU_{t}^{\top} \mU_{t} \mW_{t}+\mu \mV_{\mX_s}^{\top}\mDelta_t \mU_{t} \mW_{t}\label{signal-1-2} \\
    &= (\mI+\mu\mSigma_{s}^2) \mV_{\mX_s}^{\top} \mU_{t} \mW_{t} -\mu \mV_{\mX_s}^{\top} \mU_{t} \mU_{t}^{\top} \mV_{\mX_s}\mV_{\mX_s}^{\top} \mU_{t} \mW_{t} - \mu \mV_{\mX}^{\top} \mU_{t} \mU_{t}^{\top} \mV_{\mX_s,\perp}\mV_{\mX_s,\perp}^{\top} \mU_{t} \mW_{t}\nonumber\\
    &\quad + \mu \mV_{\mX_s}^{\top}\mDelta_t \mU_{t} \mW_{t}\nonumber \\
    &= (\mI+\mu\mSigma_{s}^2) \mV_{\mX_s}^{\top} \mU_{t} \mW_{t} (\mI-\mu \mW_t^{\top} \mU_{t}^{\top} \mV_{\mX_s}\mV_{\mX_s}^{\top} \mU_{t} \mW_{t}) +  \mu \mV_{\mX_s}^{\top}\mDelta_t \mU_{t} \mW_{t}\nonumber\\
    &\quad - \mu \mV_{\mX_s}^{\top} \mU_{t} \mU_{t}^{\top} \mV_{\mX_s,\perp}\mV_{\mX_s,\perp}^{\top} \mU_{t} \mW_{t} +\mu^2\mSigma_s^2 \mV_{\mX_s}^{\top} \mU_{t} \mW_{t} \mW_t^{\top} \mU_{t}^{\top} \mV_{\mX_s}\mV_{\mX_s}^{\top} \mU_{t} \mW_{t} \label{signal-1-4}
    \end{align}
\end{subequations}

where \Cref{signal-1-1} follows from $\mV_{\mX_s}^{\top} \mX\mX^{\top} = \mV_{\mX_s}^{\top} \mX_s\mX_s^{\top} + \mV_{\mX_s}^{\top} \mX_{s,\perp}\mX_{s,\perp}^{\top}$ and $\mV_{\mX_s}^{\top} \mX_{s,\perp} = 0$; \Cref{signal-1-2} follows from $\mV_{\mX_s}^{\top} \mX_s\mX_s^{\top} = \mV_{\mX_s}^{\top} \mV_{\mX_s}\mSigma_s \mV_{\mX_s}^{\top} = \mSigma_s \mV_{\mX_s}^{\top}$, and \Cref{signal-1-4} follows from $\mV_{\mX_s}^{\top} \mU_t = \mV_{\mX_s}^{\top} \mU_t \mW_t \mW_t^{\top} + \mV_{\mX_s}^{\top} \mU_t \mW_{t,\perp}\mW_{t,\perp}^{\top} = \mV_{\mX_s}^{\top} \mU_t \mW_t \mW_t^{\top}$ by definition of $\mW_t$ and $\mW_{t,\perp}$.
\\

We now relate the last three terms in \Cref{signal-1-4} to $\mV_{\mX_s}^{\top} \mU_t \mW_t$. Since $\mV_{\mX_s}^{\top} \mU_t \mW_t$ is invertible by \Cref{asmp-full-rank}, $\mV_{\mX_s}^{\top} \mV_{\mU_t\mW_t},\mSigma_{\mU_t\mW_t}$ and $\mW_{\mU_t\mW_t}$ are also of full rank, thus we have
\begin{equation}
    \label{U_tW_t}
    \begin{aligned}
    \mU_{t} \mW_{t} &= \mU_t \mW_t (\mV_{\mX_s}^{\top} \mU_t \mW_t)^{-1}\mV_{\mX_s}^{\top} \mU_t \mW_t \\
    &= \mU_t \mW_t \left( \mV_{\mX_s}^{\top} \mV_{\mU_t\mW_t}\mSigma_{\mU_t\mW_t}\mW_{\mU_t\mW_t}^{\top}\right)^{-1}\mV_{\mX_s}^{\top} \mU_t \mW_t \\
    &= \mV_{\mU_t\mW_t}\left( \mV_{\mX_s}^{\top} \mV_{\mU_t\mW_t}\right)^{-1}\mV_{\mX_s}^{\top} \mU_t \mW_t.
    \end{aligned}
\end{equation}
Plugging \Cref{U_tW_t} into the second and third terms of \Cref{signal-1} and re-arranging, we deduce that
\begin{equation}
    \label{signal-recursive}
    \begin{aligned}
    &\quad \mV_{\mX_s}^{\top} \mU_{t+1}\mW_t \\
    &= \left(\mI+\mu(\mSigma_{s}^2 + \mP_1 + \mP_2)\right) \mV_{\mX_s}^{\top} \mU_{t} \mW_{t} (\mI-\mu \mW_t^{\top} \mU_{t}^{\top} \mV_{\mX_s}\mV_{\mX_s}^{\top} \mU_{t} \mW_{t}) \\
    &\quad + \mu^2\left( \mSigma_s^2 + \mP_1+\mP_2\right) \mV_{\mX_s}^{\top} \mU_{t} \mW_{t} \mW_t^{\top} \mU_{t}^{\top} \mV_{\mX_s}\mV_{\mX_s}^{\top} \mU_{t} \mW_{t} \\
    &= \left[\mI+\mu\left(\mSigma_{s}^2 + \mP_1 + \mP_2\right) + \mu^2\left( \mSigma_s^2 + \mP_1+\mP_2\right) \mV_{\mX_s}^{\top} \mU_{t} \mW_{t} \mW_t^{\top} \mU_{t}^{\top} \mV_{\mX_s}\left(\mI-\mu \mV_{\mX_s}^{\top} \mU_{t} \mW_{t} \mW_t^{\top} \mU_{t}^{\top} \mV_{\mX_s}\right)^{-1} \right]\cdot\\
    &\quad \mV_{\mX_s}^{\top} \mU_{t} \mW_{t} (\mI-\mu \mW_t^{\top} \mU_{t}^{\top} \mV_{\mX_s}\mV_{\mX_s}^{\top} \mU_{t} \mW_{t})
    \end{aligned}
\end{equation}
where we use the equation $\mA = (\mI-\mu \mA\mA^{\top})^{-1}\mA(\mI-\mu \mA^{\top}\mA)$ with $\mA = \mV_{\mX_s}^{\top} \mU_{t} \mW_{t}$ (when $\mu < \frac{1}{9\|\mX\|^2}$, $\mI-\mu \mA \mA^{\top}$ is invertible by \Cref{bounded-iterate}), and
\begin{equation}
    \label{signal-3}
    \begin{aligned}
    \mP_1 &= \mV_{\mX_s}^{\top}\mU_t\mU_t^{\top}\mV_{\mX_s,\perp}\mV_{\mX_s,\perp}^{\top} \mV_{\mU_t\mW_t}\left( \mV_{\mX_s}^{\top}\mV_{\mU_t\mW_t}\right)^{-1} \\
    \mP_2 &= \mV_{\mX_s}^{\top}\mDelta_t\mV_{\mU_t\mW_t}\left(\mV_{\mX_s}^{\top}\mV_{\mU_t\mW_t}\right)^{-1} 
    \end{aligned}
\end{equation}

By assumption we have 
\begin{equation}
    \notag
    \sigma_{\min}\left( \mV_{\mX_s}^{\top} \mV_{\mU_t \mW_t}\right) \geq \sqrt{1-\left\|\mV_{\mX_s,\perp}^{\top} \mV_{\mU_t \mW_t}\right\|^2} \geq \frac{1}{2},
\end{equation}
so that
\begin{equation}
    \label{p1-bound}
    \|\mP_1\| \leq \left\|\mV_{\mX_s}^{\top}\mU_t\mU_t^{\top}\mV_{\mX_s,\perp}\right\|\cdot\left\|\mV_{\mX_s,\perp}^{\top} \mV_{\mU_t\mW_t}\right\|\cdot\left\|\left( \mV_{\mX_s}^{\top}\mV_{\mU_t\mW_t}\right)^{-1}\right\| \leq 5c_3\|\mX\|^2 \leq 0.1\|\mX\|^2
\end{equation}
and by our assumption we have
\begin{equation}
    \label{p2-bound}
    \|\mP_2\| \leq \left\|\left( \mV_{\mX_s}^{\top}\mV_{\mU_t\mW_t}\right)^{-1}\right\|\cdot\normsm{\mDelta_t}\leq 2\|\mDelta_t\|\leq 0.2\kappa^{-1}r_*^{-\frac{1}{2}}\|\mX\|^2.
\end{equation}
 Moreover, note that $\|\mSigma_s\|^2 = \sigma_1^2 = \|\mX\|^2$, and since $\mu < 10^{-4}\|\mX\|^{-2}$ by \Cref{asmp:step-size}, we have $\left\|\left(\mI-\mu \mV_{\mX_s}^{\top} \mU_{t} \mW_{t} \mW_t^{\top} \mU_{t}^{\top} \mV_{\mX_s}\right)^{-1}\right\|<1.1$. Thus
\begin{equation}
    \notag
    \left\| \left( \mSigma_s^2 + \mP_1+\mP_2\right) \mV_{\mX_s}^{\top} \mU_{t} \mW_{t} \mW_t^{\top} \mU_{t}^{\top} \mV_{\mX_s} \left(\mI-\mu \mV_{\mX_s}^{\top} \mU_{t} \mW_{t} \mW_t^{\top} \mU_{t}^{\top} \mV_{\mX_s}\right)^{-1}\right\| \leq 20\|\mX\|^4.
\end{equation}

The equation \Cref{signal-3} implies that
\begin{equation}
    \notag
    \begin{aligned}
    &\quad \sigma_{\min}(\mV_{\mX_s}^{\top} \mU_{t+1}\mW_t) \\
    &\geq \sigma_{\min}\left(\mI+\mu\left(\mSigma_{s}^2 + \mP_1 + \mP_2\right)+\left( \mSigma_s^2 + \mP_1+\mP_2\right) \mV_{\mX_s}^{\top} \mU_{t} \mW_{t} \mW_t^{\top} \mU_{t}^{\top} \mV_{\mX_s} \left(\mI-\mu \mV_{\mX_s}^{\top} \mU_{t} \mW_{t} \mW_t^{\top} \mU_{t}^{\top} \mV_{\mX_s}\right)^{-1}\right)\cdot\\
    &\qquad \sigma_{\min}\left( \mV_{\mX_s}^{\top} \mU_{t} \mW_{t} (\mI-\mu \mW_t^{\top} \mU_{t}^{\top} \mV_{\mX_s}\mV_{\mX_s}^{\top} \mU_{t} \mW_{t})\right) \\
    &\geq \left( 1+\mu\sigma_{\min}^2(\mSigma_s)-\mu\|\mP_1\|-\mu\|\mP_2\|-20\mu^2\|\mX\|^4\right)\sigma_{\min}(\mV_{\mX_s}^{\top} \mU_t)\left(1-\mu\sigma_{\min}^2(\mV_{\mX_s}^{\top}\mU_t)\right) \\
    &= \left( 1+\mu\sigma_s^2-\mu\|\mP_1\|-\mu\|\mP_2\|-20\mu^2\|\mX\|^4\right)\sigma_{\min}(\mV_{\mX_s}^{\top} \mU_t)\left(1-\mu\sigma_{\min}^2(\mV_{\mX_s}^{\top}\mU_t)\right)
    \end{aligned}
\end{equation}

Recall that $\mP_1$ and $\mP_2$ are bounded in \Cref{p1-bound,p2-bound} respectively, so we have that
\begin{equation}
    \notag
    \begin{aligned}
        &~~~~~\sigma_{\min}(\mV_{\mX_s}^{\top} \mU_{t+1}) \\
        &\geq \sigma_{\min}(\mV_{\mX_s}^{\top} \mU_{t+1}\mW_t)\\ 
        &\geq \left[ 1+\mu\left(\sigma_s^2-5c_3\|\mX\|^2-2\|\mDelta_t\|^2\right)-20\mu^2\|\mX\|^4\right] \left( 1- \mu\sigma_{\min}^2(\mV_{\mX_s}^{\top}\mU_t)\right)\sigma_{\min}(\mV_{\mX_s}^{\top}\mU_t).
    \end{aligned}
\end{equation}
The conclusion follows.
\end{proof}

The corollaries below immediately follow from \Cref{signal-main}.

\begin{corollary}
\label{signal-cor1}
Under the conditions in \Cref{signal-main}, if  $\sigma_{\min}^2(\mV_{\mX_s}^{\top}\mU_t)<0.3\kappa^{-1}\|\mX\|^2$, then we have
\begin{equation}
    \notag
    \sigma_{\min}(\mV_{\mX_s}^{\top} \mU_{t+1}) \geq \left( 1+0.5\mu(\sigma_s^2+\sigma_{s+1}^2)\right)\sigma_{\min}(\mV_{\mX_s}^{\top}\mU_t).
\end{equation}
\end{corollary}

\begin{proof}
    By \Cref{signal-main} it remains to check that
    \begin{equation}
        \notag
        \left[ 1+\mu\left(\sigma_s^2-5c_3\|\mX\|^2-2\|\mDelta_t\|^2\right)-20\mu^2\|\mX\|^4\right] \left( 1- 0.3\mu\kappa^{-1}\|\mX\|^2\right) \geq 1+ 0.5\mu(\sigma_s^2+\sigma_{s+1}^2).
    \end{equation}
    Indeed, recall from the conditions of \Cref{signal-main} that $5c_3\|\mX\|^2 \leq 0.01\kappa^{-1}\|\mX\|^2\leq 0.01(\sigma_s^2-\sigma_{s+1}^2)$ and similarly $\|\mDelta_t\|\leq 0.005(\sigma_s^2-\sigma_{s+1}^2)$ and $\mu^2\|\mX\|^4 \leq 10^{-4}\kappa^{-1}\mu\|\mX\|^2 \leq 10^{-4}(\sigma_s^2-\sigma_{s+1}^2)$, so that
    \begin{equation}
        \notag
        \begin{aligned}
            &\quad \left[ 1+\mu\left(\sigma_s^2-5c_3\|\mX\|^2-2\|\mDelta_t\|^2\right)-20\mu^2\|\mX\|^4\right] \left( 1- 0.3\kappa^{-1}\|\mX\|^2\right) \\
            &\geq \left( 1+ \mu(0.9\sigma_s^2+0.1\sigma_{s+1}^2)\right) \left(1-0.3\mu(\sigma_s^2-\sigma_{s+1}^2)\right) \\
            &= 1+ \mu(0.6\sigma_s^2+0.4\sigma_{s+1}^2) - \mu^2\|\mX\|^2(\sigma_s^2-\sigma_{s+1}^2) \\
            &\geq 1 + 0.5\mu(\sigma_s^2+\sigma_{s+1}^2)
        \end{aligned}
    \end{equation}
    as desired.
\end{proof}

\begin{corollary}
\label{signal-cor2}
Under the conditions in \Cref{signal-main}, if 
\begin{equation}
    \label{signal-cor2-eq}
    \sigma_{\min}^2(\mV_{\mX_s}^{\top}\mU_t) \leq \sigma_s^2-\mu\sigma_s^4 - 5c_3\|\mX\|^2-2\|\mDelta_t\|^2-20\mu^2\|\mX\|^4,
\end{equation}
then we have that $\sigma_{\min}(\mV_{\mX_s}^{\top} \mU_{t+1}) \geq \sigma_{\min}(\mV_{\mX_s}^{\top} \mU_{t})$.
\end{corollary}

\begin{proof}
    A sufficient condition for $\sigma_{\min}(\mV_{\mX_s}^{\top} \mU_{t+1}) \geq \sigma_{\min}(\mV_{\mX_s}^{\top} \mU_{t})$ to hold is that
    \begin{equation}
        \notag
        \begin{aligned}
            &\quad \left[ 1+\mu\left(\sigma_s^2-5c_3\|\mX\|^2-2\|\mDelta_t\|^2\right)-20\mu^2\|\mX\|^4\right] \left( 1- \mu\sigma_{\min}^2(\mV_{\mX_s}^{\top}\mU_t)\right) \\
            &\Leftarrow  \sigma_s^2-5c_3\|\mX\|^2-2\|\mDelta_t\|^2 - 20\mu\|\mX\|^2 - \sigma_{\min}(\mV_{\mX_s}^{\top} \mU_{t}) - \mu\sigma_s^2\sigma_{\min}(\mV_{\mX_s}^{\top} \mU_{t}) \geq 0.
        \end{aligned}
    \end{equation}
    When \Cref{signal-cor2-eq} holds, we have
    \begin{equation}
        \notag
        \begin{aligned}
            &\quad \sigma_s^2-5c_3\|\mX\|^2-2\|\mDelta_t\|^2 - 20\mu\|\mX\|^2 - \sigma_{\min}(\mV_{\mX_s}^{\top} \mU_{t}) \\
            &\geq \mu\sigma_s^4 \geq \mu\sigma_s^2\sigma_{\min}(\mV_{\mX_s}^{\top} \mU_{t}) 
        \end{aligned}
    \end{equation}
    as desired.
\end{proof}

\subsubsection{The orthogonal component}
In this section we turn to analyze the noise term.The main result of this section is presented in the following:

\begin{lemma}
\label{noise}
Suppose that \Cref{asmp:rip,asmp-full-rank,asmp:step-size} hold, $\mV_{\mX_s}^\top\mU_{t+1}\mW_t\in\R^{s\times r}$ is of full rank, $\|\mV_{\mX_s,\perp}\mV_{\mU_t \mW_t}\| \leq c_3 < 10^{-2}\kappa^{-1}$ and $\|\mDelta_t\|\leq c_3\|\mX\|^2$,
then we have
\begin{equation}
    \notag
    \| \mU_{t+1}\mW_{t+1,\perp}\| \leq\left(1+\mu\sigma_{s+1}^2+30\mu\|\mX\|^2 c_3 +0.1\mu^2\|\mX\|^4\right)\left\|\mU_{t} \mW_{t, \perp}\right\|.
\end{equation}
\end{lemma}

\begin{proof}
By the definition of $\mW_{t,\perp}$, we have $\mV_{\mX_s}^{\top} \mU_t \mW_{t,\perp} = 0$, thus $\left\|\mU_t \mW_{t,\perp}\right\| = \left\| \mV_{\mX_s,\perp}^{\top} \mU_t \mW_{t,\perp}\right\|$. The latter can be decomposed as follows:
\begin{equation}
\notag
\mV_{\mX_s,\perp}^{\top} \mU_{t+1} \mW_{t+1, \perp}=\underbrace{\mV_{\mX_s,\perp}^{\top} \mU_{t+1} \mW_{t} \mW_{t}^{\top} \mW_{t+1, \perp}}_{=(a)}+\underbrace{\mV_{\mX_s,\perp}^{\top} \mU_{t+1} \mW_{t, \perp} \mW_{t, \perp}^{\top} \mW_{t+1, \perp}}_{=(b)}.
\end{equation}

In the following, we are going to show that the term (a) is bounded by $c\cdot\mu$ where $c$ is a small constant, while (b) grows linearly with a slow speed.
\\

\textit{Bounding summand (a). } 
Since 
\begin{equation}
    \notag
    0 = \mV_{\mX_s}^{\top} \mU_{t+1} \mW_{t+1,\perp} = \mV_{\mX_s}^{\top} \mU_{t+1} \mW_t \mW_t^{\top} \mW_{t+1,\perp} + \mV_{\mX_s}^{\top} \mU_{t+1} \mW_{t,\perp} \mW_{t,\perp}^{\top} \mW_{t+1,\perp}
\end{equation}
by definition, we have
\begin{equation}
    \label{w-diff}
    \mW_{t}^{\top} \mW_{t+1,\perp} = -\left(\mV_{\mX_s}^{\top} \mU_{t+1} \mW_t\right)^{-1}\mV_{\mX_s}^{\top} \mU_{t+1} \mW_{t,\perp} \mW_{t,\perp}^{\top} \mW_{t+1,\perp}.
\end{equation}
Thus the summand (a) can be rewritten as follows:
\begin{subequations}
    \label{noise-eq0}
    \begin{align}
        &\quad \mV_{\mX_s,\perp}^{\top} \mU_{t+1} \mW_{t} \mW_{t}^{\top} \mW_{t+1, \perp} \nonumber\\
        &=-\mV_{\mX_s,\perp}^{\top} \mU_{t+1} \mW_{t}\left(\mV_{\mX_s}^{\top} \mU_{t+1} \mW_{t}\right)^{-1} \mV_{\mX_s}^{\top} \mU_{t+1} \mW_{t, \perp} \mW_{t, \perp}^{\top} \mW_{t+1, \perp}\label{noise-eq0-1}\\
        &= -\mV_{\mX_s,\perp}^{\top} \mU_{t+1} \mW_{t}\left(\mV_{\mX_s}^{\top} \mV_{\mU_{t+1}\mW_t}\mSigma_{\mU_{t+1}\mW_t}\mW_{\mU_{t+1}\mW_t}\right)^{-1} \mV_{\mX_s}^{\top} \mU_{t+1} \mW_{t, \perp} \mW_{t, \perp}^{\top} \mW_{t+1, \perp}\nonumber\\
        &= -\mV_{\mX_s,\perp}^{\top} \mV_{\mU_{t+1} \mW_{t}}\left(\mV_{\mX_s}^{\top} \mV_{\mU_{t+1} \mW_{t}}\right)^{-1} \mV_{\mX_s}^{\top} \mU_{t+1} \mW_{t, \perp} \mW_{t, \perp}^{\top} \mW_{t+1,\perp}\label{noise-eq0-2}\\
        &= -\mV_{\mX_s,\perp}^{\top} \mV_{\mU_{t+1} \mW_{t}}\left(\mV_{\mX_s}^{\top} \mV_{\mU_{t+1} \mW_{t}}\right)^{-1} \mV_{\mX_s}^{\top} \left( \mI+\mu \A^{*} \A\left(\mX \mX^{\top}-\mU_{t} \mU_{t}^{\top}\right) \right) \mU_t \mW_{t, \perp} \mW_{t, \perp}^{\top} \mW_{t+1, \perp}\nonumber\\
        &= -\mu \mV_{\mX_s,\perp}^{\top} \mV_{\mU_{t+1} \mW_{t}}\left(\mV_{\mX_s}^{\top} \mV_{\mU_{t+1} \mW_{t}}\right)^{-1} \mV_{\mX_s}^{\top}\left[ \left(\mX \mX^{\top}-\mU_{t} \mU_{t}^{\top}\right) + \mDelta_t\right] \mU_t \mW_{t, \perp} \mW_{t, \perp}^{\top} \mW_{t+1, \perp} \label{noise-eq0-3}\\
        &= \mu \mV_{\mX_s,\perp}^{\top} \mV_{\mU_{t+1} \mW_{t}}\left(\mV_{\mX_s}^{\top} \mV_{\mU_{t+1} \mW_{t}}\right)^{-1} \mV_{\mX_s}^{\top}\left[ \mU_{t} \mU_{t}^{\top}- \mDelta_t \right]
        \mU_t \mW_{t, \perp} \mW_{t, \perp}^{\top} \mW_{t+1, \perp}\nonumber\\
        &=\mu \mV_{\mX_s,\perp}^{\top} \mV_{\mU_{t+1} \mW_{t}}\left(\mV_{\mX_s}^{\top} \mV_{\mU_{t+1} \mW_{t}}\right)^{-1} \mM_{1} \mV_{\mX_s,\perp}^{\top} \mU_{t} \mW_{t, \perp} \mW_{t, \perp}^{\top} \mW_{t+1, \perp},\nonumber
    \end{align}
\end{subequations}
where $\mM_1 = \mV_{\mX_s}^{\top}\left[\mU_{t} \mU_{t}^{\top} \mV_{\mX_s,\perp}-\mDelta_t \mV_{\mX_s,\perp}\right]$. In \Cref{noise-eq0}, \Cref{noise-eq0-1} follows from \Cref{w-diff}, \Cref{noise-eq0-2} holds since $\mSigma_{\mU_{t+1}\mW_t}\mW_{\mU_{t+1}\mW_t}^{\top} \in \R^{s\times s}$ is invertible, and in \Cref{noise-eq0-3} we use $\mV_{\mX_s}^{\top} \mU_t \mW_{t,\perp}=0$. It follows that
\begin{equation}
    \label{noise-1}
    \|(a)\| \leq \mu  \left\|\mV_{\mX_s,\perp}^{\top} \mV_{\mU_{t+1} \mW_{t}}\right\|\cdot\left\|\left(\mV_{\mX_s}^{\top} \mV_{\mU_{t+1} \mW_{t}}\right)^{-1}\right\|\left\| \mM_{1}\right\| \left\|\mV_{\mX_s,\perp}^{\top} \mU_{t} \mW_{t, \perp}\right\|.
\end{equation}
By \Cref{appendix-saddle-2} we have $\left\| \mV_{\mX_s,\perp}^{\top} \mV_{\mU_{t+1} \mW_{t}}\right\| \leq 0.01$, which implies that
\begin{equation}
    \label{noise-2}
    \left\|\left(\mV_{\mX_s}^{\top} \mV_{\mU_{t+1} \mW_{t}}\right)^{-1}\right\| = \sigma_{\min}^{-1}\left(\mV_{\mX_s}^{\top} \mV_{\mU_{t+1} \mW_{t}}\right) = \left( 1-\left\| \mV_{\mX_s,\perp}^{\top} \mV_{\mU_{t+1} \mW_{t}}\right\|^2\right)^{-\frac{1}{2}} \leq 1.1.
\end{equation}
Lastly, we bound $\mM_1$ as follows:
\begin{equation}
    \label{noise-3}
    \begin{aligned}
    \|\mM_1\| &\leq \left\| \mV_{\mX_s}^{\top}\mU_{t} \mU_{t}^{\top} \mV_{\mX_s,\perp}\right\| + \left\| \left(\mathcal{A}^{*} \mathcal{A}-\mI \right)\left(\mX \mX^{\top}-\mU_{t} \mU_{t}^{\top}\right)\right\| \\
    &\leq \left\| \mV_{\mX_s}^{\top}\mU_{t}\mW_t\right\|\cdot \left\| \mV_{\mX_s,\perp}^{\top}\mU_{t}\mW_t\right\| + 10^{-3}\kappa^{-1} c_3 \|\mX\|^2 \\
    &\leq 10\|\mX\|^2 c_3.
    \end{aligned}
\end{equation}
where the second inequality follows from our assumption on $\left\| \left(\mathcal{A}^{*} \mathcal{A}-\mI \right)\left(\mX \mX^{\top}-\mU_{t} \mU_{t}^{\top}\right)\right\|$.
Combining \Cref{noise-1}, \Cref{noise-2} and \Cref{noise-3} yields
\begin{equation}
    \notag
    \|(a)\| \leq 20\mu\|\mX\|^2 c_3\| \mU_t \mW_{t,\perp}\|.
\end{equation}
\\

\textit{Bounding summand (b). } This is the main component in the error term. We'll see that although this term can grow exponentially fast, the growth speed is slower than the minimal eigenvalue of the parallel component.

We have
\begin{subequations}
    \label{noise1-b-1}
    \begin{align}
    &\quad \mV_{\mX_s,\perp}^{\top} \mU_{t+1} \mW_{t, \perp}\nonumber\\
    &= \mV_{\mX_s,\perp}^{\top} \left[ \mI  + \mu(\mX \mX^{\top}-\mU_t \mU_t^{\top}) + \mu\left(\mathcal{A}^{*} \mathcal{A}-\mI \right)\left(\mX \mX^{\top}-\mU_{t} \mU_{t}^{\top}\right)\right] \mU_t \mW_{t, \perp}\label{noise1-b-1-1}\\
    &=\left(\mI +\mu\mSigma_{s,\perp}^2 -\mu \mV_{\mX_s,\perp}^{\top} \mU_{t} \mU_{t}^{\top} \mV_{\mX_s,\perp}+\mu \underbrace{\mV_{\mX_s,\perp}^{\top}\mDelta_t \mV_{\mX_s,\perp}}_{=:\mM_2}\right) \mV_{\mX_s,\perp}^{\top} \mU_{t} \mW_{t, \perp}\label{noise1-b-1-2} \\
    &= \left(\mI +\mu\mSigma_{s,\perp}^2-\mu \mV_{\mX_s,\perp}^{\top} \mU_{t} \mW_{t} \mW_{t}^{\top} \mU_{t}^{\top} \mV_{\mX_s,\perp}+\mu \mM_2\right) \mV_{\mX_s,\perp}^{\top} \mU_{t} \mW_{t, \perp} \left(\mI -\mu \mW_{t, \perp}^{\top} \mU_{t}^{\top} \mU_{t} \mW_{t, \perp}\right) \\
    &\quad + \mu^2 \left( \mSigma_{s,\perp}^2- \mV_{\mX_s,\perp}^{\top} \mU_{t} \mW_{t} \mW_{t}^{\top} \mU_{t}^{\top} \mV_{\mX_s,\perp}+ \mM_2\right)\mV_{\mX_s,\perp}^{\top} \mU_{t} \mW_{t, \perp} \mW_{t, \perp}^{\top} \mU_{t}^{\top} \mU_{t} \mW_{t, \perp}\label{noise1-b-1-3}
    \end{align}
\end{subequations}
where we recall that $\mSigma_{s,\perp}^2 = \diag{\sigma_{s+1}^2,\cdots,\sigma_r^2,0,\cdots,0} \in\R^{(d-s)\times(d-s)}$. In \Cref{noise1-b-1}, \Cref{noise1-b-1-1} follows from the update rule of GD, \Cref{noise1-b-1-2} is obtained from $\mV_{\mX_s,\perp}^{\top} \mX \mX^{\top} = \mSigma_{s,\perp}^2 \mV_{\mX_s,\perp}^{\top}$ and $\mU_t \mW_{t,\perp} = \mV_{\mX_s}\mV_{\mX_s}^{\top}\mU_t \mW_{t,\perp} + \mV_{\mX_s,\perp}\mV_{\mX_s,\perp}^{\top} \mU_t \mW_{t,\perp} = \mV_{\mX_s,\perp}\mV_{\mX_s,\perp}^{\top} \mU_t \mW_{t,\perp}$, and lastly in \Cref{noise1-b-1-3} we use
\begin{equation}
    \notag
    \begin{aligned}
    &\quad \mV_{\mX_s,\perp}^{\top} \mU_{t} \mU_{t}^{\top} \mV_{\mX_s,\perp}\mV_{\mX_s,\perp}^{\top} \mU_{t} \mW_{t, \perp} \\
    &= \mV_{\mX_s,\perp}^{\top} \mU_{t} \mW_t \mW_t^{\top} \mU_{t}^{\top} \mV_{\mX_s,\perp}\mV_{\mX_s,\perp}^{\top} \mU_{t} \mW_{t, \perp} + \mV_{\mX_s,\perp}^{\top} \mU_t \mW_{t,\perp} \mW_{t,\perp}^{\top} \mU_t^{\top} \mV_{\mX_s,\perp}\mV_{\mX_s,\perp}^{\top} \mU_{t} \mW_{t, \perp} \\
    &= \mV_{\mX_s,\perp}^{\top} \mU_{t} \mW_t \mW_t^{\top} \mU_{t}^{\top} \mV_{\mX_s,\perp}\mV_{\mX_s,\perp}^{\top} \mU_{t} \mW_{t, \perp} + \mV_{\mX_s,\perp}^{\top} \mU_t \mW_{t,\perp} \mW_{t,\perp}^{\top} \mU_t^{\top} \mU_{t} \mW_{t, \perp}.
    \end{aligned}
\end{equation}

It follows that
\begin{equation}
    \notag
    \begin{aligned}
    &\quad \left\|\mV_{\mX_s,\perp}^{\top} \mU_{t+1} \mW_{t, \perp}\right\| \\
    &\leq \left( \left\| \mI - \mu \mV_{\mX_s,\perp}^{\top} \mU_{t} \mW_{t} \mW_{t}^{\top} \mU_{t}^{\top} \mV_{\mX_s,\perp}\right\| + \mu\|\mSigma_{s,\perp}\|^2 + \mu\|\mM_2\|\right)  \|\mV_{\mX_s,\perp}^{\top} \mU_{t} \mW_{t, \perp}\|\left( \mI-\mu \|\mV_{\mX_s,\perp}^{\top} \mU_{t} \mW_{t, \perp}\|^2\right) \\ 
    &\quad + \mu^2\left\|\mU_t \mW_{t,\perp}\right\|^3\left( \sigma_{s+1}^2+\|\mU_t\|^2+10^{-3}\kappa^{-1} c_3 \|\mX\|^2\right) \\
    &\leq \left( 1 + \mu\sigma_{s+1}^2+\mu\|\mDelta_t\|\right)\|\mU_{t} \mW_{t, \perp}\|\left( 1-\mu \|\mU_{t} \mW_{t, \perp}\|^2\right) + 0.1\mu^2\|\mX\|^4\left\|\mU_t \mW_{t,\perp}\right\| \\
    &\leq \left\|\mU_{t} \mW_{t, \perp}\right\|\left(1+\mu\sigma_{s+1}^2+\mu c_3\|\mX\|^2 +0.1\mu^2\|\mX\|^4\right)
    \end{aligned}
\end{equation}
To summarize, we have
\begin{equation}
    \notag
    \| \mU_{t+1}\mW_{t+1,\perp}\| \leq\left(1+\mu\sigma_{s+1}^2+30\mu\|\mX\|^2 c_3 +0.1\mu^2\|\mX\|^4\right)\left\|\mU_{t} \mW_{t, \perp}\right\|
\end{equation}
as desired.
\end{proof}

To bound the growth speed of the orthogonal component, we need to show that the quantity $\left\|\mV_{\mX_s,\perp}^{\top} \mV_{\mU_{t} \mW_{t}}\right\|$ remains small. The following lemma serves to complete an induction step from $t$ to $t+1$:
\begin{lemma}
\label{noise-lemma-2}
Suppose $\mV_{\mX_s}^{\top}\mU_t$ is of full rank, $\|\mV_{\mX_s,\perp}\mV_{\mU_t \mW_t}\| \leq c_3$ and $\|\mU_t \mW_{t,\perp}\|\leq \min\left\{\sigma_{\min}(\mU_t\mW_t), c_4\right\}$ with $\max\left\{ c_3,c_4\|\mX\|^{-1}\right\} \leq 10^{-2}\kappa^{-1}$, and $\mDelta_t = (\A^*\A-\mI)(\mX\mX^{\top}-\mU_t\mU_t^{\top})$ satisfies $\left\|\mDelta_t\right\| \leq 10^{-3}\kappa^{-1}c_3\|\mX\|^2$  and $\mu\leq 10^{-4}\kappa^{-1}\|\mX\|^{-2}c_3$, then we have $\|\mV_{\mX_s,\perp}\mV_{\mU_{t+1} \mW_{t+1}}\| \leq c_3$.
\end{lemma}

\begin{proof}
Let $\mM_t = \A^*\A(\mX\mX^{\top}-\mU_t \mU_t^{\top})$, so the update rule of GD implies that
\begin{equation}
    \notag
    \begin{aligned}
    \mU_{t+1}\mW_{t+1} &= (\mI+\mu \mM_t)\mU_t \mW_{t+1} \\
    &=(\mI+\mu \mM_t) \left( \mU_{t} \mW_{t} \mW_{t}^{\top} \mW_{t+1}+ \mU_{t} \mW_{t, \perp} \mW_{t, \perp}^{\top} \mW_{t+1}\right)\\
    &=(\mI +\mu \mM_t) \left( \mV_{\mU_{t} \mW_t} \mV_{\mU_{t} \mW}^{\top} \mU_{t} \mW_{t} \mW_{t}^{\top} \mW_{t+1}+ \mU_{t} \mW_{t, \perp} \mW_{t, \perp}^{\top} \mW_{t+1}\right) \\
    &= \underbrace{(\mI +\mu \mM_t) (\mI +\mP) \mV_{\mU_{t} \mW_{t}}}_{:= \mH} \mV_{\mU_{t} \mW_{t}}^{\top} \mU_{t} \mW_{t} \mW_{t}^{\top} \mW_{t+1},
    \end{aligned}
\end{equation}
where
\begin{equation}
    \notag
    \mP = \mU_{t} \mW_{t, \perp} \mW_{t, \perp}^{\top} \mW_{t+1}\left(\mV_{\mU_{t} \mW_{t}}^{\top} \mU_{t} \mW_{t} \mW_{t}^{\top} \mW_{t+1}\right)^{-1} \mV_{\mU_{t} \mW_{t}}^{\top}
\end{equation}
and $\mV_{\mU_{t} \mW_{t}}^{\top} \mU_{t} \mW_{t} \mW_{t}^{\top} \mW_{t+1}$ is invertible since $\mV_{\mU_t\mW_t}^{\top}\mU_t\mW_t$ is invertible by our assumption that $\mV_{\mX_s}^{\top}\mU_t$ is of full rank and $\rank{\mU_{t} \mW_{t}} \geq \rank{\mV_{\mX_s}^{\top}\mU_t\mW_t}=\rank{\mV_{\mX_s}^{\top}\mU_t}=s$, and $\mW_t^{\top}\mW_{t+1}$ is invertible by \Cref{row-space-change}. Indeed, \Cref{row-space-change} implies that $\sigma_{\min}\left(\mW_t^{\top}\mW_{t+1}\right) \geq \frac{1}{2}$ by our condition on $\mu$.

The key observation here is that because the (square) matrix $\mV_{\mU_{t} \mW_{t}}^{\top} \mU_{t} \mW_{t} \mW_{t}^{\top} \mW_{t+1}$ is invertible, so that the column space of $\mU_{t+1}\mW_{t+1}$ is the same as that of $\mH$. Following the line of proof of \citealp[Lemma 9.3]{stoger2021small} (for completeness, we provide details in \Cref{Z-lem}), we deduce that
\begin{equation}
    \label{noise-5}
    \begin{aligned}
    &\quad \left\|\mV_{\mX_s,\perp}^{\top} \mV_{\mU_{t+1} \mW_{t+1}}\right\| = \left\|\mV_{\mX_s,\perp}^{\top} \mV_{\mH} \mW_{\mH}^{\top}\right\| \\
    &\leq \left\|\mV_{\mX_s,\perp}^{\top} \left[ \left(\mI+\mB-\frac{1}{2} \mV_{\mU_{t} \mW_{t}} \mV_{\mU_{t} \mW_{t}}^{\top}\left(\mB+\mB^{\top}\right)\right) \mV_{\mU_{t} \mW_{t}}- \mB \mV_{\mU_{t} \mW_{t}} \mV_{\mU_{t} \mW_{t}}^{\top}\left(\mB+\mB^{\top}\right) \mV_{\mU_{t} \mW_{t}}+ \mD\right]   \right\| \\
    &\leq \left\| \mV_{\mX_s,\perp}^{\top}  \left(\mI+\mB-\frac{1}{2} \mV_{\mU_{t} \mW_{t}} \mV_{\mU_{t} \mW_{t}}^{\top}\left(\mB+\mB^{\top}\right)\right) \mV_{\mU_{t} \mW_{t}}\right\| + 2\|\mB\|^2 + \|\mD\|
    \end{aligned}
\end{equation}
where $\mB = (\mI+\mu \mM_t) (\mI+\mP) - \mI$ and $\|\mD\|\leq 100\|\mB\|^2$. By assumption we have
\begin{equation}
    \notag
    \begin{aligned}
    \|\mP\| &\leq \frac{\left\| \mU_t \mW_{t,\perp}\right\|\left\| \mW_{t,\perp} \mW_{t+1}\right\|}{\sigma_{\min}(\mU_t \mW_t)\sigma_{\min}(\mW_t^{\top} \mW_{t+1})} \\
    &\leq 2\left\| \mW_{t,\perp} \mW_{t+1}\right\|,
    \end{aligned}
\end{equation}
so that
\begin{equation}
    \label{noise-4}
    \begin{aligned}
    &\quad \left\| \mB - \mu(\mX\mX^{\top}-\mU_t \mU_t^{\top})\right\|\\
    &\leq \mu\|\mM_t - (\mX\mX^{\top}-\mU_t \mU_t^{\top})\| + \|\mP\| + \mu\|\mM_t\|\|\mP\| \\
    &\leq \mu \left\| \mDelta_t\right\| + 2\left\| \mW_{t,\perp} \mW_{t+1}\right\| + 4\mu\|\mX\|^2\left\| \mW_{t,\perp} \mW_{t+1}\right\| \\
    &\leq \mu \left\| \mDelta_t\right\| + 6\left\| \mW_{t,\perp} \mW_{t+1}\right\| \\
    &\leq  18\mu\left(10\mu\|\mX\|^3+ c_4\right)c_3\|\mX\| + 7\mu\left\|\mDelta_t\right\| \\
    &\leq 18\mu\left(10\mu\|\mX\|^3+ c_4\right)c_3\|\mX\| + 0.01\mu \kappa^{-1}c_3\|\mX\|^2
    \end{aligned}
\end{equation}
where we use \Cref{row-space-change} to bound $\left\|\mW_{t,\perp}^{\top} \mW_{t+1}\right\|$. Let $\mB_1 = \mu(\mX\mX^{\top}-\mU_t \mU_t^{\top})$ and $\mR_1 = \mV_{\mX_s,\perp}^{\top}  \left(\mI+\mB_1-\mV_{\mU_{t} \mW_{t}} \mV_{\mU_{t} \mW_{t}}^{\top}\mB_1\right) \mV_{\mU_{t} \mW_{t}}$, then we have

\begin{equation}
    \begin{aligned}
    \mR_1 &= \mV_{\mX_s,\perp}^{\top}\left( \mI+\mu\left(\mI-\mV_{\mU_{t} \mW_{t}} \mV_{\mU_{t} \mW_{t}}^{\top}\right)\left(\mX \mX^{\top}-\mU_{t} \mU_{t}^{\top}\right)\right) \mV_{\mU_{t} \mW_{t}} \\
    &= \left(\mI+\mu\mSigma_{s,\perp}^2\right) \mV_{\mX_s,\perp}^{\top} \mV_{\mU_{t} \mW_{t}}\left(\mI-\mu \mV_{\mU_{t} \mW_{t}}^{\top} \mX \mX^{\top} \mV_{\mU_{t} \mW_{t}}\right) \\
    &\quad -\mu \mV_{\mX_s,\perp}^{\top}\left(\mI-\mV_{\mU_{t} \mW_{t}} \mV_{\mU_{t} \mW_{t}}^{\top}\right) \mU_{t} \mW_{t, \perp} \mW_{t, \perp}^{\top} \mU_{t}^{\top} \mV_{\mX_s,\perp} \mV_{\mX_s,\perp}^{\top} \mV_{\mU_{t} \mW_{t}} \\
    &\quad + \mu^2\mSigma_{s,\perp}^2\mV_{\mX_s,\perp}^{\top} \mV_{\mU_{t} \mW_{t}}\mV_{\mU_{t} \mW_{t}}^{\top} \mX \mX^{\top} \mV_{\mU_{t} \mW_{t}}.
    \end{aligned}
\end{equation}
By Weyl's inequality (cf. \Cref{weyl}) and our assumption on $c_3$,
\begin{equation}
    \notag
    \begin{aligned}
    \sigma_{\min}\left(\mV_{\mU_{t} \mW_{t}}^{\top} \mX \mX^{\top} \mV_{\mU_{t} \mW_{t}}\right) 
    &\geq \sigma_{\min}\left( \mV_{\mU_{t} \mW_{t}}^{\top} \mX_s \mX_s^{\top} \mV_{\mU_{t} \mW_{t}}\right) - \left\|\mV_{\mU_{t} \mW_{t}}^{\top} \mX_{s,\perp} \mX_{s,\perp}^{\top} \mV_{\mU_{t} \mW_{t}}\right\|^2 \\
    &\geq \sigma_{\min}\left( \mV_{\mU_{t} \mW_{t}}^{\top} \mX_s \mX_s^{\top} \mV_{\mU_{t} \mW_{t}}\right) - \sigma_{s+1}^2\left\| \mV_{\mX_s,\perp}^{\top} \mV_{\mU_t \mW_t}\right\|^2 \\
    &\geq \sigma_s^2 \left\|\mV_{\mU_{t} \mW_{t}}^{\top} \mV_{\mX_s}\right\|^2 - \sigma_{s+1}^2 c_3^2 \\
    &= \sigma_s^2 - (\sigma_s^2+\sigma_{s+1}^2)c_3^2 > \frac{1}{2}\left(\sigma_s^2+\sigma_{s+1}^2\right).
    \end{aligned}
\end{equation}
So we have
\begin{equation}
    \notag
    \begin{aligned}
    \|\mR_1\| &\leq \left( 1-\frac{\mu}{2}(\sigma_s^2-\sigma_{s+1}^2)\right)\left\|\mV_{\mX_s,\perp}^{\top} \mV_{\mU_{t} \mW_{t}}\right\| + \mu c_3 c_4^2 + \mu^2 \|\mX\|^4.
    \end{aligned}
\end{equation}
It thus follows from \Cref{noise-5} that
\begin{equation}
    \notag
    \begin{aligned}
    &\quad \left\|\mV_{\mX_{s}^{\perp}}^{\top} \mV_{\mU_{t+1} \mW_{t+1}}\right\| \\
    &\leq \|\mR_1\| + 2\|\mB-\mB_1\| + 102\|\mB\|^2 \\
    &\leq \left( 1-\frac{\mu}{2}(\sigma_s^2-\sigma_{s+1}^2)\right)\left\|\mV_{\mX_s,\perp}^{\top} \mV_{\mU_{t} \mW_{t}}\right\| + 40\mu c_3 c_4 \|\mX\| + 0.02\mu\kappa^{-1}c_3\|\mX\|^2 + 10^3\mu^2\|\mX\|^4.
    \end{aligned}
\end{equation}
Since $\left\|\mV_{\mX_s,\perp}^{\top} \mV_{\mU_{t} \mW_{t}}\right\| \leq c_3$, it follows from our assumption on $c_3,c_4$ and $\mu$ that $\left\|\mV_{\mX_{s}^{\perp}}^{\top} \mV_{\mU_{t+1} \mW_{t+1}}\right\| \leq c_3$ as well, which concludes the proof.
\end{proof}

\subsection{Induction}
\label{subsec:induct}
Let
\begin{equation}
    \notag
    T_{\alpha,s}^{\tpi}=\min \left\{t \geqslant 0: \sigma_{\min }^2\left(\mV_{\mX_{s}}^{\top} \mU_{\alpha,t+1}\right)>0.3\kappa^{-1}\|\mX\|^2\right\}.
\end{equation}
where $\mathtt{pi}$ stands for the parallel improvement phase.
In this section, we show that when $T_{\alpha}^{\sp} \leq t < T_{\alpha,s}^{\tpi}$, the parallel component grows exponentially faster than the orthogonal component. We prove this via induction and the base case is already shown in \Cref{spectral-end}.

\begin{lemma}[\Cref{main:induction-lemma}, detailed version]
\label{induction-lemma}
Suppose that \Cref{asmp:rip,asmp-full-rank,asmp:step-size} hold and let $c_3 = 10^4\kappa\sqrt{r_*}\delta, c_4 \leq 10^{-3}\kappa^{-1}\|\mX\|$.
Then the following holds for all $T_{\alpha}^{\sp} \leq t < T_{\alpha,s}^{\tpi}$ as long as $\alpha \leq C_4(\mX,\bar{\mU}) =\left(\kappa\frac{C_2(\mX,\bar{\mU})^2}{C_3(\mX,\bar{\mU})^2}\right)^{-2\kappa}$ is sufficiently small:
\begin{subequations}
\label{induction-eq}
\begin{align}
    \sigma_{\min }\left(\mV_{\mX_{s}}^{\top} \mU_{t+1}\right)
    &\geq \sigma_{\min }\left(\mV_{\mX_{s}}^{\top} \mU_{t+1}\mW_{t}\right)
    \geq \left(1+0.5 \mu\left(\sigma_{s}^{2}+\sigma_{s+1}^{2}\right)\right) \sigma_{\min }\left(\mV_{\mX_{s}}^{\top} \mU_{\alpha,t}\right) \label{induction-1} \\
    \left\|\mU_{t+1} \mW_{t+1, \perp}\right\| &\leq \min\left\{\left(1+\mu \left(0.4 \sigma_s^2+0.6 \sigma_{s+1}^{2}\right)\right)\left\|\mU_{t} \mW_{t,\perp}\right\|,c_4\right\}  \label{induction-2}\\
    \left\|\mV_{\mX_s,\perp}^{\top} \mV_{\mU_{t+1}\mW_{t+1}}\right\| &\leq c_3.\label{induction-3} \\
    \rank{\mV_{\mX_{s}}^{\top} \mU_{t+1}} &= \rank{\mV_{\mX_{s}}^{\top} \mU_{t+1}\mW_{t}} = s.\label{induction-4}
\end{align}
\end{subequations}
\end{lemma}

\begin{proof}
The base case $t = T_{\alpha,s}^{\tpi}$ is already proved in \Cref{induction-base}. Now suppose that the lemma holds for $t$, we now show that it holds for $t+1$ as well.

To begin with, we bound the term $\left\|\mDelta_t\right\|$ as follows:
\begin{equation}
    \label{delta-t-bound}
    \begin{aligned}
        \norm{\mDelta_t} &=
        \left\|(\A^*\A-\mI)(\mX\mX^{\top}-\mU_{t}\mU_{t}^{\top})\right\|\\
        &\leq \left\|(\A^*\A-\mI)(\mX\mX^{\top}-\mU_{t}\mW_{t}\mW_{t}^{\top}\mU_{t}^{\top})\right\|+\left\|(\A^*\A-\mI) \mU_{t}\mW_{t,\perp}\mW_{t,\perp}^{\top}\mU_{t}^{\top}\right\|\\
        &\leq 10\delta\sqrt{r_*}\|\mX\|^2 + \delta\left\|\mU_{t}\mW_{t,\perp}\mW_{t,\perp}^{\top}\mU_{t}^{\top}\right\|_{*}\\
        &\leq 10\delta\sqrt{r_*}\|\mX\|^2 + \delta d\left\|\mU_t\mW_{t,\perp}\right\|^2
    \end{aligned}
\end{equation}
where in the second inequality we use \Cref{rip-prop-1,rip-cor} and in the third inequality we use $\|\mA\|_{*}\leq \sqrt{d}\|\mA\|, \forall \mA\in\R^{d\times d}$ and the induction hypotheses.

By induction hypothesis, there exists a constant $\hat{C}_4(\mX,\bar{\mU})=\frac{C_2(\mX,\bar{\mU})}{C_3(\mX,\bar{\mU})}$ (see \Cref{spectral-end}) such that
\begin{equation}
    \label{induction-eq1}
    \begin{aligned}
    \frac{\sigma_{\min }\left(\mV_{\mX_{s}}^{\top} \mU_{t}\right)}{\left\|\mU_{t} \mW_{t, \perp}\right\|} \geq \frac{\sigma_{\min }\left(\mV_{\mX_{s}}^{\top} \mU_{T_{\alpha}^{\sp}}\right)}{\left\|\mU_{T_{\alpha}^{\sp}} \mW_{T_{\alpha}^{\sp}, \perp}\right\|} \geq \hat{C}_4\cdot \alpha^{-\gamma_s}
    \end{aligned}
\end{equation}
where
\begin{equation}
    \notag
    \gamma_{s}=\frac{2\left(\log \left(1+\mu \hat{\sigma}_{s}^{2}\right)-\log \left(1+\mu \hat{\sigma}_{s+1}^{2}\right)\right)}{3 \log \left(1+\mu \hat{\sigma}_{1}^{2}\right)-\log \left(1+\mu \hat{\sigma}_{s+1}^{2}\right)}\geq \frac{1}{4\kappa}.
\end{equation}
Since we must have $\sigma_{\min}^2\left(\mV_{\mX_{s}}^{\top} \mU_{t}\right) \leq
0.3\kappa^{-1}\|\mX\|^2$ by definition of $T_{\alpha,s}^{\tpi}$, it follows that
$\left\|\mU_{t}\mW_{t,\perp}\right\|^2\leq
10\kappa\|\mX\|^2 \hat{C}_4^{2}\alpha^{\frac{1}{2\kappa}}$, so for 
$\alpha \leq (\hat{C}_4^{-2}\kappa)^{-2\kappa}$, $\left\|\mDelta_t\right\| \leq
11\delta\sqrt{r_*}\|\mX\|^2$ holds.

The above inequality combined with our assumption on $\delta$ implies that the
conditions on $\normsm{\mDelta_t}$
in \Cref{signal-main,noise,noise-lemma-2} hold. We now show that
\Cref{induction-1,induction-2,induction-3,induction-4} hold for $t+1$, which
completes the induction step.

First, since $t < T_{\alpha,s}^{\tpi}$, we have $\sigma_{\min}\left( \mV_{\mX_s}^{\top} \mU_{t+1}\right) \leq \kappa^{-1}\|\mX\|^2$. Moreover, the induction hypothesis implies that $\left\|\mV_{\mX_s,\perp}^{\top} \mV_{\mU_{t-1}\mW_{t-1}}\right\| \leq c_3$ and that $\mV_{\mX_s}^{\top}\mU_{\alpha,t}$ is of full rank. Thus the conditions of \Cref{signal-cor1} are all satisfied, and we deduce that \Cref{induction-1} holds.

Second, the assumptions on $c_3,c_4$ and $\delta$, combined with \Cref{noise}, immediately implies 
\begin{equation}
    \notag
    \left\|\mU_{t+1} \mW_{t+1, \perp}\right\| \leq \left(1+\mu \left(0.4 \sigma_s^2+0.6 \sigma_{s+1}^{2}\right)\right)\left\|\mU_{t} \mW_{t, \perp}\right\|.
\end{equation}
As a result, similar to \Cref{induction-eq1} we observe that 
\begin{equation}
    \notag
    \frac{\sigma_{\min }\left(\mV_{\mX_{s}}^{\top} \mU_{t+1}\right)}{\left\|\mU_{t+1} \mW_{t+1, \perp}\right\|} \geq \frac{\sigma_{\min }\left(\mV_{\mX_{s}}^{\top} \mU_{T_{\alpha}^{\sp}}\right)}{\left\|\mU_{T_{\alpha}^{\sp}} \mW_{T_{\alpha}^{\sp}, \perp}\right\|} \geq \hat{C}_4\cdot \alpha^{-\frac{1}{4\kappa}}.
\end{equation}

Since $\sigma_{\min }\left(\mV_{\mX_{s}}^{\top} \mU_{t+1}\right) \leq \|\mX\|$, when $\alpha$ is sufficiently small we must have that $\left\|\mU_{t+1} \mW_{t+1, \perp}\right\| \leq c_4$.

Finally, \Cref{noise-lemma-2} implies that \Cref{induction-3} is true, and \Cref{induction-4} follows from our application of \Cref{signal-main}. This concludes the proof.
\end{proof}

\subsection{The refinement phase and concluding the proof of \Cref{main-lemma}}
\label{subsec:refine}
We have shown that the parallel component $\sigma_{\min}\left( \mV_{\mX_s}^{\top} \mU_{t+1}\right)$ grows exponentially faster than the orthogonal component $\left\|\mU_t \mW_{t,\perp}\right\|$. In this section, we characterize the GD dynamics \textit{after} $T_{\alpha,s}^{\tpi}$. We begin with the following lemma, which is straightforward from the proof of \Cref{induction-lemma}.

\begin{lemma}[\Cref{main:refinement-start}, formal version]
\label{refinement-start}
Under the conditions of \Cref{main:induction-lemma}, the following inequality holds when $\alpha \leq C_4(\mX,\bar{\mU})$:
\begin{equation}
    \notag
    \left\| \mU_{T_{\alpha,s}^{\tpi}} \mW_{T_{\alpha,s}^{\tpi},\perp}\right\| \leq C_5(\mX,\bar{\mU})\cdot \alpha^{\frac{1}{4\kappa}}
\end{equation}
where $C_5 = \sqrt{10\kappa}\|\mX\|\frac{C_2(\mX,\bar{\mU})}{C_3(\mX,\bar{\mU})}$.
\end{lemma}
The following lemma states that in a certain time period after $T_{\alpha,s}^{\tpi}$, the parallel and orthogonal components still behave similarly to the second (parallel improvement) phase.

\begin{lemma}
\label{continue-induct}
    Under the conditions in \Cref{induction-lemma}, there exists $\widetilde{t}_{\alpha,s} \geq \frac{1}{\log\left(1+\mu\sigma_s^2\right)}\log\left(\frac{10^{-4}c_3\|\mX\|^2}{\sqrt{d}\kappa C_5}\alpha^{-\frac{1}{4\kappa}}\right) = \Theta\left(\log\alpha^{-1}\right)$ when $\alpha\to 0$ such that when $0\leq t-T_{\alpha,s}^{\tpi} \leq \widetilde{t}_{\alpha,s}$, we have
    \begin{subequations}
        \label{refine-induct-eq}
        \begin{align}
            \sigma_{\min}\left(\mV_{\mX_s}^{\top}\mU_t\right) &\geq \sigma_{\min}\left(\mU_t\mW_t\right) \geq 0.3\kappa^{-1}\|\mX\|^2, \label{refine-induct-eq1} \\
            \left\|\mU_{t}\mW_t\right\| &\leq \left( 1+ \mu(0.4\sigma_s^2+0.6\sigma_{s+1}^2)\right)^{t-T_{\alpha,s}^{\tpi}}\left\|\mU_{T_{\alpha,s}^{\tpi}}\mW_{T_{\alpha,s}^{\tpi}}\right\|, \label{refine-induct-eq2} \\
            \left\|\mV_{\mX_s,\perp}\mV_{\mU_t\mW_t}\right\| &\leq c_3. \label{refine-induct-eq3}
        \end{align}
    \end{subequations}
\end{lemma}

\begin{proof}
    We choose 
    \begin{equation}
        \label{continue-induct-eq0}
        \widetilde{t}_{\alpha,s} = \min\left\{ t\geq 0: \| \mU_{t+1}\mW_{t+1,\perp}\|^2 \leq c_5\right\}
    \end{equation}
    where 
    \begin{equation}
        \label{def-c5}
        c_5= 10^{-4}d^{-\frac{1}{2}}\kappa^{-1}c_3\|\mX\|^2
    \end{equation}
    We prove \Cref{refine-induct-eq} by induction. The proof follows the idea of
    \Cref{induction-lemma}, except that we need to bound
    $\normsm{\mDelta_t}$ in
    each induction step. Concretely, suppose that \Cref{refine-induct-eq} holds
    at time $t$, then
    \begin{equation}
        \label{refine-induct-rip-error}
        \begin{aligned}
            \normsm{\mDelta_t} 
            &= \left\|(\A^*\A-\mI)(\mX\mX^{\top}-\mU_{t}\mU_{t}^{\top})\right\|\\
            &\leq \left\|(\A^*\A-\mI)(\mX\mX^{\top}-\mU_{t}\mW_{t}\mW_{t}^{\top}\mU_{t}^{\top})\right\|+\left\|(\A^*\A-\mI) \mU_{t}\mW_{t,\perp}\mW_{t,\perp}^{\top}\mU_{t}^{\top}\right\|\\
            &\leq 10\delta\sqrt{r_*}\|\mX\|^2 + \delta\left\|\mU_{t}\mW_{t,\perp}\mW_{t,\perp}^{\top}\mU_{t}^{\top}\right\|_{*}\\
            &\leq 10\delta\sqrt{r_*}\|\mX\|^2 + \delta c_5 \sqrt{d} \leq 0.02\kappa^{-1}c_3\|\mX\|^2
        \end{aligned}
    \end{equation}
    where we used the definition of $c_5$ in the last step. As a result, we can apply the conclusion of \Cref{signal-main,noise,noise-lemma-2} which implies that \Cref{refine-induct-eq} holds for $t+1$. Finally, combining \Cref{refinement-start} and \Cref{refine-induct-eq2} yields $\widetilde{t}_{\alpha,s}=\Theta\left(\log\frac{1}{\alpha}\right)$.
\end{proof}

We now present the main result of this section:

\begin{lemma}
\label{refinement-main}
Suppose that $0\leq t-T_{\alpha,s}^{\pi}\leq \widetilde{t}_{\alpha,s}$, $\left\| \mV_{\mX_s,\perp}^{\top} \mV_{\mU_t \mW_t}\right\| \leq c_3$ and the conditions in \Cref{induction-lemma} hold, then we have
\begin{equation}
    \notag
    \begin{aligned}
    &\quad \left\| \mV_{\mX_s}^{\top}( \mX\mX^{\top}-\mU_{t+1}\mU_{t+1}^{\top})\right\|_F \\
    &\leq \left( 1-\frac{1}{2}\mu\tau\right)\left\| \mV_{\mX_s}^{\top}\left(\mX \mX^{\top}-\mU_{t} \mU_{t}^{\top}\right)\right\|_F + 20\mu\|\mX\|^4\left( \delta+5c_3\right) + 2000\mu^2\sqrt{r_*}\|\mX\|^6.
    \end{aligned}
\end{equation}
where we recall that $\tau = \min_{1\leq s\leq \hr\wedge r_*}(\sigma_s^2-\sigma_{s+1}^2)>0$.
\end{lemma}

\begin{proof}

Recall that $\mM_t = \A^*\A\left(\mX\mX^{\top}-\mU_t\mU_t^{\top}\right)$. The update of GD implies that
\begin{equation}
    \notag
    \begin{aligned}
    &\quad \mX \mX^{\top} - \mU_{t+1}\mU_{t+1}^{\top} \\
    &= \mX \mX^{\top} - (\mI+\mu \mM_t)\mU_t \mU_t^{\top} (\mI+\mu \mM_t) \\
    &=\underbrace{\left(\mI-\mu \mU_{t} \mU_{t}^{\top}\right)\left(\mX \mX^{\top}-\mU_{t} \mU_{t}^{\top}\right)\left(\mI-\mu \mU_{t} \mU_{t}^{\top}\right)}_{=(i)}+\mu \underbrace{\mDelta_t \mU_{t} \mU_{t}^{\top}}_{=(ii)} \\
    &\quad +\underbrace{\mu \mU_{t} \mU_{t}^{\top} \mDelta_t}_{=(iii)} + \mu^2 \left(\mathcal{E}_{t,1}+\mathcal{E}_{t,2}\right),
    \end{aligned}
\end{equation}
where $\mathcal{E}_{t,1}=-\mU_{t} \mU_{t}^{\top}\left(\mX \mX^{\top}-\mU_{t} \mU_{t}^{\top}\right) \mU_{t} \mU_{t}^{\top}$ and $\mathcal{E}_{t,2} = -\mM_t\mU_t\mU_t^{\top}\mM_t$. Since $\|\mU_t\|\leq 3\|\mX\|$ by \Cref{bounded-iterate}, and $\left\|\mM_t-(\mX\mX^{\top}-\mU_t\mU_t^{\top})\right\|=\|\mDelta_t\|\leq \|\mX\|^2$ which is shown in \Cref{refine-induct-rip-error}, we have
\begin{equation}
    \notag
    \begin{aligned}
        \left\| \mV_{\mX_s}^{\top}\mathcal{E}_{t,1}\right\|_F 
        &= \left\| \mV_{\mX_s}^{\top} \mU_{t} \mU_{t}^{\top}\left(\mX \mX^{\top}-\mU_{t} \mU_{t}^{\top}\right) \mU_{t} \mU_{t}^{\top} \right\|_F \\
        &\leq \sqrt{r_*} \left\| \mV_{\mX_s}^{\top} \mU_{t} \mU_{t}\left(\mX \mX^{\top}-\mU_{t} \mU_{t}^{\top}\right) \mU_{t} \mU_{t}^{\top} \right\|_2 \\
        &\leq 10^3\sqrt{r_*}\|\mX\|^6
    \end{aligned}
\end{equation}
and
\begin{equation}
    \notag
    \begin{aligned}
        \left\|\mV_{\mX_s}^{\top}\mathcal{E}_{t,2}\right\|_F
        &= \left\|\mV_{\mX_s}^{\top}\left[\left(\mathcal{A}^{*} \mathcal{A}\right)\left(\mX \mX^{\top}-\mU_{t} \mU_{t}^{\top}\right)\right] \mU_{t} \mU_{t}^{\top}\left[\left(\mathcal{A}^{*} \mathcal{A}\right)\left(\mX \mX^{\top}-\mU_{t} \mU_{t}^{\top}\right)\right]\right\|_F \\
        &\leq \sqrt{r_*} \left\|\left[\left(\mathcal{A}^{*} \mathcal{A}\right)\left(\mX \mX^{\top}-\mU_{t} \mU_{t}^{\top}\right)\right] \mU_{t} \mU_{t}^{\top}\left[\left(\mathcal{A}^{*} \mathcal{A}\right)\left(\mX \mX^{\top}-\mU_{t} \mU_{t}^{\top}\right)\right]\right\| \\
        &\leq 10^3\sqrt{r_*}\|\mX\|^6.
    \end{aligned}
\end{equation}
Note that we would like to bound $\left\| \mV_{\mX_s}^{\top}\left( \mX \mX^{\top} - \mU_{t+1}\mU_{t+1}^{\top}\right)\right\|_F$. We deal with the above three terms separately. For the first term, we have
\begin{subequations}
    \label{refine-eq1}
    \begin{align}
    &\quad \left\|\mV_{\mX_s}^{\top}\left(\mI-\mu \mU_{t} \mU_{t}^{\top}\right)\left(\mX \mX^{\top}-\mU_{t} \mU_{t}^{\top}\right)\left(\mI-\mu \mU_{t} \mU_{t}\right)\right\|_F \nonumber \\
    &= \left\|\mV_{\mX_s}^{\top}\left(\mI-\mu \mU_{t} \mU_{t}^{\top}\right) \mV_{\mX_s} \mV_{\mX_s}^{\top}\left(\mX \mX^{\top}-\mU_{t} \mU_{t}^{\top}\right)\left(\mI-\mu \mU_{t} \mU_{t}^{\top}\right)\right\|_F \nonumber \\
    &\quad +\left\|\mV_{\mX_s}^{\top}\left(\mI-\mu \mU_{t} \mU_{t}^{\top}\right) \mV_{\mX_s,\perp} \mV_{\mX_s,\perp}^{\top}\left(\mX \mX^{\top}-\mU_{t} \mU_{t}^{\top}\right)\left(\mI-\mu \mU_{t} \mU_{t}^{\top}\right)\right\|_F \nonumber \\
    &\leq \left\| \mI - \mu \mV_{\mX_s}^{\top} \mU_t \mU_t^{\top} \mV_{\mX_s}\right\|\left\| \mV_{\mX_s}^{\top}\left(\mX \mX^{\top}-\mU_{t} \mU_{t}^{\top}\right)\right\| + \mu \left\| \mV_{\mX_s}^{\top} \mU_t \mU_t^{\top} \mV_{\mX_s,\perp} \mV_{\mX_s,\perp}^{\top}\left(\mX \mX^{\top}-\mU_{t} \mU_{t}^{\top}\right)\right\|_F\label{refine-eq1-1} \\
    &\leq \left( 1-\mu\sigma_{\min}^2(\mU_t \mW_t)\sigma_{\min}^2\left( \mV_{\mX_s}^{\top} \mV_{\mU_t \mW_t}\right)\right)\left\| \mV_{\mX_s}^{\top}\left(\mX \mX^{\top}-\mU_{t} \mU_{t}^{\top}\right)\right\|_F + 100\mu\|\mX\|^4 c_3\label{refine-eq1-2} \\
    &\leq \left( 1-\frac{1}{2}\mu\tau\right)\left\| \mV_{\mX_s}^{\top}\left(\mX \mX^{\top}-\mU_{t} \mU_{t}^{\top}\right)\right\|_F + 100\mu\|\mX\|^4 c_3,\label{refine-eq1-3}
    \end{align}
\end{subequations}
where in \Cref{refine-eq1-1} we use $\left\|\mI-\mu \mU_t \mU_t^{\top}\right\|\leq 1$, \Cref{refine-eq1-2} follows from
\begin{equation}
    \notag
    \begin{aligned}
    \sigma_{\min}\left( \mV_{\mX_s}^{\top} \mU_t \mU_t^{\top} \mV_{\mX_s}\right) &= \sigma_{\min}\left( \mV_{\mX_s}^{\top} \mU_t \mW_t \mW_t^{\top} \mU_t^{\top} \mV_{\mX_s}\right) \geq \sigma_{\min}\left( \mV_{\mX_s}^{\top} \mU_t \mW_t\right)^2 \\
    &\geq \sigma_{\min}^2(\mU_t \mW_t)\sigma_{\min}^2\left( \mV_{\mX_s}^{\top} \mV_{\mU_t \mW_t}\right)
    \end{aligned}
\end{equation}
and 
\begin{equation}
    \notag
    \left\|\mV_{\mX_s}^{\top} \mU_t \mU_t^{\top} \mV_{\mX_s,\perp}\right\| = \left\|\mV_{\mX_s}^{\top} \mU_t \mW_t \mW_t^{\top} \mU_t^{\top} \mV_{\mX_s,\perp}\right\| \leq \left\|\mU_t\right\|^2\left\|\mV_{\mX_s,\perp}^{\top} \mU_t \mW_t\right\| \leq c_3\|\mU_t\|^2,
\end{equation}
and lastly \Cref{refine-eq1-3} is obtained from
\begin{equation}
    \notag
    \sigma_{\min}^2\left( \mV_{\mX_s}^{\top} \mV_{\mU_t \mW_t}\right) \geq 1-\left\|\mV_{\mX_s,\perp}^{\top} \mV_{\mU_t \mW_t}\right\|^2\geq 1-c_3^2.
\end{equation}

For the second and the third terms, we have
\begin{equation}
    \label{refine-2}
    \begin{aligned}
     \left\| \mDelta_t \mU_{t} \mU_{t}^{\top} + \mU_{t} \mU_{t}^{\top}\mDelta_t\right\| 
    \leq 0.1\kappa c_3\|\mX\|^4
    \end{aligned}
\end{equation}
where we use the estimate in \Cref{refine-induct-rip-error}.
Combining \Cref{refine-eq1} and \Cref{refine-2} yields
\begin{equation}
    \notag
    \begin{aligned}
    &\quad \left\| \mV_{\mX_s}^{\top}( \mX\mX^{\top}-\mU_{t+1}\mU_{t+1}^{\top})\right\|_F \\ 
    &\leq \left( 1-\frac{1}{2}\mu\tau\right)\left\| \mV_{\mX_s}^{\top}\left(\mX \mX^{\top}-\mU_{t} \mU_{t}^{\top}\right)\right\|_F + 200\mu\|\mX\|^4 c_3 + 110\mu^2 \sqrt{r_*}\|\mX\|^6.
    \end{aligned}
\end{equation}
\end{proof}

To apply the result of \Cref{main:refinement-main}, we need to verify that $\left\| \mV_{\mX_s,\perp}\mV_{\mU_t \mW_t}\right\|\leq c_3$ still holds when $t \geq T_{\alpha,s}^{\tpi}$. In fact, this is true as long as $t - T_{\alpha,s}^{\tpi} \leq \O\left( \log\frac{1}{\alpha}\right)$.

\begin{lemma}
\label{apply-refinement}
Under the conditions in \Cref{induction-lemma}, if 
\begin{equation}
    \notag
    T_{\alpha,s}^{\tpi} \leq t \leq T_{\alpha,s}^{\tpi} + \frac{\gamma_s\log\frac{c_4}{C_5(\mX,\bar{\mU})}\cdot\log\frac{1}{\alpha}}{\log\left(1+\mu\sigma_s^2\right)} =: T_{\alpha,s}^{\mathtt{re}},
\end{equation}
then $\left\| \mU_{t+1} \mW_{t+1,\perp}\right\|\leq (1+\mu\sigma_s^2)\left\| \mU_t \mW_{t,\perp}\right\|$ and $\left\| \mV_{\mX_s,\perp}^{\top} \mV_{\mU_{t}\mW_{t}}\right\| \leq c_3$. As a consequence, we have $\left\| \mU_t \mW_{t,\perp}\right\|\leq (1+\mu\sigma_s^2)^{t-T_{\alpha,s}^{\tpi}} C_5(\mX,\bar{\mU})\cdot \alpha^{\gamma_s} \leq c_4$.
\end{lemma}

\begin{proof}
The proof is basically the same as that of \Cref{induction-lemma} and we only provide a sketch here.

We induct on $t$. The base case $t = T_{\alpha,s}^{\tpi}$ is already proved in \Cref{induction-lemma}. Suppose that the lemma holds for $t-1$ with $t < T_{\alpha,s}^{\mathtt{re}}$, then the choice of $T_{\alpha,s}^{\mathtt{re}}$ combined with \Cref{noise} imply that
\begin{equation}
    \notag
    \left\| \mU_t \mW_{t,\perp}\right\| \leq \left( 1+\mu\sigma_s^2\right)\left\| \mU_{t-1} \mW_{t-1,\perp}\right\| \leq \left( 1+\mu\sigma_s^2\right)^{t-T_{\alpha,s}^{\tpi}}\left\| \mU_{T_{\alpha,s}^{\tpi}} \mW_{T_{\alpha,s}^{\tpi},\perp}\right\|.
\end{equation}
Since we have $\left\| \mU_{T_{\alpha,s}^{\tpi}} \mW_{T_{\alpha,s}^{\tpi},\perp}\right\| \leq C_5(\mX,\bar{\mU})\cdot\alpha^{\gamma_s}$ by \Cref{refinement-start}, the choice of $T_{\alpha,s}^{\mathtt{re}}$ implies that $\left\| \mU_t \mW_{t,\perp}\right\| \leq c_4$. The bound $\left\| \mV_{\mX_s,\perp}^{\top} \mV_{\mU_t \mW_t}\right\| \leq c_3$ then follows from \Cref{noise-lemma-2}.
\end{proof}

We will only use a weaker version of this lemma, namely that the bounds holds for all $T_{\alpha,s}^{\tpi} \leq t\leq T_{\alpha,s}^{\mathtt{ft}}$. When $\alpha$ is sufficiently small, this can be directly derived from \Cref{refinement-main,apply-refinement} since $\tilde{t}_{\alpha,s}, T_{\alpha,s}^{\mathtt{re}} - T_{\alpha,s}^{\tpi} = \Theta\left(\log\frac{1}{\alpha}\right)$. Specifically, we have proven \Cref{main:refinement-main} in the main text:

\refineMain*

We are now ready to present our first main result, which states that with small initialization, GD would visit the $\O(\delta)$-neighborhood of the rank-$s$ minimizer of the full observation loss i.e. $\mX_s \mX_s^{\top}$.
\begin{theorem}[Restatement of \Cref{main-lemma}]
\label{main-thm-formal}
Under \Cref{asmp:rip,asmp-full-rank,asmp:step-size}, 
if the initialization scale $\alpha$ is sufficiently small,
then for all $1\leq s\leq \hr\wedge r_*$ there exists a time $T_{\alpha,s}^{\mathtt{ft}} \in\mathbb{Z}_{+}$ (where $\mathtt{ft}$ stands for \textit{fitting} the ground-truth) such that
\begin{equation}
    \notag
    \left\|\mX_{s} \mX_{s}^{\top} - \mU_{T_{\alpha,s}^{\tpi}} \mU_{T_{\alpha,s}^{\tpi}}^{\top}\right\|_F \leq 10^7\kappa^3 r_*\|\mX\|^2\delta.
\end{equation}
\end{theorem}

\begin{proof}
First, observe that for all $t\geq 0$,
\begin{equation}
\label{linear-convergence}
    \begin{aligned}
    \left\|\mX_{s} \mX_{s}^{\top}-\mU_{t} \mU_{t}^{\top}\right\|_F & \leq \left\|\left(\mX_{s} \mX_{s}^{\top}-\mU_{t} \mU_{t}^{\top}\right) \mV_{\mX_{s}} \mV_{\mX_{s}}^{\top}\right\|_F +\left\|\mU_{t} \mU_{t}^{\top} \mV_{\mX_{s}^{\perp}} \mV_{\mX_{s}^{\perp}}^{\top}\right\|_F \\
    &\leq \left\|\left(\mX_{s} \mX_{s}^{\top}-\mU_{t} \mU_{t}^{\top}\right) \mV_{\mX_{s}} \mV_{\mX_{s}}^{\top}\right\|_F +\left\|\mV_{\mX_{s}^{\perp}}^{\top} \mU_{t} \mU_{t}^{\top} \mV_{\mX_{s}^{\perp}}\right\|_F \\
    &\leq \left\| \mV_{\mX_{s}}^{\top}\left(\mX_{s} \mX_{s}^{\top}-\mU_{t} \mU_{t}^{\top}\right)\right\|_F + \sqrt{r_*}\left\|\mV_{\mX_{s}^{\perp}}^{\top} \mU_{t} \mW_{t}\right\|^2+\sqrt{d}\left\|\mV_{\mX_{s}^{\perp}}^{\top} \mU_{t} \mW_{t, \perp}\right\|^2 \\
    &\leq \left\| \mV_{\mX_{s}}^{\top}\left(\mX_{s} \mX_{s}^{\top}-\mU_{t} \mU_{t}^{\top}\right)\right\| + 9\sqrt{r_*}\|\mX\|^2 \left\| \mV_{\mX_s,\perp}\mV_{\mU_t \mW_t}\right\|^2 + \sqrt{d}\| \mU_t \mW_{t,\perp}\|^2.
    \end{aligned}
\end{equation}
We set $c_3 = 10^{3}\kappa\sqrt{r_*} \delta$ and
\begin{equation}
    \label{def-hat-T}
    T_{\alpha,s}^{\mathtt{ft}} = T_{\alpha,s}^{\mathtt{pi}} - \frac{\log\left( 10^{-2}\|\mX\|^{-2}\tau c_3^{-1} \right)}{\log\left( 1-\frac{1}{2}\mu\tau\right)},
\end{equation}
where we recall that $\tau=\kappa^{-1}\|\mX\|^2$, then for small $\alpha$ we have $T_{\alpha,s}^{\mathtt{ft}} \leq T_{\alpha,s}^{\mathtt{pi}} + \widetilde{t}_{\alpha,s}$ (defined in \Cref{continue-induct}). Hence for $T_{\alpha,s}^{\mathtt{pi}} \leq t < T_{\alpha,s}^{\mathtt{ft}}$ we always have $\left\|\mV_{\mX_s,\perp}\mV_{\mU_t\mW_t}\right\|\leq c_3$. By \Cref{refinement-main} and the choice of $c_3$ and $\delta$, we have for $T_{\alpha,s}^{\mathtt{pi}} \leq t < T_{\alpha,s}^{\mathtt{ft}}$ that
\begin{equation}
    \notag
    \left\|\mV_{\mX_{s}}^{\top}\left(\mX \mX^{\top}-\mU_{t+1} \mU_{t+1}^{\top}\right)\right\|_F \leq \left(1-\frac{1}{2} \mu \tau\right)\left\|\mV_{\mX_{s}}^{\top}\left(\mX \mX^{\top}-\mU_{t} \mU_{t}^{\top}\right)\right\|_F+30 \mu\|\mX\|^{4}\sqrt{r_*}c_3 
\end{equation}
which implies that for $t = T_{\alpha,s}^{\mathtt{ft}}$,
\begin{equation}
    \notag
    \left\|\mV_{\mX_{s}}^{\top}\left(\mX \mX^{\top}-\mU_{T_s} \mU_{T_s}^{\top}\right)\right\|_F \leq 80\kappa\|\mX\|^2 \sqrt{r_*} c_3.
\end{equation}
Meanwhile, by \Cref{continue-induct} we have $\left\| \mU_t \mW_{t,\perp}\right\| \leq c_5$ ($c_5$ is defined in \Cref{def-c5}) and $\left\| \mV_{\mX_s,\perp}^{\top} \mV_{\mU_{t}\mW_{t}}\right\| \leq c_3$ at $t = T_{\alpha,s}^{\mathtt{ft}}$. Plugging into \Cref{linear-convergence} yields
\begin{equation}
    \notag
    \left\| \mX_s \mX_s^{\top} - \mU_{T_{\alpha,s}^{\mathtt{ft}}}\mU_{T_{\alpha,s}^{\mathtt{ft}}}^{\top}\right\|_F \leq 80\kappa\|\mX\|^2\sqrt{r_*} c_3 + 9\|\mX\|^2 c_3^2\sqrt{r_*} + c_5^2\sqrt{d}.
\end{equation}
By definition of $c_3$ and $c_5$ we deduce that $\left\| \mX_s \mX_s^{\top} - \mU_{T_{\alpha,s}^{\mathtt{ft}}}\mU_{T_{\alpha,s}^{\mathtt{ft}}}^{\top}\right\|_F \leq 10^{2}\tau^{-2}\|\mX\|^6\sqrt{r_*}c_3 \leq 10^5\kappa^3 r_*\|\mX\|^2\delta$, as desired.
\end{proof}

\begin{corollary}
    \label{cor-C6}
    There exists a constant 
    \begin{equation}
        \label{def-C6}
        C_6(\mX,\bar{\mU}) = C_5(\mX,\bar{\mU})\cdot (1+\mu\sigma_s^2)^{T_{\alpha,s}^{\mathtt{ft}}-T_{\alpha,s}^{\mathtt{pi}}}
    \end{equation}
    such that
    \begin{equation}
        \notag
        \max_{0\leq t\leq T_{\alpha,s}^{\mathtt{ft}}}\left\|\mU_t\mW_{t,\perp}\right\| \leq C_1\cdot\alpha^{\frac{1}{4\kappa}}.
    \end{equation}
\end{corollary}

\begin{proof}
    The case of $t \leq T_{\alpha,s}^{\mathtt{pi}}$ directly follows from \Cref{refinement-start}. For $t > T_{\alpha,s}^{\mathtt{pi}}$, we know from \Cref{continue-induct} that
    \begin{equation}
        \notag
        \left\|\mU_t\mW_{t,\perp}\right\| \leq \left\|\mU_{T_{\alpha,s}^{\mathtt{pi}}}\mW_{T_{\alpha,s}^{\mathtt{pi}},\perp}\right\|\cdot \left(1+\mu\sigma_s^2\right)^{T_{\alpha,s}^{\mathtt{ft}}-T_{\alpha,s}^{\mathtt{pi}}}.
    \end{equation}
    By \Cref{def-hat-T}, the second term is a constant independent of $\alpha$, so the conclusion follows.
\end{proof}

\section{Auxiliary results for proving \Cref{main-lemma}}
This section contains a collection of auxiliary results that are used in the previous section.
\label{appsec-aux}
\subsection{The spectral phase}
In the section, we provide auxiliary results for the analysis in the spectral phase.

Recall that $\mK_{t} = (\mI+\mu \mM)^t$ and $\mU_t = \mU_t^{\sp} + \mE_t = \mK_{t} \mU_0 + \mE_t$ and $\mU_0 = \alpha \bar{\mU}$ with $\|\bar{\mU}\|=1$. Also recall that $\mM = \sum_{i=1}^{\rank{\mM}}\hat{\lambda}_i\hat{\vv}_i\hat{\vv}_i^{\top}$; we additionally define $\mM_s = \sum_{i=1}^{\min\left\{s,\rank{\mM}\right\}}\hat{\lambda}_i\hat{\vv}_i\hat{\vv}_i^{\top}$. Similarly, let $\mL_t$ be the span of the top-$s$ left singular vectors of $\mU_t$. The following lemma shows that power iteration would result in large eigengap of $\mU_t$.

\begin{lemma}
\label{appendix-spectral-lma1}
Let $\hat{\rho} = \sigma_{\min }\left(\mV_{\mM_s}^{\top} \bar{\mU}\right) > 0$, then the following three inequalities hold, given that the denominator of the third is positive.
\begin{subequations}
    \label{appendix-spectral-1}
    \begin{align}
    \sigma_s(\mU_t) &\geq   \alpha\left(\hat{\rho} \sigma_{s}\left(\hat{\mZ}_{t}\right)-\sigma_{s+1}\left(\hat{\mZ}_{t}\right) \right) -\left\|\mE_{t}\right\|,\label{appendix-spectral-1-1} \\
    \sigma_{s+1}(\mU_t) &\leq \alpha \sigma_{s+1}\left(\hat{\mZ}_{t}\right)+\left\|\mE_{t}\right\|,\label{appendix-spectral-1-2} \\
    \left\|\mV_{\mM_s^{\perp}}^{\top} \mV_{L_{t}}\right\| & \leq \frac{\alpha \sigma_{s+1}\left(\hat{\mZ}_{t}\right)+\left\|\mE_{t}\right\|}{\alpha\hat{\rho} \sigma_{s}\left(\hat{\mZ}_{t}\right) -2\left(\alpha \sigma_{s+1}\left(\hat{\mZ}_{t}\right)+\left\|\mE_{t}\right\|\right)} .\label{appendix-spectral-1-3}
    \end{align}
\end{subequations}
\end{lemma}

\begin{proof}
By Weyl's inequality we have
\begin{equation}
    \notag
    \begin{aligned}
        \sigma_{s+1}(\mU_t) &= \sigma_{s+1}\left( (1+\mu \mM)^t \mU_0\right) + \|\mE_t\| \\
        &= \alpha \sigma_{s+1}\left( (1+\mu \mM)^t \bar{\mU}\right) + \|\mE_t\| \\
        &\leq \alpha\sigma_{s+1}\left( (1+\mu \mM_s)^t \bar{\mU}\right) + \alpha\left\| \left[ (1+\mu \mM)^t-(1+\mu \mM_s)^t\right] \bar{\mU}\right\|+\|\mE_t\| \\
        &\leq \alpha (1+\mu\hat{\lambda}_{s+1})^t + \|\mE_t\|.
    \end{aligned}
\end{equation}
Thus \Cref{appendix-spectral-1-2} holds. Similarly,
\begin{equation}
    \notag
    \begin{aligned}
        \sigma_s(\mU_t) &\geq \alpha\sigma_s\left( \mN_{t} \mV_{\mM_s}\mV_{\mM_s}^{\top} \bar{\mU}\right) - \alpha (1+\mu\hat{\lambda}_{s+1})^t  - \|\mE_t\| \\
        &\geq \alpha \sigma_s\left( \mN_{t} \mV_{\mM_s}\right)\sigma_{\min}\left( \mV_{\mM_s}^{\top} \bar{\mU}\right) - \alpha (1+\mu\hat{\lambda}_{s+1})^t  - \|\mE_t\| \\
        &\geq \alpha\hat{\rho} (1+\mu\hat{\lambda}_{s})^t - \alpha (1+\mu\hat{\lambda}_{s+1})^t  - \|\mE_t\|.
    \end{aligned}
\end{equation}
Finally, note that we can write
\begin{equation}
    \notag
    \alpha(1+\mu \mM_s)^t \bar{\mU} = \mV_{\mM_s}\underbrace{(1+\mu\mSigma_{\mM_s})^t \mV_{\mM_s}^{\top} \bar{\mU}}_{\text{invertible}},
\end{equation}
so that the subspace spanned by the left singular vectors of $\alpha(1+\mu \mM_s)^t \bar{\mU}$ coincides with the column span of $\mV_{\mM_s}$. Since $L_t$ is the span of top-$s$ left singular vectors of $\mU_t$, we apply Wedin's sin theorem \citep{wedin1972perturbation} and deduce \Cref{appendix-spectral-1-3}.
\end{proof}

The next lemma relates the quantities studied in \Cref{appendix-spectral-lma1} with those that are needed in the induction. The proof is the same as \citealp[Lemma 8.4]{stoger2021small}, so we omit it here.

\begin{lemma}
Suppose that $\left\|\mV_{\mX_s,\perp}^{\top} \mV_{\mL_{t}}\right\| \leq 0.1$ for some $t \geq 1$. Then it holds that
\begin{subequations}
    \label{appendix-spectral-2}
    \begin{align}
    \sigma_{s}\left(\mU_{t} \mW_{t}\right) & \geq \frac{1}{2} \sigma_{s}\left(\mU_{t}\right),\label{appendix-spectral-2-1}\\
    \left\|\mV_{\mX_s,\perp}^{\top} \mV_{\mU_{t} \mW_{t}}\right\| & \leq 10\left\|\mV_{\mX_s,\perp}^{\top} \mV_{\mL_{t}}\right\|, \label{appendix-spectral-2-2}\\
    \left\|\mU_{t} \mW_{t, \perp}\right\| & \leq 2 \sigma_{s+1}\left(\mU_{t}\right) .\label{appendix-spectral-2-3}
    \end{align}
\end{subequations}
\end{lemma}

Combining the above two lemmas, we directly obtain the following corollary:

\begin{corollary}
\label{signal-noise-separation}
Suppose that $\alpha\sigma_s(\mK_{t}) > 10\left( \alpha\sigma_{s+1}(\mK_{t})+\|\mE_t\|\right)$, then we have that
\begin{equation}
\begin{aligned}
    \sigma_{s}\left(\mU_{t} \mW_{t}\right) & \geq 0.4\alpha \sigma_{r_{\star}}\left(\mK_{t}\right) \sigma_{\min }\left(\mV_{\mL}^{\top} \bar{\mU}\right) \\
    \left\|\mU_{t} \mW_{t, \perp}\right\| & \leq 2\left(\alpha \sigma_{s+1}\left(\mK_{t}\right)+\left\|\mE_{t}\right\|\right) \\
    \left\|\mV_{\mX_s,\perp}^{\top} \mV_{\mU_{t} \mW_{t}}\right\| & \leq 100\left( \delta + \frac{\alpha \sigma_{s+1}\left(\mK_{t}\right)+\left\|\mE_{t}\right\|}{\alpha\hat{\rho} \sigma_{s}\left(\mK_{t}\right)}\right)
\end{aligned}
\end{equation}
\end{corollary}

\subsection{The parallel improvement phase}

In the section, we provide auxiliary results for the analysis in the parallel improvement phase.

\begin{lemma}
\label{bounded-iterate}
(\citealp[Lemma 9.4]{stoger2021small}) For sufficiently small $\mu$ and $\delta$, suppose that $\|\mU_t\|\leq 3\|\mX\|$, then we also have $\|\mU_{t+1}\| \leq 3\|\mX\|$.
\end{lemma}

\begin{lemma}
\label{appendix-saddle-2}
Under the assumptions in \Cref{noise}, we have
\begin{equation}
    \notag
    \left\|\mV_{\mX_s,\perp}^{\top} \mV_{\mU_{t+1} \mW_{t}}\right\| \leq 2\left( c_3 + 10\mu\|\mX\|^2\right) \leq 0.01.
\end{equation}
\end{lemma}

\begin{proof}
The proof of this lemma is essentially the same as \citealp[Lemma B.1]{stoger2021small}, and we omit it here.
\end{proof}

\begin{lemma}
\label{appendix-saddle-1}
Under the assumptions in \Cref{noise-lemma-2}, we have
\begin{equation}
    \notag
    \sigma_{\min}\left(\mV_{\mX_s}^{\top} \mU_{t+1}\right) \geq \frac{1}{2}\sigma_{\min} (\mU_t \mW_t).
\end{equation}
\end{lemma}

\begin{proof}
We have
\begin{equation}
    \notag
    \begin{aligned}
    &\quad \sigma_{\min}\left(\mV_{\mX_s}^{\top} \mU_{t+1}\right) \geq \sigma_{\min}\left(\mV_{\mX_s}^{\top} \mU_{t+1} \mW_t\right) \\
    &= \sigma_{\min}\left(\mV_{\mX_s}^{\top} \left( \mI + \mu \mM_t \right) \mU_t \mW_t\right) \\
    &\geq \sigma_{\min}\left( \mV_{\mX_s}^{\top} \left( \mI + \mu \mM_t \right) \mV_{\mU_t \mW_t}\right)\cdot \sigma_{\min}\left( \mV_{\mU_t \mW_t}^{\top} \mU_t \mW_t\right) \\
    &\geq \left[ \sigma_{\min}\left( \mV_{\mX_s}^{\top}  \mV_{\mU_t \mW_t}\right) - \mu\|\mM_t\|\right]\cdot \sigma_{\min}(\mU_t \mW_t)\\
    &\geq \left(\sqrt{1-c_3^2}-10\mu\|\mX\|^2\right)\sigma_{\min}(\mU_t \mW_t) \geq \frac{1}{2}\sigma_{\min} (\mU_t \mW_t)
    \end{aligned}
\end{equation}
where the last step follows from 
\begin{equation}
    \notag
    \sigma_{\min}\left( \mV_{\mX_s}^{\top}  \mV_{\mU_t \mW_t}\right)^2 \geq 1-\left\| \mV_{\mX_s,\perp}^{\top}  \mV_{\mU_t \mW_t}\right\|^2 \geq 1-c_3^2.
\end{equation}
The conclusion follows.
\end{proof}

\begin{lemma}
\label{row-space-change}
Under the assumptions in \Cref{noise-lemma-2}, we have
\begin{equation}
    \notag
    \left\|\mW_{t, \perp}^{\top} \mW_{t+1}\right\| \leq 3\mu\left(10\mu\|\mX\|^2+ c_4\right)c_3\|\mX\| + \mu\left\|(\A^*\A-\mI)(\mX\mX^{\top}-\mU_t\mU_t^{\top})\right\|.
\end{equation}
\end{lemma}

\begin{proof}
The proof roughly follows [\citealp[Lemma B.3]{stoger2021small}], but we include it here for completeness. 

Since $\mV_{\mX_s}^{\top} \mU_{t+1} = \mV_{t+1}\mSigma_{t+1}\mW_{t+1}$ and $\mV_{t+1}\mSigma_{t+1}\in\R^{s\times s}$ is invertible, we have
\begin{equation}
    \notag
    \left\|\mW_{t,\perp}^{\top} \mW_{t+1}\right\|=\left\|\mW_{t,\perp}^{\top} \mU_{t+1}^{\top} \mV_{\mX_s}\left(\mV_{\mX_s}^{\top} \mU_{t+1} \mU_{t+1}^{\top} \mV_{\mX_s}\right)^{-\frac{1}{2}}\right\|.
\end{equation}
Since
\begin{subequations}
    \label{app-improve-1}
    \begin{align}
    &\quad \mV_{\mX_s}^{\top} \mU_{t+1} \mW_{t,\perp}\nonumber \\
    &= \mV_{\mX_s}^{\top}\left( \mI + \mu\A^*\A(\mX\mX^{\top}-\mU_t\mU_t^{\top})\right) \mU_t \mW_{t,\perp}\nonumber\\
    &= \mV_{\mX_s}^{\top}\left( \mI + \mu(\mX\mX^{\top}-\mU_t\mU_t^{\top})\right) \mU_t \mW_{t,\perp} + \mu \mV_{\mX_s}^{\top} \mDelta_t \mU_t \mW_{t,\perp}\nonumber \\
    &= -\mu \mV_{\mX_s}^{\top} \mU_t \mU_t^{\top} \mU_t \mW_{t,\perp} + \mu \mV_{\mX_s}^{\top} \mDelta_t \mU_t \mW_{t,\perp}\label{app-improve-1-1} \\
    &= -\mu \mV_{\mX_s}^{\top} \mU_t \mW_t \mW_t^{\top} \mU_t^{\top} \mU_t \mW_{t,\perp} + \mu \mV_{\mX_s}^{\top} \mDelta_t \mU_t \mW_{t,\perp}\label{app-improve-1-2} \\
    &= -\mu \underbrace{\mV_{\mX_s}^{\top} \mU_t \mW_t \mW_t^{\top} \mU_t^{\top} \mV_{\mX_s,\perp}\mV_{\mX_s,\perp}^{\top} \mU_t \mW_{t,\perp}}_{=:\mK_1} + \mu \underbrace{\mV_{\mX_s}^{\top} \mDelta_t \mU_t \mW_{t,\perp}}_{:=\mK_2}\label{app-improve-1-3}
    \end{align}
\end{subequations}
where \Cref{app-improve-1-1} follows from $\mV_{\mX_s}^{\top} \mX\mX^{\top} \mU_t \mW_{t,\perp} = \mSigma_s \mV_{\mX_s}^{\top} \mU_t \mW_{t,\perp} = 0$, and in \Cref{app-improve-1-2} and \Cref{app-improve-1-3} we use $\mV_{\mX_s}^{\top}\mU_t \mW_{t,\perp}=0$.

For $\mK_1$, note that
\begin{equation}
    \notag
    \begin{aligned}
    &\quad \left\|\left(\mV_{\mX_s}^{\top} \mU_{t+1} \mU_{t+1}^{\top} \mV_{\mX_s}\right)^{-\frac{1}{2}}\mV_{\mX_s}^{\top} \mU_t\right\| \\
    &\leq \left\|\left(\mV_{\mX_s}^{\top} \mU_{t+1} \mU_{t+1}^{\top} \mV_{\mX_s}\right)^{-\frac{1}{2}}\mV_{\mX_s}^{\top} \mU_{t+1}\right\| + \mu \left\|\left(\mV_{\mX_s}^{\top} \mU_{t+1} \mU_{t+1}^{\top} \mV_{\mX_s}\right)^{-\frac{1}{2}}\mV_{\mX_s}^{\top} \A^*\A(\mX\mX^{\top}-\mU_t \mU_t^{\top})\mU_t\right\| \\
    &\leq 1+ 10\mu\|\mX\|^3\sigma_{\min}^{-1}\left(\mV_{\mX_s}^{\top} \mU_{t+1}\right)
    \end{aligned}
\end{equation}
so that
\begin{equation}
    \notag
    \left\|\left(\mV_{\mX_s}^{\top} \mU_{t+1} \mU_{t+1}^{\top} \mV_{\mX_s}\right)^{-\frac{1}{2}}K_1\right\| \leq \left[ 1+10\mu\|\mX\|^3\sigma_{\min}^{-1}\left(\mV_{\mX_s}^{\top} \mU_{t+1}\right)\right]\left\|\mV_{\mX_s,\perp}^{\top}\mU_t \mW_t\right\|\left\|\mU_t \mW_{t,\perp}\right\|.
\end{equation}
Plugging into \Cref{app-improve-1}, we deduce that
\begin{equation}
    \notag
    \begin{aligned}
    &\quad \left\|\mW_{t,\perp}^{\top}\mW_{t+1}\right\| \\
    &\leq 3\mu \left( 1+10\mu\|\mX\|^3\sigma_{\min}^{-1}\left(\mV_{\mX_s}^{\top} \mU_{t+1}\right)\right)\left\|\mV_{\mX_s,\perp}^{\top} \mV_{\mU_t \mW_t}\right\|\|\mX\|\left\|\mU_t \mW_{t,\perp}\right\| \\
    &\quad + \mu\sigma_{\min}^{-1}\left(\mV_{\mX_s}^{\top} \mU_{t+1}\right)\left\|\mU_t\mW_{t,\perp}\right\|\left\|(\A^*\A-\mI)(\mX\mX^{\top}-\mU_t\mU_t^{\top})\right\| \\
    &\leq 3\mu \left( \left\|\mU_t \mW_{t,\perp}\right\|+10\mu\|\mX\|^3\right)\left\|\mV_{\mX_s,\perp}^{\top} \mV_{\mU_t \mW_t}\right\| \|\mX\| \\
    &\quad + \mu\sigma_{\min}^{-1}\left(\mV_{\mX_s}^{\top} \mU_{t+1}\right)\left\|\mU_t\mW_{t,\perp}\right\|\left\|(\A^*\A-\mI)(\mX\mX^{\top}-\mU_t\mU_t^{\top})\right\|\\
    &\leq 3\mu\left(10\mu\|\mX\|^2+ c_4\right)c_3\|\mX\| + \mu\left\|(\A^*\A-\mI)(\mX\mX^{\top}-\mU_t\mU_t^{\top})\right\|.
    \end{aligned}
\end{equation}
where in the last step we use \Cref{appendix-saddle-1} and the induction hypothesis which implies that $\sigma_{\min}(\mU_t \mW_t) \geq \left\|\mU_t \mW_{t,\perp}\right\|$.
\end{proof}

\begin{lemma}
\label{Z-lem}
The matrix $\mH$ defined in the proof of \Cref{noise-lemma-2} satisfies the following:
\begin{equation}
    \notag
    \mH(\mH^{\top} \mH)^{-\frac{1}{2}} = \mV_{\mU_{t} \mW_{t}}+\mB \mV_{\mU_{t} \mW_{t}}-\frac{1}{2}(\mI +\mB) \mV_{\mU_{t} \mW_{t}} \mV_{\mU_{t} \mW_{t}}^{\top}\left(\mB+\mB^{\top}\right) \mV_{\mU_{t} \mW_{t}}-\mD,
\end{equation}
where $\|\mD\| \leq 30\|\mB\|^2$.
\end{lemma}

\begin{proof}
By definition of $\mH$ we have
\begin{equation}
    \notag
    \begin{aligned}
    &\quad \mH(\mH^{\top} \mH)^{-\frac{1}{2}}\\
    &= (\mI +\mu \mM)(\mI +\mP) \mV_{\mU_{t} \mW_{t}}\left(\mV_{\mU_{t} \mW_{t}}^{\top}\left(\mI +\mP^{\top}\right)(\mI +\mu \mM)^{2}(\mI +\mP) \mV_{\mU_{t} \mW_{t}}\right)^{-\frac{1}{2}}\\
    &= (\mI +\mB) \mV_{\mU_{t} \mW_{t}}\left[\mV_{\mU_t \mW_{t}}^{\top}\left(\mI +\mB^{\top}+\mB+\mB^{\top} \mB\right) \mV_{\mU_{t} \mW_{t}}\right]^{-\frac{1}{2}}\\
    &= (\mI +\mB) \mV_{\mU_t \mW_t} \left[ \mI  + \underbrace{\mV_{\mU_t \mW_{t}}^{\top}\left(\mB^{\top}+\mB+\mB^{\top} \mB\right) \mV_{\mU_{t} \mW_{t}}}_{=:\bm{\Theta}}\right]^{-\frac{1}{2}}.
    \end{aligned}
\end{equation}
It follows from \Cref{noise-4} and our assumptions on $c_3$ and $c_4$ that
\begin{equation}
    \notag
    \begin{aligned}
    \|\mB\| &\leq \mu\left\| \mX\mX^{\top}-\mU_t \mU_t^{\top}\right\| + 6\mu\left( c_3 c_4 \|\mX\|+50\|\mX\|^2\delta\right)\\
    &\leq 10\mu\|\mX\|^2 + 6\mu c_3 \left( c_4 + 1\right)\|\mX\| < 0.1
    \end{aligned}
\end{equation}
(note that this step is independent and does not rely on earlier derivations in the proof of \Cref{noise-lemma-2}), so by Taylor's formula, we have
\begin{equation}
    \notag
    \left\|(\mI+\bm{\Theta})^{-\frac{1}{2}}-\mI+\frac{1}{2}\bm{\Theta}\right\| \leq 3\|\bm{\Theta}\|^2.
\end{equation}
Hence,
\begin{equation}
    \notag
    \begin{aligned}
    &\quad \left\|\mH(\mH^{\top} \mH)^{-\frac{1}{2}} -\left( \mV_{\mU_{t} \mW_{t}}+\mB \mV_{\mU_{t} \mW_{t}}-\frac{1}{2}(\mI +\mB) \mV_{\mU_{t} \mW_{t}} \mV_{\mU_{t} \mW_{t}}^{\top}\left(\mB+\mB^{\top}\right) \mV_{\mU_{t} \mW_{t}}\right)\right\| \\
    &= \left\| (\mI +\mB) \mV_{\mU_{t} \mW_{t}} \left(\left( \mI +\bm{\Theta}\right)^{-\frac{1}{2}}-\mI+\frac{1}{2}\bm{\Theta}-\frac{1}{2}\mV_{\mU_t \mW_t}^{\top} \mB^{\top} \mB \mV_{\mU_t \mW_t}\right) \right\| \\
    &\leq \left( 1+\|\mB\|\right) \left( 3\|\bm{\Theta}\|^2 + \frac{1}{2}\|\mB\|^2\right) < 30\|\mB\|^2
    \end{aligned}
\end{equation}
as desired.
\end{proof}

\section{Proofs for the Landscape Results in \Cref{subsec:key-lemma}}
\label{appsec-landscape}
In this section, we study the landscape of under-parameterized matrix sensing problem
\begin{equation}
    \notag
    f_s(\mU) = \frac{1}{2}\left\|\A(\mU\mU^{\top}-\mX\mX^{\top})\right\|_2^2,\quad \mU\in\R^{d\times s}
\end{equation}
Our key result in this
section is \Cref{landscape-main-thm}, which states a local RSI condition for the
matrix sensing loss. Most existing results only study the landscape of
\Cref{matrix-sensing} in the exact- and over-parameterized case.
\citet{zhu2021global} have studied the landscape of under-parameterized matrix
factorization problem, but their main focus is the strict-saddle property of the loss.

\subsection{Analysis of the matrix factorization loss}

When the measurement satisfies the RIP condition, we can expect that the
landscape of $f_s$ looks similar to that of the (under-parameterized) matrix
factorization loss:
\begin{equation}
    \notag
    F_{s}(\mU) = \frac{1}{2}\left\| \mU \mU^{\top} - \mX \mX^{\top} \right\|_F^2, \quad \mU \in \R^{d\times s}
\end{equation}
for some $s<\hr$.
For this reason, we first look into the landscape of $F_s$
before analyzing $f_s$.

Recall that $\mX \mX^{\top} = \sum_{i=1}^{r_*} \sigma_i^2 \vv_i \vv_i^{\top}$. The critical points of $F_{s}(\mU)$ is characterized by the following lemma:
\begin{lemma}
\label{fac-critical-point}
$\mU \in \R^{d\times s}$ is a critical point of $F_{s}(\mU)$ if and only if there exists an orthogonal matrix $\mR \in \R^{s\times s}$, such that all columns of $\mU \mR$ are in $\left\{ \sigma_i v_i: 1\leq i\leq r_*\right\}$.
\end{lemma}

\begin{proof}
    Assume WLOG that $\mX\mX^{\top} = \mathrm{diag}(\sigma_1^2,\sigma_2^2,\cdots,\sigma_r^2,0,\cdots,0)=:\mSigma$.
    Let $\mU$ be a critical point of $F_s$, then we have that $\left(\mU\mU^{\top}-\mX\mX^{\top}\right)\mU=0$. Let $\mW=\mU\mU^{\top}$, then $(\mSigma-\mW)\mW=0$.

    Since $\mW$ is symmetric, so is $\mW^2$, and we obtain that $\mSigma\mW$ is
    also symmetric. It is then easy to see that that if
    $\mSigma=\mathrm{diag}\left(\lambda_1
    \mI_{m_1},\cdots,\lambda_t\mI_{m_t}\right)$ with
    $\lambda_1>\lambda_2>\cdots>\lambda_t\geq 0$, then $\mW$ is also in
    block-diagonal form:
    $\mW=\mathrm{diag}\left(\mW_1,\mW_2,\cdots,\mW_t\right)$ where $\mW_i \in
    \R^{m_i\times m_i}$. For each $1\leq i \leq t$, we then have the equation
    $\left(\lambda_i\mI_{m_i}-\mW_i\right)\mW_i=0$. Hence, there exists an
    orthogonal matrix $\mR_i$ such that $\mR_i^{\top}\mW_i\mR_i$ is a diagonal
    matrix where the diagonal entries are either $0$ or
    $\sqrt{\lambda_i}=\sigma_i$. Let
    $\mR=\mathrm{diag}\left(\mR_1,\mR_2,\cdots,\mR_t\right)$, then
    $\mR^{\top}\mW\mR$ is diagonal and its nonzero diagonal entries form an
    $s$-subset of the multi-set $\{\sigma_i:1\leq i\leq r_*\}$. The conclusion
    follows.
\end{proof}

In the case of $s=1$, the global minimizers of $F_{s}$ are $\pm\sigma_1 \vv_1$,
and we can show that $F_{s}$ is locally strongly convex around these minimizers.
Therefore, we can deduce that $f$ is locally strongly-convex as well. Since our
main focus is on $s>1$, we put these details in \Cref{appendix_rank1}. When
$s>1$, $F_{s}(\mU)$ is not locally strongly-convex due to rotational invariance:
if $\mU$ is a global minimizer, then so is $\mU \mR$ for any orthogonal matrix
$\mR\in\R^{s\times s}$. Instead, we establish a Restricted Secant
Inequality for $F_s$, as shown below.
\begin{lemma}
\label{matrix-factorization-PL}
For $\mU\in\R^{d\times s}$, let $\mR$ be an orthogonal matrix that minimizes
$\left\|\mU-\mX_s \mR\right\|_F$. Suppose that
$\mathrm{dist}(\mU,\mX_s) \leq 0.1\|\mX\|^{-1}\tau$ (where we recall that
$\tau=\min_{s\in[r_*]}\left(\sigma_s^2-\sigma_{s+1}^2\right)$ is the eigengap of
$\mX\mX^{\top}$), then we have
\begin{equation}
    \notag
    \left\langle \nabla F_{s}(\mU), \mU - \mX_s \mR \right\rangle \geq 0.1 \tau \cdot\mathrm{dist}^2(\mU,\mX_s).
\end{equation}
\end{lemma}

\begin{proof}
Assume WLOG that $\mR = \mI$.
Then by \Cref{lm:R-orth},
$\mU^{\top} \mX_s$ is symmetric and positive semi-definite. Let $\mH = \mU - \mX_s$, then
\begin{equation}
    \notag
    \begin{aligned}
    \nabla F_{s}(\mU) &= (\mU\mU^{\top}-\mX\mX^{\top}) \mU \\
    &= \left[ (\mH+\mX_s)(\mH+\mX_s)^{\top} - \mX\mX^{\top}\right] (\mH+\mX_s).
    \end{aligned}
\end{equation}
So we have
\begin{subequations}
    \label{landscape-1}
    \begin{align}
    \left\langle \nabla F_{s}(\mU), \mU - \mX_s\right\rangle
    &= \left\langle \left[ (\mH+\mX_s)(\mH+\mX_s)^{\top} - \mX\mX^{\top}\right] (\mH+\mX_s),\mH\right\rangle\nonumber \\
    &= \tr{\mH^{\top} \left[(\mH+\mX_s)(\mH+\mX_s)^{\top} -\mX \mX^{\top}\right] \mH + \mH^{\top} \left( \mH\mH^{\top} + \mH \mX_s^{\top} + \mX_s \mH^{\top}\right) \mX_s}\nonumber \\
    &\geq -\tr{\mH^{\top} \mX_{s,\perp}\mX_{s,\perp}^{\top} \mH} - 3\|\mX\|\|\mH\|_F^3 + \tr{\mH^{\top} \mH \mX_s^{\top} \mX_s}\label{landscape-1-1} \\
    &\geq \left(\sigma_s^2-\sigma_{s+1}^2\right) \|\mH\|_F^2 - 3\|\mX\|\|\mH\|_F^3 \label{landscape-1-2} \\
    &\geq 0.1\tau\|\mH\|_F^2 \nonumber
    \end{align}
\end{subequations}
where in \Cref{landscape-1-1} we use $\tr{(\mH^{\top} \mX_s)^2}\geq 0$ (since
$\mH^{\top} \mX_s$ is symmetric as noticed in the beginning of the proof), and
\Cref{landscape-1-2} is because of
\begin{equation}
    \notag
    \begin{aligned}
    \tr{\mH^{\top} \mH \mX_s^{\top} \mX_s}  
    &\geq \sigma_{\min}\left(\mX_s^{\top} \mX_s\right)\cdot \tr{\mH^{\top} \mH} = \sigma_s^2 \|\mH\|_F^2
    \end{aligned}
\end{equation}
and
\begin{equation}
    \notag
    \begin{aligned}
    \tr{\mH^{\top} \mX_{s,\perp}\mX_{s,\perp}^{\top} \mH}
    &= \tr{\mH^{\top} \mV_{\mX_s,\perp}\mSigma_{s,\perp}\mV_{\mX_s,\perp}^{\top} \mH}\\
    &\leq \left\|\mSigma_{s,\perp}\right\|\cdot\left\|\mH^{\top} \mV_{\mX_s,\perp}\right\|_F^2 \leq \sigma_{s+1}^2\|\mH\|_F^2.
    \end{aligned}
\end{equation}
\end{proof}

\begin{corollary}
\label{cor:matrix-factorization-PL}
Under the conditions of \Cref{matrix-factorization-PL}, we have $\left\|\nabla F_{s}(\mU)\right\|_F \geq 0.1\tau\mathrm{dist}(\mU,\mX_s)$.
\end{corollary}

\subsection{Analysis of the matrix sensing loss}

The following lemma states that the minimizer of matrix sensing loss is also near-optimal for the matrix factorization loss. 

\begin{lemma}
\label{func-value-close}
Let $\mZ_s^*$ be a best rank-$s$ solution as defined in \Cref{def:best-rank}, then we have
\begin{equation}
    \notag
    \left\|\mZ_s^* - \mX\mX^{\top}\right\|_F^2 \leq \left\|\mX_s\mX_s^{\top}-\mX\mX^{\top}\right\|_F^2 + 10\delta\left\|\mX\mX^{\top}\right\|_F^2.
\end{equation}
\end{lemma}

\begin{proof}
By the RIP property \Cref{def-rip} we have 
\begin{equation}
    \notag
    \begin{aligned}
    \left\| \mX\mX^{\top} - \mZ_s^*\right\|_F^2
    &\leq (1-\delta)^{-1}\left\|\A\left(\mX\mX^{\top} - \mZ_s^*\right)\right\|_2^2 \\
    &\leq (1-\delta)^{-1}\left\|\A\left(\mX\mX^{\top} - \mX_s \mX_s^{\top}\right)\right\|_2^2 \\
    &\leq \frac{1+\delta}{1-\delta} \left\|\mX\mX^{\top} - \mX_s \mX_s^{\top}\right\|_F^2 \\
    &\leq \left\|\mX\mX^{\top} - \mX_s \mX_s^{\top}\right\|_F^2 + 10\delta\|\mX\mX^{\top}\|_F^2,
    \end{aligned}
\end{equation}
where the second inequality holds due to \Cref{def:best-rank}.
\end{proof}

We now recall \Cref{lemma:minima-close}.

\minimaClose*

We prove the statements in this lemma separately in \Cref{app:minima-close,landscape-cor-1} below.

\begin{lemma}
\label{app:minima-close}
Suppose that \Cref{asmp:rip} holds. Let $\mU_s^*$ be a global minimizer of $f_s$, then we have
\begin{equation}
    \notag
    \mathrm{dist}(\mU_s^*, \mX_s) \leq 40\delta\kappa\|\mX\|_F.
\end{equation}
\end{lemma}

\begin{proof}
Define
\begin{equation}
    \notag
    S = \left\{ \mU\in\R^{d\times s}: \mathrm{dist}(\mU,\mX_s) < 0.1\kappa^{-1}\|\mX\| \right\}.
\end{equation}
First we can show that $\mU_s^* \in S$. The main idea is to apply
\Cref{opt-prelim-1}. Indeed, it is easy to see that
\begin{equation}
    \notag
    \lim_{\|\mU\|_F\to +\infty} F_{s}(\mU) = +\infty,
\end{equation}
so the condition \textit{(1)} in \Cref{opt-prelim-1} holds. To check condition \textit{(2)}, we separately analyze the two cases $\mU\in\partial S$ and $\mU \notin S$.

\textit{Firstly}, let $\mU \in\partial S$, i.e., $\mathrm{dist}^2(\mU,\mX_s) = 0.1\|\mX\|^{-1}\tau$. Assume WLOG that $\mathrm{dist}(\mU,\mX_s) = \left\|\mU-\mX_s\right\|_F$, then by \Cref{matrix-factorization-PL} we have
\begin{equation}
    \notag
    \begin{aligned}
    F_{s}(\mU)-F_{s}(\mX_s) &= \int_0^1 t\left\langle \nabla F_{s}(t \mU+(1-t)\mX_s),\mU-\mX_s\right\rangle \d t \\
    &\geq \int_0^1 0.1\tau t^2\left\|\mU-\mX_s\right\|_F^2 \d t\\
    &\geq 10^{-3}\|\mX\|^{-2}\tau^3 = 10^{-3}\kappa^{-3}\|\mX\|^2. 
    \end{aligned}
\end{equation}

\textit{Secondly}, let $\mU\notin S$ be a stationary point of $f_s$. Recall that all the stationary points of $F_{s}$ are characterized in \Cref{fac-critical-point}, so that for all $\mU \notin S$ with $\nabla F_{s}(\mU) = 0$, we have
\begin{equation}
    \notag
    F_{s}(\mU) - F_{s}^* \geq 0.5 \left( \sigma_s^4-\sigma_{s+1}^4\right) \geq 0.5 \tau^2.
\end{equation}
On the other hand, we know from \Cref{func-value-close} that 
\begin{equation}
    \label{minima-close-eq1}
    F_{s}(\mU_s^*) - F_{s}^* \leq 5\delta r_*\|\mX\|^4.
\end{equation}
By \Cref{asmp:rip}, we have $5\delta r_* \|\mX\|^4 < 10^{-3}\kappa^{-3}\|\mX\|^2 <0.5 \tau^2$, so \Cref{opt-prelim-1} implies that $\mU_s^*\in S$.

Since $\nabla f_s(\mU_s^*) = 0$, we have $\A^*\A\left(\mX\mX^{\top}- \mU_s^*\left(\mU_s^*\right)^{\top} \right) \mU_s^* = 0$, so that
\begin{equation}
    \notag
    \begin{aligned}
    \left\|\nabla F_{s}(\mU_s^*)\right\|_F &= \left\| \left(\mX\mX^{\top}- \mU_s^*\left(\mU_s^*\right)^{\top} \right) \mU_s^*\right\|_F \\
    &= \left\| \left(\A^*\A-\mI\right) \left(\mX\mX^{\top}- \mU_s^*\left(\mU_s^*\right)^{\top} \right) \mU_s^* \right\|_F \\
    &\leq \delta \left\|\mX\mX^{\top}- \mU_s^*\left(\mU_s^*\right)^{\top} \right\|_F\left\|\mU_s^*\right\| \\
    &\leq 4\delta\|\mX\|\cdot \left\|\mX\mX^{\top}\right\|_F.
    \end{aligned}
\end{equation}
From $\mU_s^*\in S$ and \Cref{cor:matrix-factorization-PL} we can deduce that
\begin{equation}
    \notag
    \mathrm{dist}(\mU_s^*,\mX_s) \leq 40\delta\tau^{-1}\|\mX\|^2\|\mX\|_F = 40\delta\kappa\|\mX\|_F.
\end{equation}
\end{proof}

\begin{corollary}
\label{landscape-cor-1}
Suppose that \Cref{asmp:rip} holds, then we have $\left\|\mZ_s^*-\mX_s\mX_s^\top\right\|_F \leq 80\delta\kappa\sqrt{r_*}\|\mX\|^2$ and $\sigma_{\min}\left(\left(\mU_s^*\right)^{\top} \mU_s^*\right) \geq \sigma_s^2 - 80\delta\kappa\sqrt{r_*}\|\mX\|^2$.
\end{corollary}

\begin{proof}
We assume WLOG that $\left\|\mU_s^*-\mX_s\right\|_F = \mathrm{dist}(\mU_s^*,\mX_s)$ i.e. $\mR = \mI$ in \Cref{def-procrutes}.
By \Cref{lemma:minima-close}, we have that
\begin{equation}
    \notag
    \begin{aligned}
    \left\| \mU_s^*\left(\mU_s^*\right)^{\top} - \mX_s\mX_s^{\top}\right\|_F
    &\leq 2\max\left\{ \left\|\mU_s^*\right\|, \left\|\mX_s\right\|\right\}\cdot \left\|\mU_s^*-\mX_s\right\|_F\\
    &\leq 160\delta\kappa\|\mX\|\|\mX\|_F
    \leq 160\delta\kappa\sqrt{r_*}\|\mX\|^2.
    \end{aligned}
\end{equation}

which proves the first inequality. Similarly, we have
\begin{equation}
    \notag
    \left\|\left(\mU_s^*\right)^{\top}\mU_s^* -\mX_s^{\top}\mX_s\right\|_F \leq 160\delta\kappa\sqrt{r_*}\|\mX\|^2.
\end{equation}
Hence $\sigma_s^2-\sigma_{\min}\left(\left(\mU_s^*\right)^{\top} \mU_s^*\right) \leq \left\|\left(\mU_s^*\right)^{\top}\mU_s^* -\mX_s^{\top}\mX_s\right\| \leq 160\delta\kappa\sqrt{r_*}\|\mX\|^2$, as desired.
\end{proof}

\minimaCloseCor*

\begin{proof}
    \Cref{asmp:rip} implies that $160\delta\kappa\sqrt{r_*}\|\mX\|^2 \leq 0.1\kappa^{-1}\|\mX\|^2 \leq 0.1\sigma_s^2$, so that the conclusion immediately follows from \Cref{landscape-cor-1}.
\end{proof}

\begin{lemma}
    \label{app-lemma:main-landscape}
    Under \Cref{asmp:rip},
    suppose that $\mU, \mU_s^*\in\R^{d\times s}$ such that $\mU_s^*$ is a global minimizer of $f_s$ and $\|\mU-\mU_s^*\|_F = \mathrm{dist}(\mU,\mU_s^*) \leq 10^{-2}\kappa^{-1}\|\mX\|$ (recall that $\mathrm{dist}$ is defined in \Cref{def-procrutes}),
    then we have
    \begin{equation}
    \notag
    \left\langle \nabla f_s(\mU), \mU -
    \mU_s^*\right\rangle \geq 0.1\kappa^{-1}\|\mX\|^2
    \| \mU - \mU_s^* \|_F^2.
    \end{equation}
\end{lemma}

\begin{proof}
By \Cref{lm:R-orth},
$\mU^{\top} \mU_s^*$ is symmetric and positive semi-definite. Let $\mH = \mU - \mU_s^*$, then
\begin{equation}
    \notag
    \begin{aligned}
    \nabla f_s(\mU) &= \left(\A^*\A\right)( \mU\mU^{\top} - \mX\mX^{\top}) \mU \\
    &= \left(\A^*\A\right)\left[ (\mH + \mU_s^*)(\mH + \mU_s^*)^{\top} - \mX \mX^{\top}\right] (\mH + \mU_s^*) \\
    &= \left[\left(\A^*\A\right)\left( \mH \mH^{\top} + \mU_s^* \mH^{\top} + \mH \left(\mU_s^*\right)^{\top}\right)\right] (\mH+\mU_s^*) - \A^*\A \left( \mX \mX^{\top} - \mU_s^*\left(\mU_s^*\right)^{\top}\right) \mH
    \end{aligned}
\end{equation}
where we use the first-order optimality condition 
\begin{equation}
    \notag
    \A^*\A \left( \mX \mX^{\top} - \mU_s^*\left(\mU_s^*\right)^{\top}\right) \mU_s^* = 0.
\end{equation}
Since $\left\|\mU_s^*\right\| \leq 2\|\mX\|$ by \Cref{lemma:minima-close}, we may thus deduce that
\begin{equation}
    \notag
    \begin{aligned}
    &\quad \left\| \nabla f_s(\mU) - \left[ \left( \mH \mH^{\top} + \mU_s^* \mH^{\top} + \mH \left(\mU_s^*\right)^{\top}\right) (\mH+\mU_s^*) -  \left( \mX \mX^{\top} - \mU_s^*\left(\mU_s^*\right)^{\top}\right) \mH \right]\right\|_F \\
    &\leq \left\|\left(\A^*\A-\mI\right)\left( \mH \mH^{\top} + \mU_s^* \mH^{\top} + \mH \left(\mU_s^*\right)^{\top}\right) (\mH+\mU_s^*)\right\|_{F} + \left\|\left(\A^*\A-\mI\right)\left( \mX \mX^{\top} - \mU_s^*\left(\mU_s^*\right)^{\top}\right) \mH\right\| \\
    &\leq 50\delta\|\mX\|^2\|\mH\|_{F}
    \end{aligned}
\end{equation}
Hence
\begin{equation}
    \notag
    \begin{aligned}
    &\quad \left\langle \nabla f_s(\mU), \mU - \mU_s^*\right\rangle \\
    &\geq \left\langle \left( \mH \mH^{\top} + \mU_s^* \mH^{\top} + \mH \left(\mU_s^*\right)^{\top}\right) (\mH+\mU_s^*) -  \left( \mX \mX^{\top} - \mU_s^*\left(\mU_s^*\right)^{\top}\right) \mH, \mH \right\rangle -50\delta\|\mX\|^2\|\mH\|_{F}^2 \\
    &\geq \mathrm{tr}\left(\mH(\mH+\mU_s^*)^{\top} (\mH+\mU_s^*)\mH^{\top}+\mH^{\top} \mU_s^* \mH^{\top} \mH + \left( \left(\mU_s^*\right)^{\top} \mH\right)^2 \right.\\
    &\quad \left.- \mH^{\top} \left( \mX \mX^{\top} - \mU_s^*\left(\mU_s^*\right)^{\top}\right) \mH\right) -50\delta\|\mX\|^2\|\mH\|_F^2 \\
    &\geq \left[\sigma_{\min}\left(\left(\mU_s^*\right)^{\top} \mU_s^*\right) - \left\|\mX \mX^{\top} - \mU_s^*\left(\mU_s^*\right)^{\top}\right\|-50\delta\|\mX\|^2-3\|\mU_s^*\|\|\mH\| - \|\mH\|^2\right] \|\mH\|_F^2.
    \end{aligned}
\end{equation}
By \Cref{landscape-cor-1} we have $\sigma_{\min}\left(\left(\mU_s^*\right)^{\top} \mU_s^*\right) \geq \sigma_s^2 - 80\delta\kappa\|\mX\|\|\mX\|_F$ and $\left\|\mX \mX^{\top} - \mU_s^*\left(\mU_s^*\right)^{\top}\right\| \leq \sigma_{s+1}^2 + 80\delta\kappa\|\mX\|_F^2$, so that
\begin{equation}
    \notag
    \left\langle \nabla f_s(\mU), \mU - \mU_s^*\right\rangle \geq \left( \sigma_{s}^2-\sigma_{s+1}^2 - 160\delta\kappa\|\mX\|\|\mX\|_F-50\delta\|\mX\|^2-3\|\mU_s^*\|\|\mH\| - \|\mH\|^2\right)\|\mH\|_F^2.
\end{equation}
When \Cref{asmp:rip} on $\delta$ is satisfied and $\|\mH\| \leq 10^{-2}\tau\|\mX\|^{-1}$, the above implies that $\left\langle \nabla f_s(\mU), \mU - \mU_s^*\right\rangle \geq 0.5\tau\|\mH\|_F^2$, as desired.
\end{proof}

We now prove \Cref{landscape-main-global-unique,landscape-main-thm} to conclude this section. Both results follow immediately from \Cref{app-lemma:main-landscape}.

\mainGlobalUnique*

\begin{proof}
    By \Cref{app:minima-close} we have that $\mathrm{dist}(\mU_s^*,\mX_s^*) \leq 40\delta\kappa\|\mX\|_F$ holds for \textit{any} $\mU_s^* \in \argmin f_s$. Suppose now that $\mU_s^*,\hat{\mU}_s^* \in \argmin f_s$ such that $\mathrm{dist}(\mU_s^*,\hat{\mU}_s^*) > 0$. Then we also have that $\mathrm{dist}(\mU_s^*,\hat{\mU}_s^*) \leq \mathrm{dist}(\mU_s^*,\mX_s) + \mathrm{dist}(\mX_s,\hat{\mU}_s^*) \leq 80\delta\kappa\|\mX\|_F < 10^{-2}\kappa^{-1}\|\mX\|$, where the last inequality follows from \Cref{asmp:rip}. Without loss of generality, we can assume that $\|\mU_s^*-\hat{\mU}_s^*\|_F = \mathrm{dist}(\mU_s^*,\hat{\mU}_s^*)$, so that we can apply \Cref{app-lemma:main-landscape} to obtain
    \begin{equation}
    \notag
        \left\langle \nabla f_s(\hat{\mU}_s^*), \hat{\mU}_s^* -
        \mU_s^*\right\rangle \geq 0.1\kappa^{-1}\|\mX\|^2
        \| \hat{\mU}_s^* - \mU_s^* \|_F^2.
    \end{equation}
    However, since $\hat{\mU}_s^*$ is a global minimizer of $f_s$, we have $\nabla f_s(\hat{\mU}_s^*)$ which is a contradiction. Thus the global minimizer must be unique under the procrustes distance.
\end{proof}

We recall \Cref{def:procrutes-projection} which is now guaranteed to be well-defined by \Cref{landscape-main-global-unique}.

\procrustesProjection*

Equipped with this definition, \Cref{app-lemma:main-landscape} directly translates into \Cref{landscape-main-thm}:

\mainLandscape*

\section{Proofs for \Cref{main_result,main-thm-cor}}
\label{appsec-main-proof}
In this section, we prove the main theorems based on our key lemmas introduced in \Cref{subsec:key-lemma}.

Based on \Cref{landscape-main-thm}, we first prove the following lemma, which shows that GD initialized near global minimizers
converges linearly.


\begin{lemma}
\label{lm:GD-rsi}
Suppose that \Cref{asmp:rip,asmp:step-size} hold.
Let $\{ \hat{\mU}_t \}_{t \ge 0}$ be a trajectory of GD that optimizes $f_s$ with step size $\mu$,
starting with $\hat{\mU}_0$.
Also let $\mU_s^*$ be a global minimizer of $f_s$.
If $\mathrm{dist}(\hat{\mU}_0, \mU_s^*) \le 10^{-2} \kappa^{-1} \|\mX\|$, then for all $t \ge 0$,
\begin{equation} \label{eq:GD-rsi-dist}
    \mathrm{dist}^2(\hat{\mU}_{t},\mU_s^*) \leq \left( 1-0.05\tau\mu\right)^t \mathrm{dist}^2(\hat{\mU}_{\alpha,0},\mU_s^*).
\end{equation}
\end{lemma}
\begin{proof}
    We prove \Cref{eq:GD-rsi-dist} by induction.
    It is easy to check that \Cref{eq:GD-rsi-dist} holds for $t = 0$.
    Now we show that \Cref{eq:GD-rsi-dist} holds for $t + 1$ assuming it holds for $t$.

    Since $\mathrm{dist}(\hat{\mU}_0, \mU_s^*) \le 10^{-2} \kappa^{-1} \|\mX\|$, we have
\begin{equation} \label{eq:GD-rsi-hat-mU-bound}
    \left\|\hat{\mU}_{0}\right\| \leq \|\mU_s^*\| + 10^{-2} \kappa^{-1} \|\mX\| \leq 2\|\mX\|.
\end{equation}
Let $\mR$ be the orthogonal matrix such that $\mU_s^* \mR \in \Pi(\mU_t)$,
then $\| \mU -\mU_s^* \mR \|_F = \mathrm{dist}(\mU_t,\mU_s^*)$. We first bound the gradient $\nabla f(\hat{\mU}_{\alpha,t})$ as follows:
\begin{equation}
    \label{gradient-lipschitz}
    \begin{aligned}
    \left\|\nabla f(\hat{\mU}_{\alpha,t})\right\|_F
    &= \left\|\A^*\A\left( \mX\mX^{\top} - \hat{\mU}_{\alpha,t}\hat{\mU}_{\alpha,t}^{\top}\right)\hat{\mU}_{\alpha,t}\right\|_F \\
    &\leq \left\|\A^*\A\left( \mX\mX^{\top} - \hat{\mU}_{\alpha,t}\hat{\mU}_{\alpha,t}^{\top}\right)\right\|\left\|\hat{\mU}_{\alpha,t} - \mU_s^*\right\|_F + \left\|\left( \hat{\mU}_{\alpha,t}\hat{\mU}_{\alpha,t}^{\top} - \mU_s^* \left( \mU_s^*\right)^{\top}\right) \mU_s^*\right\|_F \\
    &\leq 20\|\mX\|^2\left\|\hat{\mU}_{\alpha,t} - \mU_s^*\right\|_F
    \end{aligned}
\end{equation}
where we use \Cref{eq:GD-rsi-hat-mU-bound} and the RIP property (\Cref{asmp:rip}).
It follows that
\begin{subequations}
    \label{GD-PL-eq1}
    \begin{align}
        \mathrm{dist}^2(\hat{\mU}_{t+1},\mU_s^*)
        &\leq \left\| \hat{\mU}_{t+1} - \mU_s^* \mR\right\|_F^2 \label{GD-PL-eq1-1}\\
        &= \left\| \hat{\mU}_{t} - \mu \nabla f(\hat{\mU}_{\alpha,t}) - \mU_s^* \mR\right\|_F^2 \nonumber \\
        &= \left\| \hat{\mU}_{\alpha,t} - \mU_s^* \mR\right\|_F^2 - \mu\left\langle \nabla f(\hat{\mU}_{\alpha,t}), \hat{\mU}_{t} - \mU_s^* \mR\right\rangle + \mu^2 \left\|\nabla f(\hat{\mU}_{\alpha,t})\right\|_F^2\nonumber \\
        &\leq \left(1-0.1\tau\mu + 400\|\mX\|^4\mu^2\right)\left\| \hat{\mU}_{\alpha,t} - \mU_s^* \mR\right\|_F^2   \label{GD-PL-eq1-2}
    \end{align}
\end{subequations}
where \Cref{GD-PL-eq1-1} follows from the definition of $\mathrm{dist}$, and
\Cref{GD-PL-eq1-2} is due to \Cref{landscape-main-thm} and
\Cref{gradient-lipschitz}. Finally, \Cref{eq:GD-rsi-dist} follows from
\Cref{asmp:step-size}.
\end{proof}

We then note the following proposition, which
is straightforward from \Cref{continue-induct} and \Cref{main-thm-formal}. In
the following we use $\mU_{\alpha,t}$ to denote the $t$-th iteration of GD when
initialized at $\mU_0=\alpha \bar{\mU}$.

\begin{proposition}
\label{approx-rank-s}
Suppose that \Cref{asmp:rip,asmp-full-rank,asmp:step-size} hold and $\alpha$ is sufficiently small.
Then there exist matrices $\mU_{\alpha,t}^{\mathtt{lr}}$  for $t = -T_{\alpha,s}^{\mathtt{ft}}, -T_{\alpha,s}^{\mathtt{ft}}+1, \cdots, 0$ with rank $\leq s$ (where $\mathtt{lr}$ stands for low rank) and a constant $C_6=C_6(\mX,\bar{\mU})$ (defined in \Cref{def-C6}) such that
\begin{equation}
    \notag
    \max_{T_{\alpha,s}^{\mathtt{ft}} \leq t \leq 0}\left\| \mU_{\alpha,t}^{\mathtt{lr}} - \mU_{\alpha,T_{\alpha,s}^{\mathtt{ft}}+t}\right\|_F = C_6\cdot \alpha^{\frac{1}{4\kappa}}
\end{equation}
where $T_{\alpha,s}^{\mathtt{ft}}$ is defined in \Cref{main-thm-formal} and moreover
\begin{equation}
    \notag
    \left\|\mU_{\alpha,0}^{\mathtt{lr}}\left(\mU_{\alpha,0}^{\mathtt{lr}}\right)^{\top} - \mZ_s^*\right\|_F \leq 2\times 10^5 \kappa^3\|\mX\|^2 r_*\delta.
\end{equation}
where $\mZ_s^* = \mU_s^*\left(\mU_s^*\right)^{\top}$ is the best rank-$s$ solution as defined in \Cref{def:best-rank}.
\end{proposition}

\begin{proof}
It follows from \Cref{cor-C6} that $\max_{1\leq t\leq T_{\alpha,s}^{\mathtt{ft}}}\left\|\mU_{t}\mW_{t,\perp}\right\| \leq C_6(\mX,\bar{\mU})\cdot\alpha^{\frac{1}{4\kappa}}$
(recall that $C_5$ is defined in \Cref{refinement-start} and $T_{\alpha,s}^{\mathtt{ft}}$ is defined in \Cref{def-hat-T}). We choose $\mU_{\alpha,t}^{\mathtt{lr}} = \mU_{T_{\alpha,s}^{\mathtt{ft}}+t}\mW_{T_{\alpha,s}^{\mathtt{ft}}+t}\mW_{T_{\alpha,s}^{\mathtt{ft}}+t}^{\top}$, then $\rank{\bar{\mU}_t}\leq s$ and moreover by \Cref{main-thm-formal} we have $\left\|\mX_s \mX_s^{\top} - \mU_{\alpha,0}^{\mathtt{lr}}\left(\mU_{\alpha,0}^{\mathtt{lr}}\right)^{\top}\right\|_F \leq 10^5\kappa^3\|\mX\|^2 r_*\delta$. On the other hand, by \Cref{lemma:minima-close} we have that $\left\|\mZ_s^*-\mX_s \mX_s^{\top}\right\|_F\leq 80\delta\kappa\sqrt{r_*}\|\mX\|^2$. Thus $\left\| \mU_{\alpha,0}^{\mathtt{lr}}\left(\mU_{\alpha,0}^{\mathtt{lr}}\right)^{\top} - \mZ_s^*\right\|_F \leq 2\times 10^5\kappa^3\|\mX\|^2 r_*\delta$ as desired.
\end{proof}

Let $\hat{\mU}_{\alpha,0} = \mU_{T_{\alpha,s}^{\mathtt{ft}}}\mW_{T_{\alpha,s}^{\mathtt{ft}}} \in \R^{d\times s}$, then it satisfies $\hat{\mU}_{\alpha,0}\hat{\mU}_{\alpha,0}^{\top} = \mU_{\alpha,0}^{\mathtt{lr}}\left(\mU_{\alpha,0}^{\mathtt{lr}}\right)^{\top}$. The following corollary shows that $\hat{\mU}_{\alpha,0}$ is close to $\mU_s^*$ in terms of the procrustes distance.

\begin{corollary}
\label{approx-rank-s-cor}
We have $\mathrm{dist}(\hat{\mU}_{\alpha,0},\mU_s^*) \leq 3\times 10^{6}\kappa^4 r_* \|\mX\|\delta$.
\end{corollary}

\begin{proof}
We know from \Cref{lemma:minima-close} that $\mathrm{dist}(\mU_s^*,\mX_s) \leq 40\delta\kappa\|\mX\|_F$, so it remains to bound $\mathrm{dist}(\hat{\mU}_{\alpha,0}, \mX_s)$.

The proof idea is the same as that of \Cref{lemma:minima-close}, so we only provide a proof sketch here. It has been shown in the proof of \Cref{approx-rank-s} that
\begin{equation}
    \notag
    F_s(\hat{\mU}_{\alpha,0}) := \frac{1}{2}\left\| \mX_s \mX_s^{\top} - \hat{\mU}_{\alpha,0}\hat{\mU}_{\alpha,0}^{\top}\right\|_F^2 \leq r_* \left\| \mX_s \mX_s^{\top} - \hat{\mU}_{\alpha,0}\hat{\mU}_{\alpha,0}^{\top}\right\|^2 \leq 4\times 10^{10}\kappa^6 r_*^2 \|\mX\|^4\delta^2 \leq 0.5\tau^2.
\end{equation}
Note that $F_s$ is the matrix factorization loss with $\mX_s \mX_s^{\top}$ being
the ground-truth, so the local RSI condition (\Cref{matrix-factorization-PL})
still holds. By the same reason as \Cref{minima-close-eq1}, we deduce
that $\mathrm{dist}(\hat{\mU}_{\alpha,0},\mX_s)\leq 0.1\|\mX\|^{-1}\tau$, i.e.,
$\hat{\mU}_{\alpha,0}$ is in the local region around $\mX_s$ in which the RSI condition holds. Finally, it
follows from the local RSI condition that
\begin{equation}
    \notag
    \mathrm{dist}(\hat{\mU}_{\alpha,0},\mX_s)\leq 10\tau^{-1}\left\|\nabla F_s(\hat{\mU}_{\alpha,0})\right\|_F \leq 10\tau^{-1}\|\hat{\mU}_{\alpha,0}\|\left\|\mX_s \mX_s^{\top} - \hat{\mU}_{\alpha,0}\hat{\mU}_{\alpha,0}^{\top}\right\|_F \leq 3\times 10^6\kappa^4 r_*\|\mX\|\delta.
\end{equation}
The conclusion follows.
\end{proof}

We are now ready to complete the proof of \Cref{main_result,main-thm-cor}.

\mainThmUnderParam*

\begin{proof}
    When $\hr \leq r_*$, the parameterization itself ensures that
    $\mU_{\alpha,t}$ is low-rank, so that we can choose
    $\mU_{\alpha,t}^{\mathtt{lr}}=\mU_{\alpha,T_{\alpha,\hr}^{\mathtt{ft}}+t}$
    and $\hat{\mU}_{\alpha,0}=\mU_{\alpha,T_{\alpha,s}^{\mathtt{ft}}}$ in
    \Cref{approx-rank-s,approx-rank-s-cor} (for $s=\hr$). The proof that these
    choices satisfy all required conditions are identical to our proofs for
    these two lemmas in the general setting, and we omit them here.

    Applying \Cref{lm:GD-rsi}, we can thus deduce that $\lim_{t\to
    +\infty}\mathrm{dist}(\hat{\mU}_{\alpha,t},\mU_{\hr}^*) = 0$. This means
    that $\lim_{t\to +\infty}\mathrm{dist}(\mU_{\alpha,t},\mU_{\hr}^*) = 0$.
    Recall that $\mZ_{\hr}^* = \mU_{\hr}^*{\mU_{\hr}^*}^\top$, so the conclusion
    immediately follows.
\end{proof}

\mainThm*

\begin{proof}
Recall that $\left\|\mU_{T_{\alpha,s}^{\mathtt{ft}}}-\bar{\mU}_0\right\|_F = o(1)$ ($\alpha \to 0$) where $ T_{\alpha,s}^{\mathtt{ft}}$ is defined in \Cref{approx-rank-s}; we omit the dependence on $\alpha$ to simplify notations. We also note that by the update of GD, we have $\bar{\mU}_t\bar{\mU}_t^{\top} = \hat{\mU}_{\alpha,t}\hat{\mU}_{\alpha,t}^{\top}$ for all $t\geq 0$.

By \Cref{lm:GD-rsi}, we have that $\mathrm{dist}^2(\hat{\mU}_{\alpha,t},\mU_s^*) \leq \left( 1-0.05\tau\mu\right)^t \mathrm{dist}^2(\hat{\mU}_{\alpha,0},\mU_s^*)$ and, in particular, $\left\|\hat{\mU}_{\alpha,t}\right\| \leq 2\|\mX\|$ for all $t$. Thus $\left\|\bar{\mU}_t\right\| \leq 2\|\mX\|$ as well. Moreover, recall that $\|\mU_t\|\leq 3\|\mX\|$ for all $t$. It's easy to see that that the matrix sensing loss $f$ is $L$-smooth in $\left\{ \mU\in\R^{d\times r}: \|\mU\|\leq 3\|\mX\|\right\}$ for some constant $L=\O(\|\mX\|^2)$, so it follows from \Cref{gd-error} that
\begin{equation}
    \notag
    \left\|\mU_{T_{\alpha,s}^{\mathtt{ft}}+t}-\bar{\mU}_t\right\|_F \leq (1+\mu L)^t \left\|\mU_{T_{\alpha,s}^{\mathtt{ft}}}-\bar{\mU}_0\right\|_F.
\end{equation}
On the other hand, since $\mathrm{dist}^2(\hat{\mU}_{\alpha,t},\mU_s^*) \leq \left( 1-0.05\tau\mu\right)^t \mathrm{dist}^2(\hat{\mU}_{\alpha,0},\mU_s^*)$, we can deduce that
\begin{equation}
    \notag
    \begin{aligned}
    \left\| \mU_{T_{\alpha,s}^{\mathtt{ft}}+t}\mU_{T_{\alpha,s}^{\mathtt{ft}}+t}^{\top} - \mZ_s\right\|_F 
    &\leq \left\| \mU_{T_{\alpha,s}^{\mathtt{ft}}+t}\mU_{T_{\alpha,s}^{\mathtt{ft}}+t}^{\top} - \bar{\mU}_t\bar{\mU}_t^{\top}\right\|_F + \left\|\bar{\mU}_t\bar{\mU}_t^{\top}-\mU_s^* \left(\mU_s^*\right)^{\top}\right\|_F\\
    &= \left\| \mU_{T_{\alpha,s}^{\mathtt{ft}}+t}\mU_{T_{\alpha,s}^{\mathtt{ft}}+t}^{\top} - \bar{\mU}_t\bar{\mU}_t^{\top}\right\|_F + \left\|\hat{\mU}_{\alpha,t}\hat{\mU}_{\alpha,t}^{\top}-\mU_s^* \left(\mU_s^*\right)^{\top}\right\|_F\\
    &\leq 3\|\mX\|\left( \left\|\mU_{T_{\alpha,s}^{\mathtt{ft}}+t}-\bar{\mU}_t\right\|_F + \mathrm{dist}(\hat{\mU}_{\alpha,t},\mU_s^*)\right)\\
    &\leq 3\|\mX\|\left( (1+\mu L)^t \left\|\mU_{T_{\alpha,s}^{\mathtt{ft}}}-\bar{\mU}_0\right\|_F + \left( 1-0.05\tau\mu\right)^{\frac{t}{2}} \mathrm{dist}^2(\hat{\mU}_{\alpha,0},\mU_s^*)\right)
    \end{aligned}
\end{equation}
Since when $\alpha\to 0$, $\left\|\mU_{T_{\alpha,s}^{\mathtt{ft}}}-\bar{\mU}_0\right\|_F = \O(\alpha^{\frac{1}{4\kappa}})$, it's easy to see that there exists a time $t = t_{\alpha}^s$ so that we have $\max_{-T_{\alpha,s}^{\mathtt{ft}}\leq t\leq t_{\alpha}^s}\left\|\mU_{T_{\alpha,s}^{\mathtt{ft}}+t}-\bar{\mU}_t\right\|_F = \O\left(\alpha^{\frac{1}{M_1\kappa^2}}\right)$ and $\left\| \mU_{T_{\alpha,s}^{\mathtt{ft}}+t}\mU_{T_{\alpha,s}^{\mathtt{ft}}+t}^{\top} - \mZ_s\right\|_F = \O\left(\alpha^{\frac{1}{M_1\kappa^2}}\right)$ as well, where $c_1$ is a universal constant. Let $T_{\alpha}^s = T_{\alpha,s}^{\mathtt{ft}}+t_{\alpha}^s$, then $\left\| \mU_{T_{\alpha}^{s}}\mU_{T_{\alpha}^{s}}^{\top} - \mZ_s\right\|_F =o(1)$ holds. Recall that $\rank{\mU_t}\leq s$, so that $\max_{0\leq t\leq T_{\alpha}^s}\sigma_{s+1}\left(\mU_t\right)=o(1)$. Finally, for all $0\leq s < \hr\wedge r_*$, we need to show that $T_{\alpha}^s < T_{\alpha}^{s+1}$. Indeed, by \Cref{landscape-cor-1} and the \Cref{asmp:rip} we have $\sigma_{s+1}^2\left(\mU_{T_{\alpha}^{s}}\right) \geq \sigma_{s+1}\left(\mZ_{s+1}\right)-o(1) \geq 0.5\sigma_{s+1}^2$, so that $T_{\alpha}^{s+1} > T_{\alpha}^s$, as desired.
\end{proof}

\section{The landscape of matrix sensing with rank-$1$ parameterization}
\label{appendix_rank1}
In this section, we establish a local strong-convexity result
\Cref{rank1-sc-lma-restated} for rank-$1$ parameterized matrix sensing. This
result is stronger than the RSI condition we established for general ranks,
though the latter is sufficient for our analysis.

\begin{lemma}
\label{landscape-lma1}
Define the full-observation loss with rank-$1$ parameterization
\begin{equation}
    \notag
    g_1(\vu) = \frac{1}{4}\left\| \vu \vu^T - \mX \mX^T\right\|_{F}^2.
\end{equation}
Then the global minima of $g_1$ are $\vu^* = \sigma_{1}v_1$ and $-\vu^*$. Moreover, suppose that $g(\vu) - g(\vu^*) \leq 0.5\tau_1$ where $\tau_1=\sigma_1^2-\sigma_2^2$ is the eigengap, then we must have
\begin{equation}
    \notag
    \left\| \vu - \vu^* \right\|^2 \leq 20\tau_1^{-1}\left( g_1(\vu)-g_1(\vu^*)\right).
\end{equation}
\end{lemma}

\begin{proof}
We can assume WLOG that $\mX\mX^T = \diag{\sigma_1^2,\cdots,\sigma_{r_*}^2,0,\cdots,0}$. Then
\begin{subequations}
    \label{sts-ineq1}
    \begin{align}
    g_1(\vu) &= \frac{1}{4}\left( \|\vu\|_2^4 - 2\sum_{i=1}^s \sigma_i^2 \vu_i^2 + \|\mX^T \mX\|_F^2  \right) \label{sts-ineq-1-1} \\
    &\geq \frac{1}{4}\left( \|\vu\|_2^4 - 2\sigma_1^2 \|\vu\|_2^2 + \|\mX^T \mX\|_F^2  \right) \label{sts-ineq-1-2} \\
    &\geq \frac{1}{4}\left( \|\mX^T \mX\|_F^2  - \sigma_1^4 \right) \label{sts-ineq-1-3}
    \end{align}
\end{subequations}
where equality holds if and only if $\vu_2=\cdots=\vu_d=0$ and $\|\vu\|^2=\sigma_1^2$ i.e. $\vu= \pm\sigma_1 \ve_1$. Moreover, suppose that $g_1(\vu) - g_1(\vu^*) \leq 0.5\tau_{1}$, it follows from \Cref{sts-ineq-1-2} that $\tau_{1}\sum_{i=2}^d \vu_i^2 \leq 2(g_1(\vu)-g_1(\vu^*))$ which implies that $\sum_{i=2}^d \vu_i^2 \leq 2\tau_{1}^{-1}\left( g_1(\vu)-g_1(\vu^*)\right)$. Also \Cref{sts-ineq-1-3} yields $\left|\|\vu\|^2 - \sigma_1^2\right| \leq 4\sqrt{g_1(\vu)-g_1(\vu^*)}$. Assume WLOG that $\vu_1>0$, then we have
\begin{equation}
    \notag
    \begin{aligned}
    \left\| \vu-\sigma_1 \ve_1\right\|^2
    &\leq \sigma_1^{-2}\left( \vu_1^2-\sigma_1^2\right)^2 + \sum_{i=2}^d \vu_i^2 \\
    &\leq 20\tau_{1}^{-1}\left( g_1(\vu)-g_1(\vu^*)\right).
    \end{aligned}
\end{equation}
\end{proof}

\begin{lemma}
\label{rank1-sc-lma-restated}
Let
\begin{equation}
    \notag
    \label{rank1-objective}
    f_1(\vu) = \frac{1}{4}\left\|\A\left(\vu \vu^T - \mX \mX^T\right)\right\|_2^2,\quad \vu\in\R^d.
\end{equation}
Suppose that $\delta\leq 10^{-3}\|\mX\|^{-2}\tau_{1}$, then there exists constants $a_1$ and $\iota$, such that $f_1$ is locally $\iota$-strongly convex in $\mathcal{B}_1 = \mathcal{B}(\sigma_1 v_1, a_1) \subset \R^d$. Furthermore, there is a unique global minima of $f_1$ inside $\mathcal{B}_1$.
\end{lemma}

\begin{proof}
Recall that we defined the full observation loss $g_1(\vu) = \frac{1}{4}\left\| \vu \vu^T - \mX \mX^T\right\|_{F}^2$.
Let $h_1 = f_1 - g_1$, then
\begin{equation}
    \notag
    \begin{aligned}
    \left\|\nabla^2 h_1(\vu)\right\|
    &= \frac{1}{2}\left\| \left(\A^*\A-\mI\right)(\vu \vu^T - \mX \mX^T) + 2\left(\A^*\A-I\right) \vu \vu^T\right\| \\
    &\leq \delta\left( 2\|\vu\|^2 + \|\mX\|^2\right).
    \end{aligned}
\end{equation}
When $\left\| \vu-\sigma_1 v_1\right\|^2 \leq 0.1\min\left\{ \sigma_1^2, \tau_{1}\right\}$ (recall $\tau_1 = \sigma_1^2-\sigma_{2}^2$),
\begin{equation}
    \notag
    \sigma_{\min }\left(\nabla^{2} g_{1}(\vu)\right) = \frac{1}{2}\sigma_{\min}\left(\|\vu\|^{2} I+2 \vu \vu^{T}-\mX \mX^{T}\right) \geq 0.4\tau_{1}.
\end{equation}
Hence we have
\begin{equation}
    \notag
    \sigma_{\min}\left(\nabla^{2} f_{1}(\vu)\right) \geq \left(\nabla^{2} g_{1}(\vu)\right) - \|\nabla^2 h_1(\vu)\| \geq 0.4\tau_{1} - 4\|\mX\|^2\delta \geq 0.2\tau_{1},
\end{equation}
i.e. strong-convexity holds for $a_1^2 = 0.1\min\left\{ \sigma_1^2, \tau_1\right\}$ and $\iota = 0.2\tau_{1}$.

Let $\vu^*$ be a global minima of $f_1$, then we must have $\|\vu^*\| \leq 2\|\mX\|$ (otherwise $f_1(\vu)>f_1(0)$). We can thus deduce that
\begin{equation}
    \notag
    \begin{aligned}
    g_1(\vu^*) &\leq f_1(\vu^*) + \frac{1}{4}\left|\left\langle \vu \vu^T - \mX \mX^T, (\A^*\A-I)(\vu \vu^T - \mX \mX^T)\right\rangle\right| \\
    &\leq f_1(\vu) + 10\delta\|\mX\|^2 \leq g_1(\vu) + 20\delta\|\mX\|^2.
    \end{aligned}
\end{equation}
It follows from \Cref{landscape-lma1} and our assumption on $\delta$ that $\min\left\{ \left\| \vu^* - \sigma_1 \vv_1\right\|^2, \left\| \vu^* + \sigma_1 \vv_1\right\|^2\right\} \leq \frac{1}{2}a_1^2$. Moreover, by strong convexity, there exists only one global minima in $\mathcal{B}_1$, which concludes the proof.
\end{proof}

\end{document}